\documentclass[10pt]{article}
\usepackage[a4paper,top=2cm,bottom=2cm,left=3cm,right=3cm,marginparwidth=1.75cm]{geometry}
\usepackage{amsmath,amsfonts}
\usepackage{array}
\usepackage{textcomp}
\usepackage{stfloats}
\usepackage{url}
\usepackage{verbatim}
\usepackage{graphicx}
\usepackage{cite}
\usepackage{amssymb}
\usepackage{subfigure}
\usepackage{xcolor}
\usepackage{tabularx}
\usepackage{booktabs}
\usepackage{bm}
\usepackage{ragged2e}
\usepackage{authblk}
\usepackage{multirow}
\usepackage{hyperref}
\usepackage[square,numbers]{natbib}
\newcommand{\fq}[1]{{\color{black}{#1}}}

\makeatletter
\def\thanks#1{\protected@xdef\@thanks{\@thanks
        \protect\footnotetext{#1}}}
\makeatother

\begin{document}

\title{Hierarchical Vectorization for Portrait Images}

\author{Qian Fu$^{*}$,
Linlin Liu$^{*}$,
Fei Hou and Ying He
\thanks{$^{*}$ Equal contribution.}
\thanks{Qian Fu and Ying He are with the School of Computer Science and Engineering, Nanyang Technological University, Singapore. E-mail: \{qfu004, yhe\}@ntu.edu.sg.}
\thanks{Linlin Liu is with the Interdisciplinary Graduate School, Nanyang Technological University, Singapore and Alibaba Group. E-mail: linlin001@ntu.edu.sg.}
\thanks{Fei Hou is with the Institute of Software, Chinese Academy of Sciences. E-mail: houfei@ios.ac.cn.}}



\maketitle

\begin{abstract}
The explosive growth of social media makes portrait editing and retouching in high demand. Though portraits are commonly captured and stored as raster images, editing raster images is non-trivial and often requires the user to be highly skilled. Aiming at developing intuitive and easy-to-use portrait editing tools, we propose a novel vectorization method that can automatically convert raster images into a 3-tier hierarchical representation. The base layer consists of a set of sparse diffusion curves (DC) which characterize salient geometric features and low-frequency colors and provide means for semantic color transfer and facial expression editing. The middle level encodes specular highlights and shadows to large and editable Poisson regions (PR) and allows the user to directly adjust illumination via tuning the strength and/or changing shape of PR.
The top level contains two types of pixel-sized PRs for high-frequency residuals and fine details such as pimples and pigmentation. We also train a deep generative model that can produce high-frequency residuals automatically.
Thanks to the meaningful organization of vector primitives, editing portraits becomes easy and intuitive. In particular, our method supports color transfer, facial expression editing, highlight and shadow editing and automatic retouching.
To quantitatively evaluate the results, we extend the commonly used FLIP metric (which measures color and feature differences between two images) by considering illumination. The new metric, called illumination-sensitive FLIP or IS-FLIP, can effectively capture the salient changes in color transfer results, and is more consistent with human perception than FLIP and other quality measures on portrait images.
We evaluate our method on the FFHQR dataset and show that our method is effective for common portrait editing tasks, such as retouching, light editing, color transfer and expression editing. We will make the code and trained models publicly available.
\end{abstract}


\section{Introduction}
Vector images are ubiquitous in the digital era and essential for massive image production in digital media. Their key advantages include compact representation, resolution independence, being easily editable and capable of representing image content in a simple and visually pleasing manner.
Although vector graphics has a long history, early methods can only represent simple colors (such as piecewise constant or linear). Due to their limited flexibility in representation power in earlier works,
they were only confined in few applications, such as fonts and clip art.
Recent diffusion-based vector primitives, such as diffusion curves (DC)~\citep{orzan2008diffusion} and its bi-Laplacian variant~\citep{finch2011freeform}, significantly improve the expressiveness of vector graphics. The user sketches sparse curves and sets color constraints, and a solver diffuses colors to create an image.

\begin{figure}[htbp]
\centering
{\scriptsize
\includegraphics[width=0.192\columnwidth]{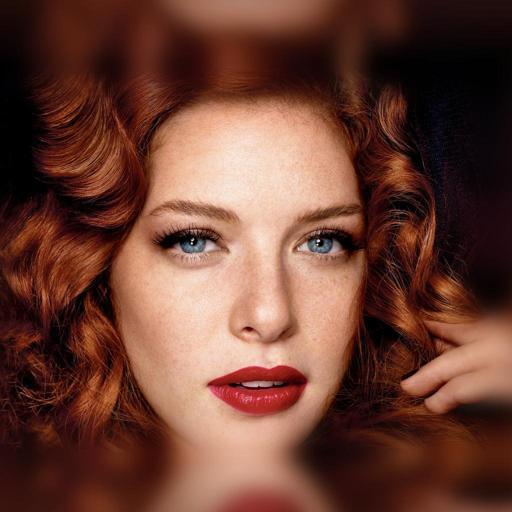}
\includegraphics[width=0.192\columnwidth]{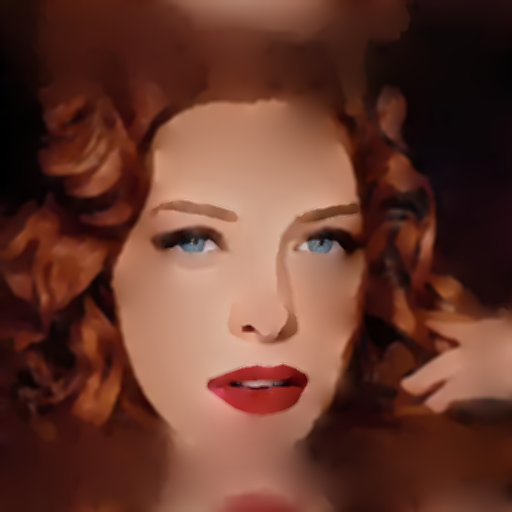}
\includegraphics[width=0.192\columnwidth]{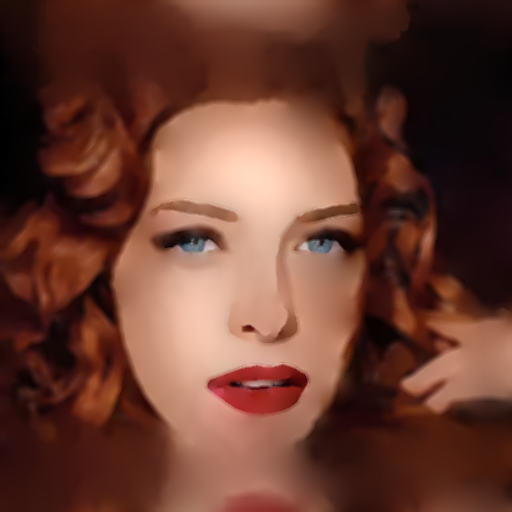}
\includegraphics[width=0.192\columnwidth]{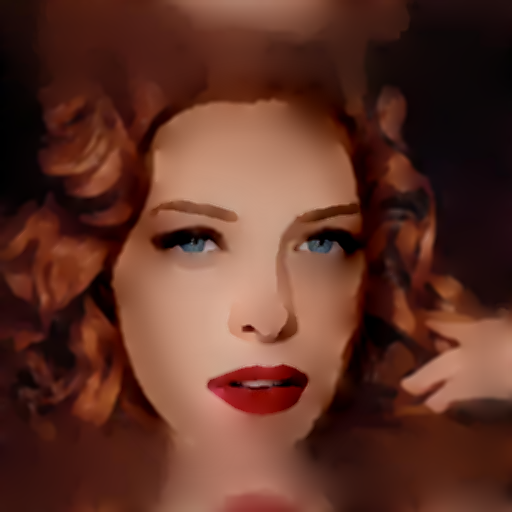}
\includegraphics[width=0.192\columnwidth]{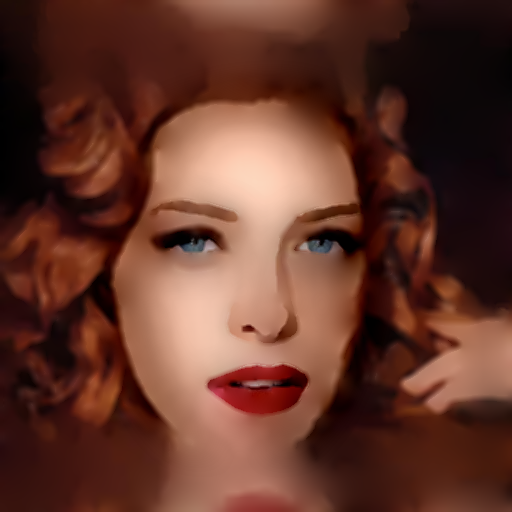}\\
\makebox[0.192\columnwidth]{Original}
\makebox[0.192\columnwidth]{DC}
\makebox[0.192\columnwidth]{DC+hPR}
\makebox[0.192\columnwidth]{{DC+sPR}}
\makebox[0.192\columnwidth]{Base+middle}\\
\includegraphics[width=0.192\columnwidth]{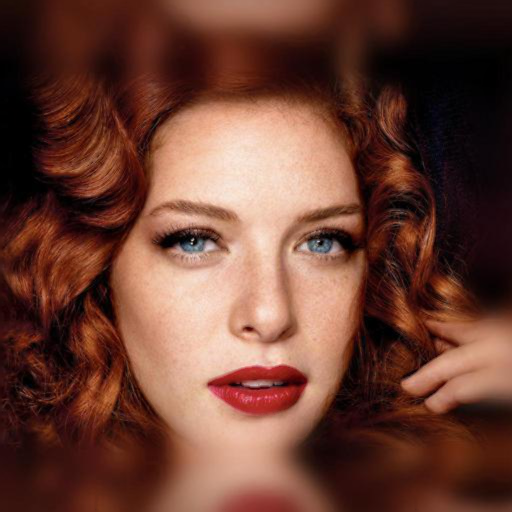}
\includegraphics[width=0.192\columnwidth]{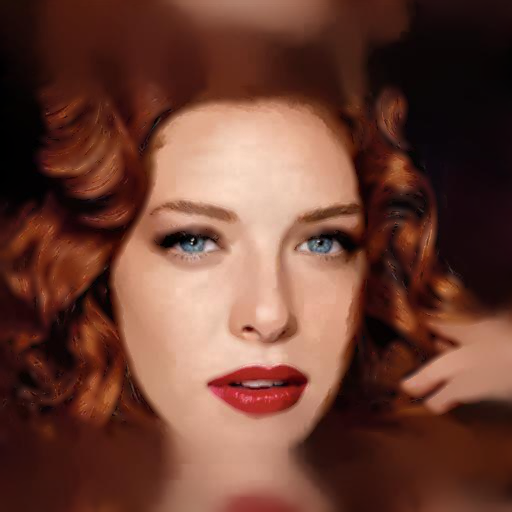}
\includegraphics[width=0.192\columnwidth]{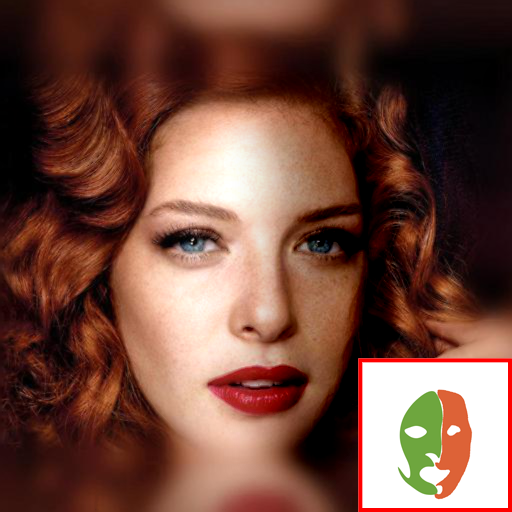}
\includegraphics[width=0.192\columnwidth]{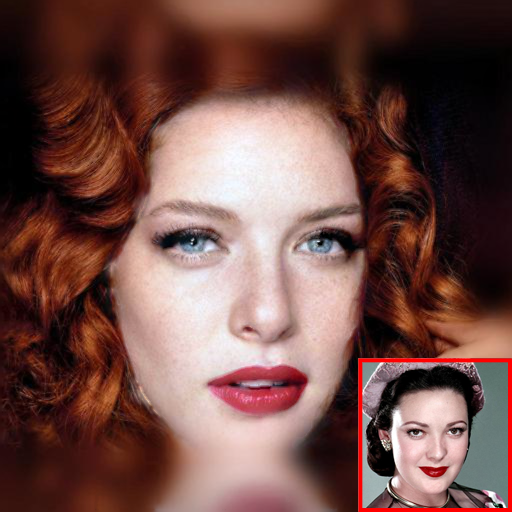}
\includegraphics[width=0.192\columnwidth]{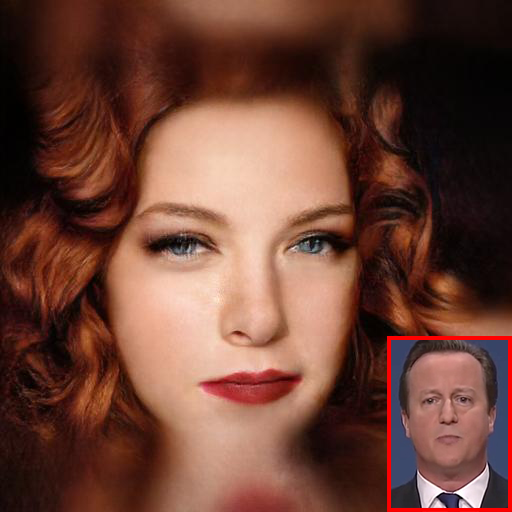}\\
\makebox[0.185\columnwidth]{Base+top}
\makebox[0.192\columnwidth]{Retouching}
\makebox[0.192\columnwidth]{Light editing}
\makebox[0.19\columnwidth]{Color transfer}
\makebox[0.2\columnwidth]{Expression editing}\\
}
\caption{Our method converts a portrait photo into a 3-level vector image, consisting of diffusion curves in the base level, and 2 types of large Poisson regions (hPR and sPR) in the middle level, and 2 types of pixel-sized Poisson regions (fPR and rPR) in the top level. Since our method explicitly separates colors from illumination and facial details in different levels, it is easy to use and enables coarse-to-fine editing in an intuitive manner. For example, it suffices to edit colors and expressions by modifying the boundary colors and geometries of DCs in the base level, to edit illumination by modifying the PRs in the middle level, and to retouch the image by removing fPR in the top level.}
\label{fig:teaser}
\end{figure}

Diffusion curves can produce smooth, non-linear color functions that could mimic photo-realistic images. At present, there are a few algorithms for converting natural photographs into diffusion curves~\cite{orzan2008diffusion,xie2014hierarchical,zhao2017inverse}.
These algorithms are fully automatic and efficient. However, they often produce a large amount of diffusion curves, making post-editing tedious and time-consuming. Moreover, all the DC vectorization methods face a common challenge: diffusion curves are not allowed to intersect each other, since intersecting curves often lead to unpleasant artifacts in color diffusion due to non-compatible or contradicting boundary colors at intersections. As a result, it is difficult to produce {hierarchical} diffusion curve images.

This paper aims at developing a method for converting portrait photos into  {hierarchical} vector graphics. To overcome the aforementioned limitations of DCs, we adopt Poisson vector graphics (PVG)~\cite{hou2018poisson}, which is a mixed representation of diffusion curves and a new type of vector primitives, called Poisson regions (PR).
Unlike diffusion curve images that define the colors by diffusing the user-specified boundary colors from the sparse curves to the entire image domain, Poisson regions, as Laplacian of colors, are the relative values. Using PRs, PVG can explicitly separate colors and tones, which facilities color and tone editing. Furthermore, PVG can tolerate intersection between diffusion curves and Poisson regions~\cite{hou2018poisson}.
The built-in advantages of PVG are favorable to artists, who can quickly sketch vector primitives and easily organize them into various {hierarchical levels} without worrying about the intersection and/or self-intersection. Despite many desired features for PVG \textit{authoring}, PVG \textit{vectorization} is an ill-posed problem, since a pixel of the input image may be assigned to several {hierarchical levels} of different types.

To tackle the challenges, we develop a hybrid method that combines both deep neural networks, which have proven highly successful in 2D image processing, and the traditional optimization techniques. Technically, we first use off-the-shelf tools including face parsing, face retouching, edge detection, and highlight/shadow extraction to automatically convert portrait photos into hierarchical vector primitives.
Then we solve linear least squares to compute the boundary colors of diffusion curves and the Laplacians of Poisson regions.
Our vectorization method converts a portrait photo into a 3-level hierarchical PVG. The base level is a set of sparse diffusion curves, which characterize the salient geometric features and encode the low-frequency colors of skin and hair. The middle level consists of highlight and shadow Poisson regions, both of which are large regions and fully editable.
Finally, the top level contains 2 types of pixel-sized Poisson regions for small facial details (e.g., pimples, pigmentation, etc) and high-frequency residuals.
The base level DCs are always used, whereas the PRs in the upper levels are flexible and can be used arbitrarily, yielding various interesting visual effects and making photo retouching easy and intuitive.
Specifically, we show that the user can edit color and illumination by modifying the boundary colors of diffusion curves and the Laplacians of Poisson regions in the base and middle levels. To support geometry editing, we train a deep generative model that can produce high-frequency residuals automatically when the geometries of diffusion curves are changed.

Thanks to the linearity of the Laplace operator, we can
borrow the blending functions from raster images to vector images. In particular, we define alpha blending, linear dodge and linear burn
to the hierarchically organized PRs and show that they are effective for editing highlights and shadows. We evaluate our method on the CelebAMask-HQ \citep{CelebAMask-HQ} and the Flickr-Faces-HQ-Retouched (FFHQR) \citep{shafaei2021auturetouch} datasets, and demonstrate its effectiveness quantitatively and visually on various portrait editing tasks, such as color transfer, illumination and expression editing. Figure~\ref{fig:teaser} shows an example of the proposed vectorization method.

Our method takes advantages of both vectorization and deep learning in that we make use of mature data-driven techniques in portrait preprocessing and geometric editing, while applying vector primitives in coarse-to-fine attribute editing. As a result, the proposed method is a unified framework for portrait editing and retouching.

In summary, we make the following contributions in the paper.
\begin{itemize}
\item  A new method for converting portrait images into hierarchical vector representation, which supports color transfer, illumination, facial features and expression editing.
\item {Three easy-to-use blending functions, namely alpha blending, linear dodge and linear burn, for the hierarchically organized Poisson regions}.
\item  A novel quantitative evaluation measure for portraits, which can detect the difference of highlights, shadows, colors and features between two images.
\end{itemize}


\section{Related Work}
\noindent\textbf{Intrinsic images.} The intrinsic image decomposition retrieves intrinsic properties of an image and separates it into illumination, reflectance, and specular components.
Some representative recent works include dense conditional random field~\cite{bell2014intrinsic}, near-infrared image aided energy minimization~\cite{cheng2019non}, and a global-local spherical harmonics lighting model
\cite{zhou2019glosh}.
There are also approaches tailored for portrait images. For example, the recent deep learning approaches
\cite{sengupta2018sfsnet,shu2017neural} are able to decompose a face image into shape, reflectance and illuminance, facilitating semantic facial editing.

\noindent\textbf{Vector graphics.}
There are various vector image formulations that can be applied to
portraits. Gradient meshes~\cite{Sun2007}, which are quadrilateral meshes,
represent a raster image by Ferguson patches on planar domain and allow color manipulation in a direct fashion.
However, due to topological constraints, this method cannot deal with regions with holes. \citet{lai2009automatic} proposed a fully automatic method for generating gradient meshes from multiply connected domains.
\citet{chen2019image} proposed a method for encoding editable geometries and fine image details in a hybrid way. Their method is efficient and supports interactive color editing, material replacement and image magnification.
Besides spline functions, subdivision surfaces  \cite{liao2012subdivision, zhou2014representing} are also a popular scheme for image vectorization. These methods represent images by triangle tessellations, thereby are easier to construct than gradient meshes.

Sparse curve based representations, such as diffusion curves~\cite{orzan2008diffusion,finch2011freeform,boye2012vectorial} and Poisson vector graphics~\cite{hou2018poisson}, are more flexible than the mesh-based representations. These methods allow the user to specify constraints only at a set of sparse curves or regions, and then compute color function by solving  partial differential equations, such as Laplace's and Poisson's equations.

\noindent\textbf{DC vectorization.}
\citet{orzan2008diffusion} adopted Canny edges~\cite{Canny} as diffusion curves and solved a linear least square to compute their boundary colors.
\citet{xie2014hierarchical} pointed out in natural images, Canny edges usually do not overlap the Laplacian maxima, hereby fitting diffusion curves in the gradient domain is not accurate. To solve the problem, they proposed extracting and fitting diffusion curves in the Laplacian and bi-Laplacian domains. Their method is fully automatic, efficient and accurate. However, for natural images with rich colors and features, it often produces thousands of diffusion curves, including many short ones. However, with so many primitives, it is difficult for post-editing of both geometries and colors.
\citet{zhao2017inverse} proposed a shape optimization method for extracting smooth DCs from natural images. But it still results in too many primitives, prohibiting the vectorized images from further editing.
\citet{lu2019depth} proposed a vectorization method for RGB-D images, which extracts DCs from both color space and depth map. Their method works only for indoor scene images, due to strict working condition of the current RGB-D cameras.

\noindent\textbf{Portrait image \& video editing}
 has been studied extensively, resulting in a large body of literature. The works that are most relevant to ours are
lighting transfer using optimal mass transport~\citep{shu2017portrait} and deep neural networks~\citep{zhou2019deep,zhang2020neural},
vectorization guided color transfer~\citep{fu2019vectorization}, style transfer through deep image analogy \citep{liao2017visual},
and deep learning based color transfer \cite{afifi2021histogan},
geometry editing~\citep{dekel2018sparse}, expression synthesis \citep{lu2018conditional}, contrast transfer~\citep{shih2014style} and style transfer \citep{sheng2018avatar}.
{The neural model based methods have demonstrated promising performance in many tasks. However training deep neural networks, especially generative adversarial networks (GAN), is prone to many problems \citep{thanh2020catastrophic, bang2021mggan}. A large amount of training data are often needed to achieve good performance. Moreover, each neural model is usually designed to handle a single task, thereby requiring users to maintain multiple models to support a wide range of tasks}. Our method, in contrast, is more flexible in that it allows the users to choose different combinations of the decomposed {levels} to suit their applications.


\section{Hierarchical Poisson Vector Graphics}
\label{sec:overview}

\subsection{Overview}
\label{subsec:objectives}

Poisson vector graphics~\cite{hou2018poisson} is an extension of the popular diffusion curves~\cite{orzan2008diffusion}. A PVG consists of at least one diffusion curve and an arbitrary number of Poisson regions.
Our goal is to convert a portrait image into a hierarchical Poisson vector graphics.
To make the resulting PVG editable, we expect a small number of vector primitives.
Each diffusion curve $\gamma$ is 2-sided curve with a boundary color function $g$ defined on both sides. It can diffuse the boundary colors to produce smooth colors between curves. Therefore, diffusion curves can represent low-frequency colors, serving at the base level.

A Poisson region $\Omega$ is defined by a closed curve and a non-zero Laplacian constraint $f$ associated to the points inside the region. Since Poisson regions can intersect each other as well as diffusion curves, they are ideal for defining additional levels on top of diffusion curves. Specifically, we define 4 types of PR in the middle and top levels to encode \underline{h}ighlights, \underline{s}hadows, \underline{f}ine details and \underline{r}esiduals, which are distinguished by prefixes h, s, f and r, respectively.

To render PVG images, we solve Poisson's equation
\begin{equation}
\label{eq:poisson}
\Delta u(\mathbf{x}) = \left
\{
\begin{array}{ll}
0, & \forall\mathbf{x} \in D\setminus \{\gamma\bigcup \Omega\}, \\
f(\mathbf{x}), & \forall\mathbf{x} \in \Omega,\\
\end{array}
\right.
\end{equation} with Dirichlet boundary condition
\begin{equation}
u(\mathbf{x})=g(\mathbf{x}),~\forall \mathbf{x}\in\gamma,
\end{equation}
where $D$ is the image domain {and $u\in [0,1]$} is the color function.

In hierarchical PVG, the user can control
the geometries and boundary colors $g$ of diffusion curves in the base level, and the geometries and Laplacian constraints of Poisson regions in the middle level. As Poisson regions in the top level are small and may be present in a large amount, direct editing them is not feasible. Instead,  the user can turn each type of PRs on or off, and multiplying them by a global coefficient.

In raster image editing, alpha blending combines a foreground $\mathbf{x}_1$ with a background $\mathbf{x}_2$ to create the appearance of partial or full transparency. The blended image is  $(1-\alpha) \mathbf{x_1}+\alpha \mathbf{x_2}$,
where $\alpha\in [0,1]$ is the opacity.
Thanks to the linearity of the Laplace operator, we can naturally define alpha blending in PVG.
Given two Poisson regions with Laplace constraints $f_1$ and $f_2$ respectively, we define the blended PVG as
\begin{equation}
\label{eq:poisson_alpha}
\Delta u(\mathbf{x}) = \left
\{
\begin{array}{ll}
0,  \forall\mathbf{x} \in D\setminus \{\gamma\bigcup \Omega\}, \\
(1-\alpha) f_1(\mathbf{x})+\alpha f_2(\mathbf{x}),  \forall\mathbf{x} \in \Omega.\\
\end{array}
\right.
\end{equation}

Linear dodge brightens the base color to reflect the blend color by increasing the brightness in each color channel. For raster images, it can be written as $\mathbf{x_1}+\mathbf{x_2}$ for two raster layers $\mathbf{x}_1$ and $\mathbf{x}_2$. In PVG, we define linear dodge as
\begin{equation}
\label{eq:poisson_dodge}
\Delta u(\mathbf{x}) = \left
\{
\begin{array}{ll}
0,  \forall\mathbf{x} \in D\setminus \{\gamma\bigcup \Omega\}, \\
f_1(\mathbf{x})+f_2(\mathbf{x}),  \forall\mathbf{x} \in \Omega.\\
\end{array}
\right.
\end{equation}

Linear burn, which is opposite to linear dodge, darkens the base color to reflect the blend color by decreasing the brightness in each channel. In raster image blending, it can be written as $\mathbf{x_1}+\mathbf{x_2}-1$. We define linear burn in PVG as
\begin{equation}
\label{eq:poisson_burn}
\left
\{
\begin{array}{ll}
\Delta u(\mathbf{x}) = 0,  \forall\mathbf{x} \in D\setminus \{\gamma\bigcup \Omega\}, \\
\Delta \left(u(\mathbf{x}) +1 \right) = f_1(\mathbf{x})+ f_2(\mathbf{x}),  \forall\mathbf{x} \in \Omega.\\
\end{array}
\right.
\end{equation}

As Figures~\ref{fig:three_mode} and~\ref{fig:light.transfer} show, applying linear dodge and linear burn to middle-level PRs allows the user to edit highlights and shadows.

\subsection{Algorithmic Pipeline}

\begin{figure}[!htbp]
\centering
\includegraphics[width=1\columnwidth]{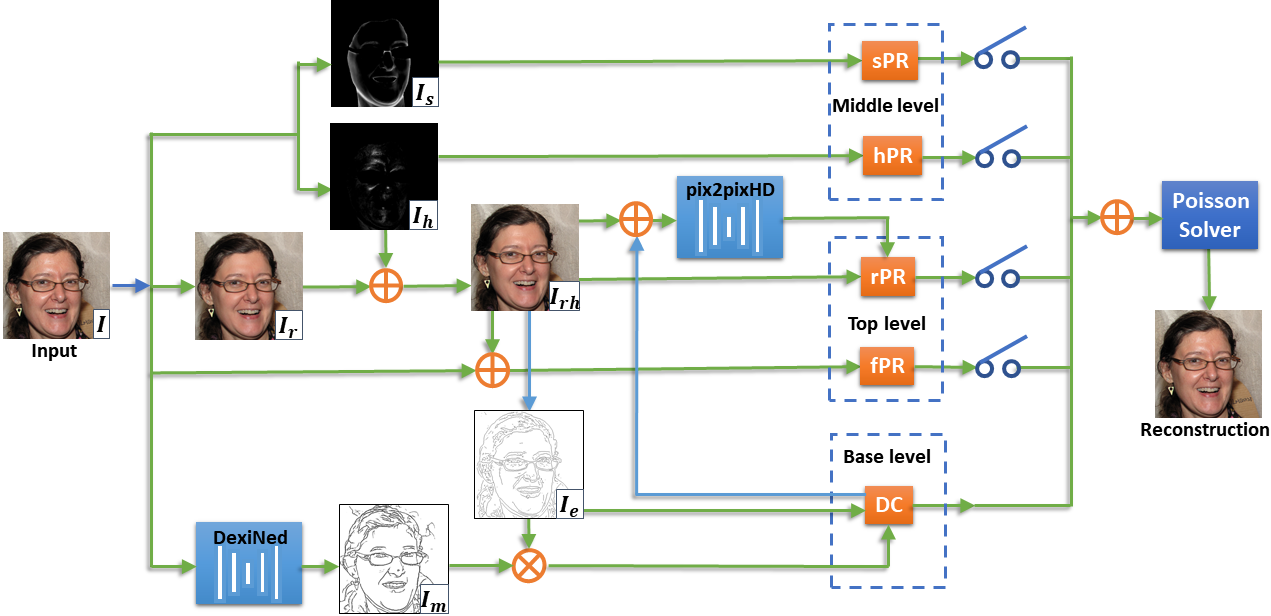}
\caption{The algorithmic pipeline of our {hierarchical vectorization.} See the text for details. 
}
\label{fig:pipeline}
\end{figure}

\begin{figure}[htbp]
\centering
{\scriptsize
\includegraphics[width=0.24\columnwidth]{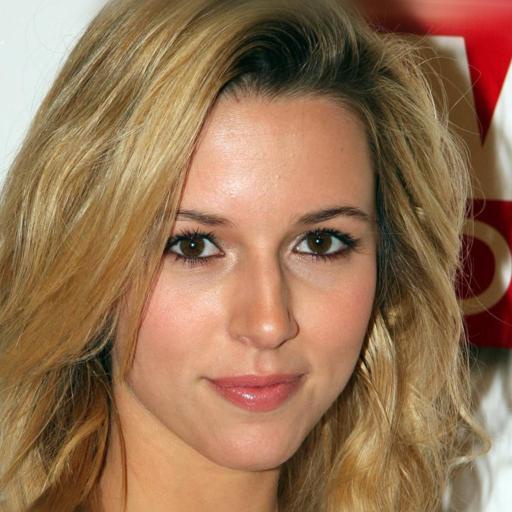}
\includegraphics[width=0.24\columnwidth]{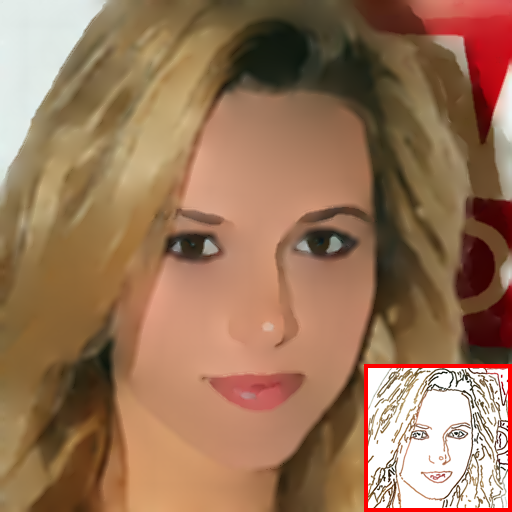}
\includegraphics[width=0.24\columnwidth]{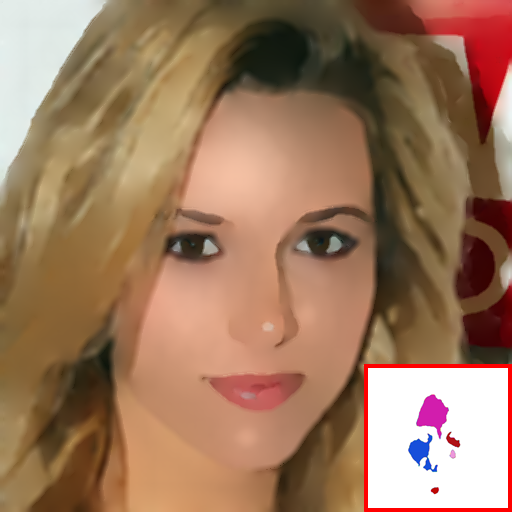}
\includegraphics[width=0.24\columnwidth]{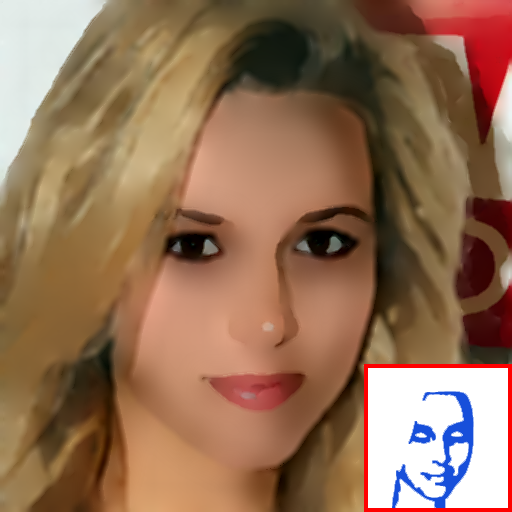}\\
\makebox[0.24\columnwidth]{(a) Original}
\makebox[0.24\columnwidth]{(b) DC}
\makebox[0.24\columnwidth]{(c) DC+hPR}
\makebox[0.24\columnwidth]{{(d) DC+sPR}}\\
\vspace{1pt}
\includegraphics[width=0.24\columnwidth]{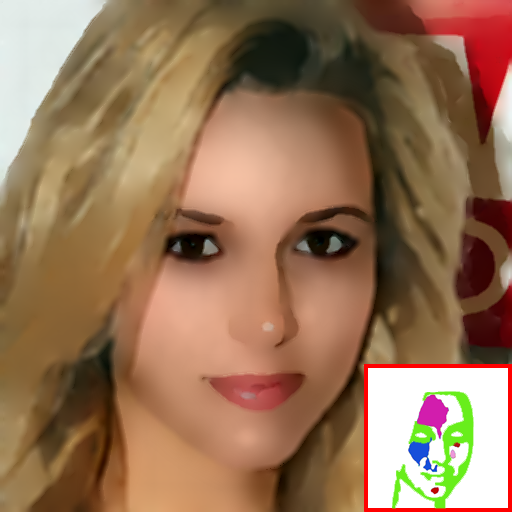}
\includegraphics[width=0.24\columnwidth]{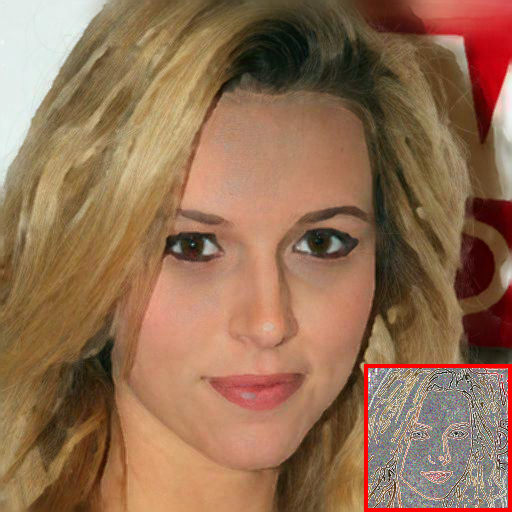}
\includegraphics[width=0.24\columnwidth]{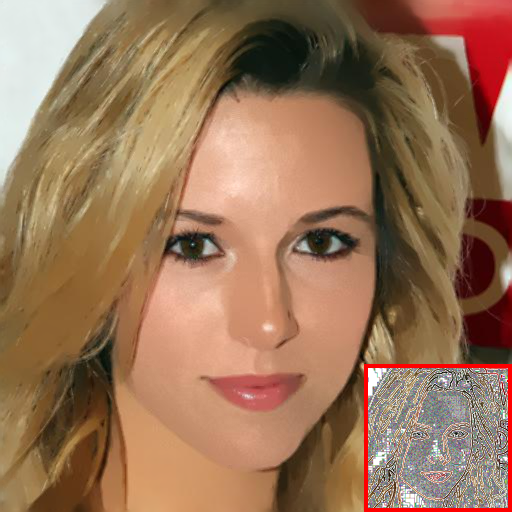}
\includegraphics[width=0.24\columnwidth]{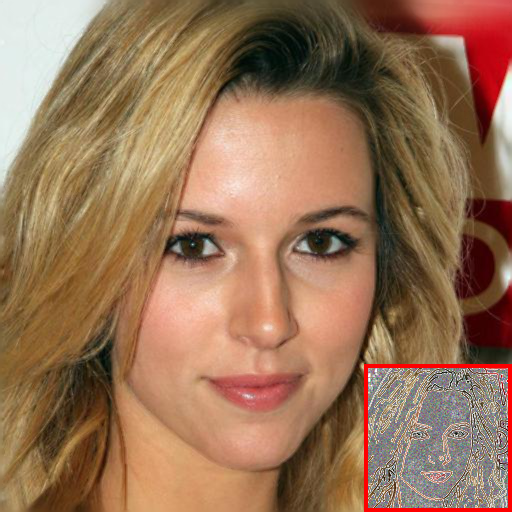}\\
\makebox[0.24\columnwidth]{{(e) DC+middle}}
\makebox[0.24\columnwidth]{(f) DC+fPR}
\makebox[0.24\columnwidth]{(g) DC+rPR}
\makebox[0.24\columnwidth]{{(h) Base+top}}\\}
\caption{Illustration of hierarchical PVG. The vector primitives are shown in the small insets. There are 358 DCs (5.2\% px) and 7 PRs in this example. Among the diffusion curves, only less than 30 of them represent the hair and face contours as well as salient facial features, such as month, nose and eyes, which are used in geometry editing. Our method can produce photo-realistic clipart effect with the base and middle levels in (e), and automatic retouching effect with the base level and rPR in (g)}.

\label{fig:layered_pvg}
\end{figure}

Figure~\ref{fig:pipeline} shows the algorithmic pipeline of our {hierarchical} vectorization method. We give a quick overview of the components in this subsection, and document the technical details in Sections~\ref{sec:dc} and~\ref{sec:pr}.
{Table~\ref{table:application} summarizes the typical usages and their corresponding operations of hierarchical PVG for portraits.} {Figure \ref{fig:layered_pvg} shows an example of our hierarchical PVG.}

\begin{table*}[!htbp]
\centering
\begin{tabularx}{1.0\textwidth}{|m{0.06\textwidth}|m{0.17\textwidth}|m{0.11\textwidth}|m{0.177\textwidth}|X|}
\hline
\textbf{~Level} & \textbf{Information} & \textbf{\small{Vector} \newline \small{Primitives}} & \textbf{Usage} & \textbf{Operation}\\
\hline
\multirow{2}{*}{~Base} & Salient \newline geometric & \multirow{2}{*}{~~~DCs} & Color transfer \newline \& editing & Changing boundary colors\\
\cline{4-5}
 & features \& low-frequency colors & &Geometry editing &  Adding/deleting curves \& changing curves' geometry \\
\hline
\multirow{2}{*}{Middle} & \multirow{2}{*}{Illumination} &  ~~~hPRs & Editing \newline highlights& Adding/editing PRs with  positive Laplacians\\
\cline{3-5}
& & \vspace{1pt}~~~sPRs &  Editing shadows &  Adding/editing PRs with negative Laplacians\\
\hline
\multirow{3}{*}{} &  & \vspace{1pt}  ~~~fPRs & Restoring \newline fine details & Multiplying the Laplacians of all fPRs by a global coefficient\\
\cline{3-5}
~Top & Fine details \& residuals & \multirow{2}{*}{~~~rPRs} & Face retouching & Setting fPR=0 and rendering only rPRs and DCs\\
\cline{4-5}
\cline{4-5}
&&& Geometry editing  & Generating rPRs using the deep generative model with the modified DCs as input \\
\hline
\end{tabularx}
\vspace{3pt}
\caption{{Hierarchical decomposition of portrait images.} }
\label{table:application}
\end{table*}

\begin{figure}[!htbp]
    \centering
    \includegraphics[width=0.32\columnwidth]{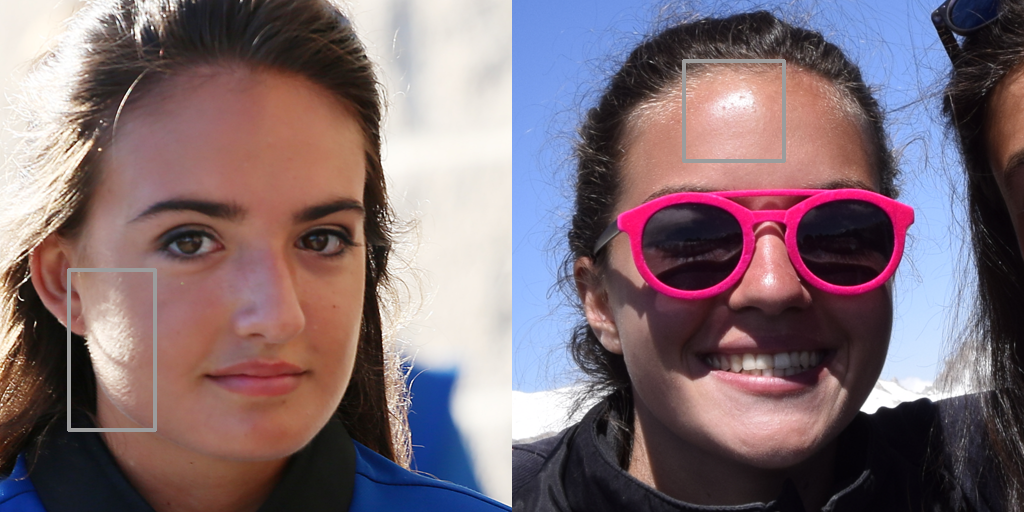}
    \includegraphics[width=0.32\columnwidth]{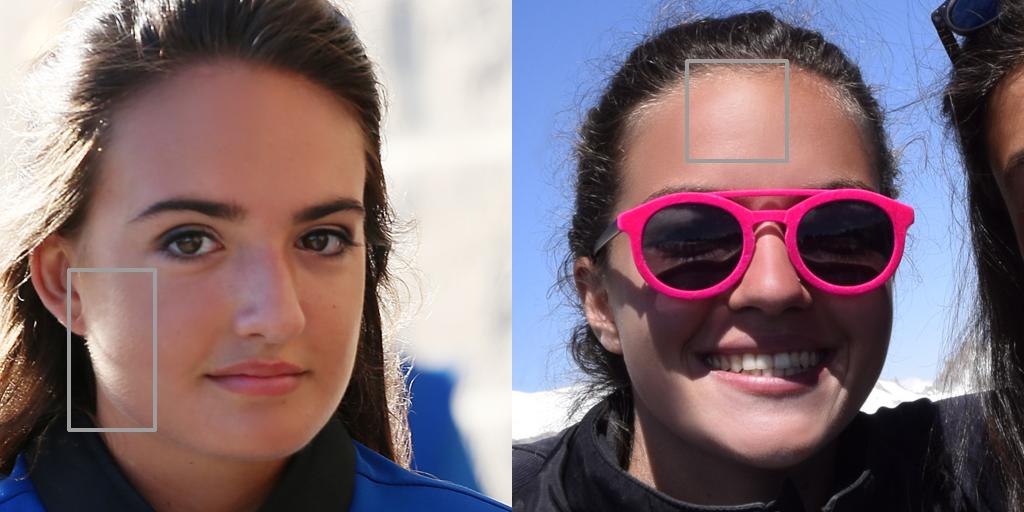}
    \includegraphics[width=0.32\columnwidth]{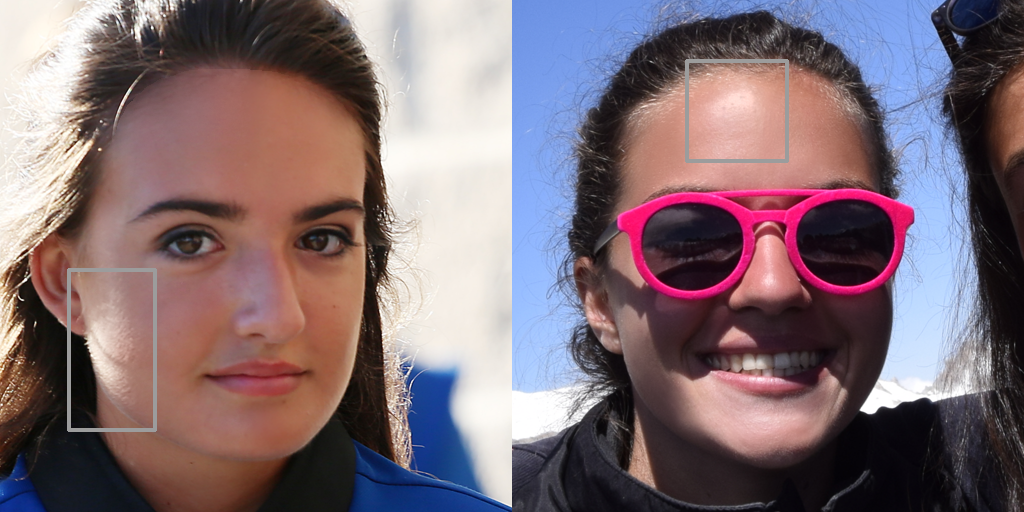}
    {\scriptsize
    \makebox[0.32\columnwidth]{(a) $I$}
    \makebox[0.32\columnwidth]{(b) $I_r$}
    \makebox[0.32\columnwidth]{(c) $I_{rh}$}}
    \caption{{Preprocessing. Given an input image $I$, we compute a retouched image $I_r$ and a highlight-compensated retouched image $I_{rh}$. We highlight the differences in greyed boxes.  }}
    \label{fig:preprocessing}
\end{figure}

\noindent\textbf{Preprocessing.} {Let $I$ be the input portrait image. Denote by $I_r$ the retouched image of $I$. Each image in the FFHQR dataset \citep{shafaei2021auturetouch} is already associated with a retouched image that was created by professional artists. For general portrait images, one can apply the auto face retouching algorithm~\citep{shafaei2021auturetouch} to even out the skin and remove imperfections and oily glare.
We also compute a highlight image $I_h$ by applying the highlight removal algorithm~\citep{shen2013real}. We then compute a highlight-compensated retouched image $I_{rh}$ by adding highlights to $I_r$. See Figure~\ref{fig:preprocessing}.}

\noindent\textbf{Extracting diffusion curves (DCs).} {We adopt a two-step method for computing DCs.} {First, we apply the probability edge algorithm~\cite{leordeanu2012efficient} to extract strong edges $I_e$ in the retouched image $I_{rh}$ and use the colors on the edges to define the boundary colors $g$ of DCs. Those DCs, which encode both skin colors and illuminations, are used in geometry editing (e.g., changing expressions). }
{In color transfer and light editing, we require skin colors are separated from  illumination. So we modify the DexiNed neural network~\cite{poma2020dense} to compute salient facial feature curves as edge mask $I_m$. Applying $I_m$ to the probability edge induced DCs can effectively reduce interference of light to skin colors.}

 \noindent\textbf{Extracting highlight{ \& shadow Poisson regions (hPRs \& sPRs)}.} We apply the highlight removal algorithm~\citep{shen2013real} to extract specular highlights $I_h$ from the original image $I$. Then we soften the boundaries of the highlights using median filter.  {Applying \citep{shen2013real} to the inverse image $(1-I)$ yields shadows $I_s$.
 We assign each hPR (resp. sPR) a positive (resp. negative) constant Laplacian. }

\noindent\textbf{Extracting fine details Poisson regions (fPRs).} {The fine details are small-scale features, such as pimples and pigmentation.
We define fPRs as the Laplacian of the difference between the input image $I$ and the highlight-compensated retouched image $I_{hr}$, i.e., $\bigtriangleup (I-I_{hr})$.}

\noindent\textbf{Extracting residual Poisson regions (rPRs).}
{Residual PRs are pixel-sized PRs defined as the Laplacian of the highlight compensated retouched image $I_{hr}$. It is worth noting that rPRs are not meant for direct editing. Instead, they are to preserve the photorealism  of the results.
For applications without changing the geometry of DCs (e.g., color transfer, light editing), we simply keep the extracted rPRs unchanged. For other applications in which the DC geometries are changed (e.g., expression editing), we use a deep generative model, which takes the modified DCs as input, to generate new rPRs. Due to this reason, we call them ``residual'' Poisson regions.
}

\subsection{Properties}
\label{subsec:properties}

Our {3-level, hierarchical} PVG has several properties that are favorable to face image retouching and editing.

First, the vector primitives are organized in a hierarchical structure: contours and salient facial features are in the base level, and fine details (such as pimples and uneven skin color) and high-frequency residuals are in the top level. The middle level models highlights and shadows, and does not contain geometry information. The user can manipulate the base level primitives (i.e., sparse diffusion curves) to modify the face geometry. Since there are only a few PRs in the middle level, the user can also directly modify their geometries and Laplacians to edit illumination. For the PRs in the top level, the user can apply filters and/or use brushes to modify them.

Second, our portrait PVG explicitly separates colors and tones, which follows the basic drawing principle adopted by artists. There are only a small number of DCs in the base level {representing hair and face contours and salient facial features, such as eyes, month and nose.} Each DC is explicitly assigned two boundary colors on both sides.
To change face and hair color, it suffices for the user to modify the boundary color of a few relevant DCs. If the input image is vectorized using DCs only~\cite{orzan2008diffusion,xie2014hierarchical,zhao2017inverse}, editing colors is tedious and time-consuming, since the user has to work on a large number of DCs.
Besides, our method supports automatic color transfer from a reference image to the input image, where the two images may have very different facial geometries, colors and head poses. {The 2 types of PRs in the middle level enable the user to directly edit illumination. Each PR is assigned with a constant integral Laplacian in Eqn. (\ref{eqn:PR-control}). The user can modify both the geometries and the integral Laplacians of PRs to edit illumination.}

Third, unlike diffusion curves that are strictly intersection-free, Poisson regions are flexible in that they allow intersection and self-intersections, and can be freely placed and blended. This feature is particularly desired in illumination editing that involves highlight and shadow PRs in the middle level (see Figure \ref{fig:light.transfer}. Also, combining the middle-level PRs and the base-level DCs yields visually pleasing clipart effect (see Figure~\ref{fig:layered_pvg}(e)).

\begin{figure}[htbp]
\centering
{\scriptsize
\includegraphics[width=0.192\columnwidth]{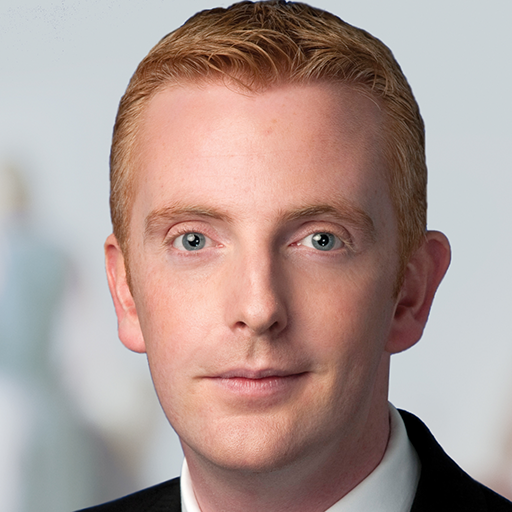}
\includegraphics[width=0.192\columnwidth]{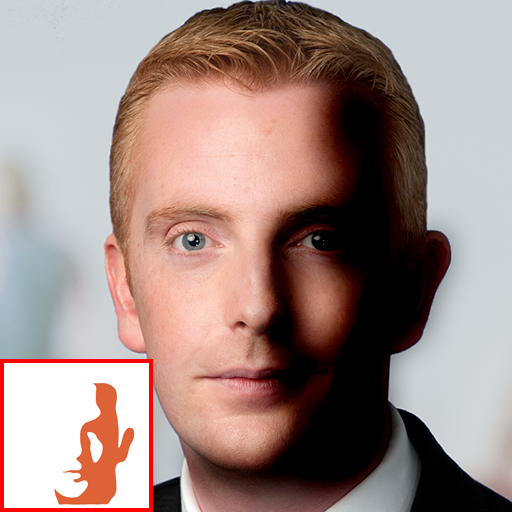}
\includegraphics[width=0.192\columnwidth]{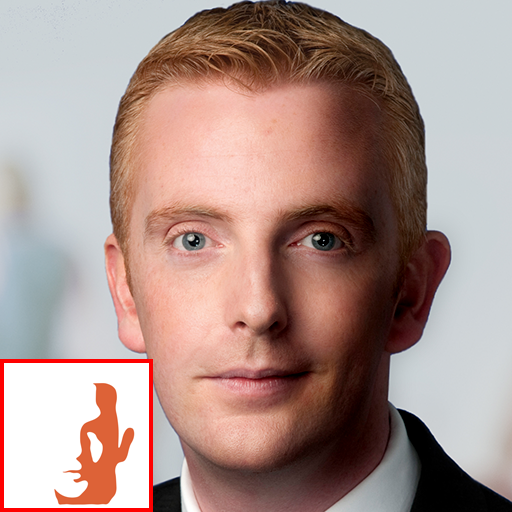}
\includegraphics[width=0.192\columnwidth]{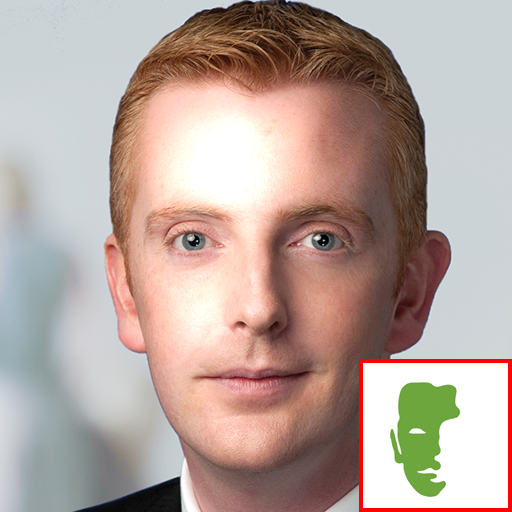}
\includegraphics[width=0.192\columnwidth]{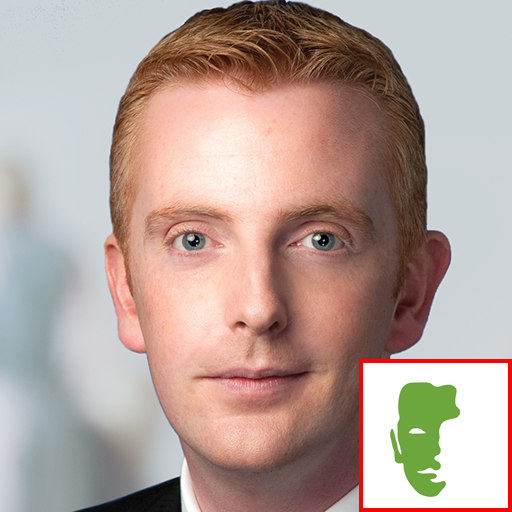}\\
\makebox[0.192\columnwidth]{(a) Original}
\makebox[0.192\columnwidth]{(b) Burn: 100\%}
\makebox[0.192\columnwidth]{(c) Burn: 40\%}
\makebox[0.192\columnwidth]{(d) Dodge: 100\%}
\makebox[0.192\columnwidth]{(e) Dodge: 40\%}\\
}
\caption{{Applying the linear blending functions to edit highlights and shadows. (b)-(c): We add an sPR with integral Laplacian $(-20,-20,-20)$ to create shadows on the right cheek. (d)-(e): We add an hPR with integral Laplacian $(20,20,20)$ to create highlights. By tuning the opacity percentage, we obtain visually pleasing results in (c) and (e).}}
\label{fig:three_mode}
\end{figure}

Fourth, thanks to the linearity of the Laplace operator, we introduce three linear blending modes, i.e., alpha blending, linear dodge and linear burn, which allow the user to edit the highlight and shadow PRs in the middle level in an easy and intuitive manner. See Figure \ref{fig:three_mode} for an example.

Fifth, our method takes advantages of editability of vector primitives.
All the vector primitives in the base and middle levels are free to edit by the user. However, the high-frequency signals in the top level are not meant for direct editing. They are used to preserve photorealism of the results. For applications that do not involve changing the geometries of diffusion curves (e.g., color transfer and illumination editing), we simply keep the top-level PR unchanged. For other applications (such as expression editing) in which the user changes the geometries of diffusion curves, we adopt a deep generative model to automatically yield the primitives of residual PRs. As a result, the user can freely modify the base level primitives without worrying about the primitives of upper levels. See Figure \ref{fig:dc_ai} for illustration.


\section{DC Extraction}
\label{sec:dc}

Diffusion curves are a set of sparse curves with boundary colors assigned on both sides. Diffusion curve images are rendered by diffusing the boundary colors from the curves to the entire image domain. The existing DC extraction methods rely on edge and feature detector. Traditional edge detector methods, such as  \citep{canny1986computational,leordeanu2012efficient}, consider only local color changes and often yield short and incomplete curves.
It is known that diffusion curve images are highly sensitive to the positions of curves. For example, if a region's boundary, which is supposed to be a closed curve, turns out to be incomplete, the colors inside the region will be diffused out of it, causing color leaking artifact.

\begin{figure}[htbp]
\centering
\includegraphics[width=0.185\columnwidth]{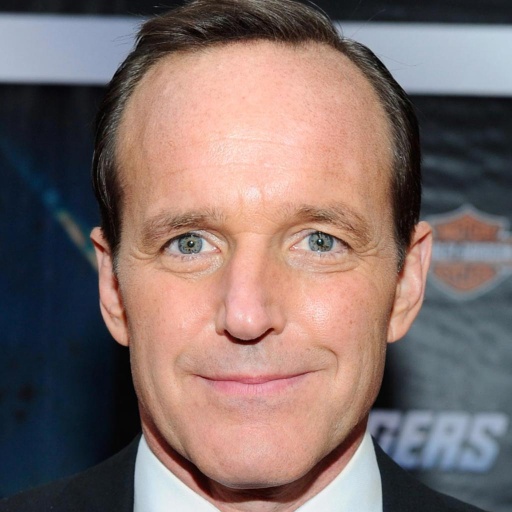}
\includegraphics[width=0.185\columnwidth]{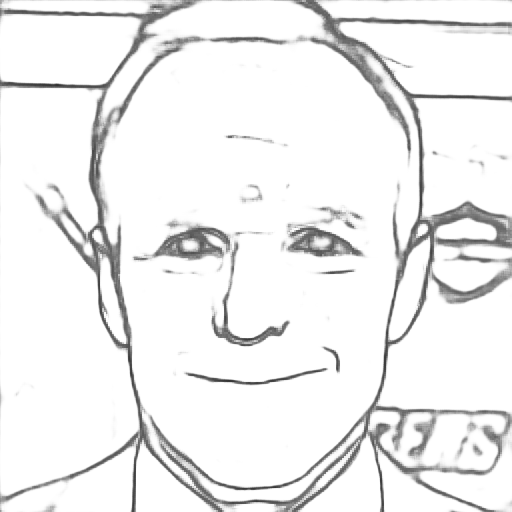}
\includegraphics[width=0.185\columnwidth]{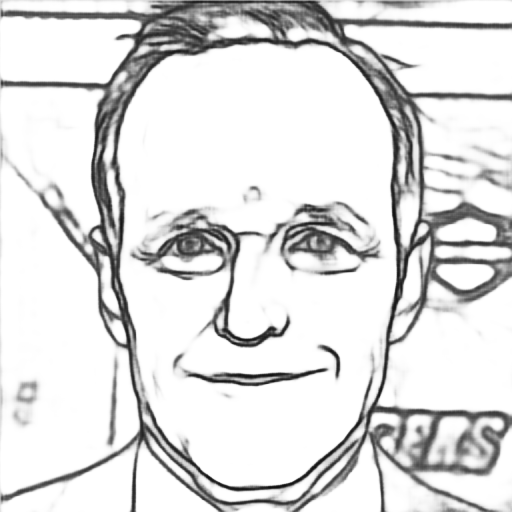}
\includegraphics[width=0.185\columnwidth]{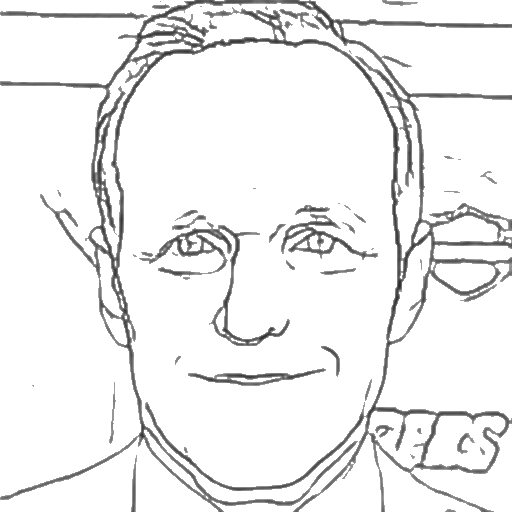}
\includegraphics[width=0.185\columnwidth]{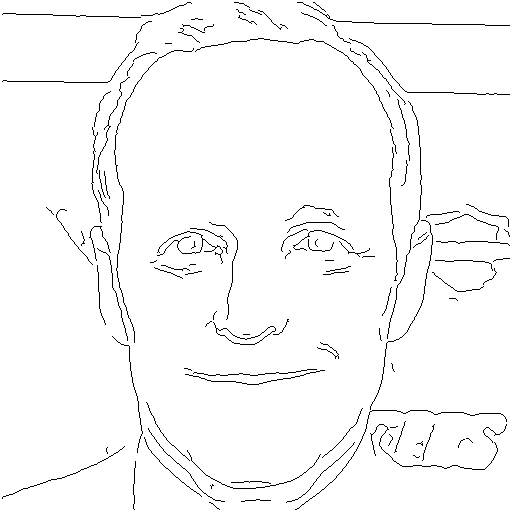}\\
{\scriptsize
\makebox[0.185\columnwidth]{(a)}
\makebox[0.185\columnwidth]{(b)}
\makebox[0.185\columnwidth]{(c)}
\makebox[0.185\columnwidth]{(d)}
\makebox[0.185\columnwidth]{(e)}}
\caption{An example of DC mask. (a) is the input image {$I$}; (b) is the output of the original DexiNed model; (c) is the output of our retrained model; (d) is the DC mask {$I_m$} by smoothing (c); and (e) is the extracted features $I_e\cap I_m$ after applying the mask $I_m$ to the probability edges.}
\label{fig:dc_mask_example}
\end{figure}

\begin{figure}[htbp]
\centering
\includegraphics[width=0.24\columnwidth]{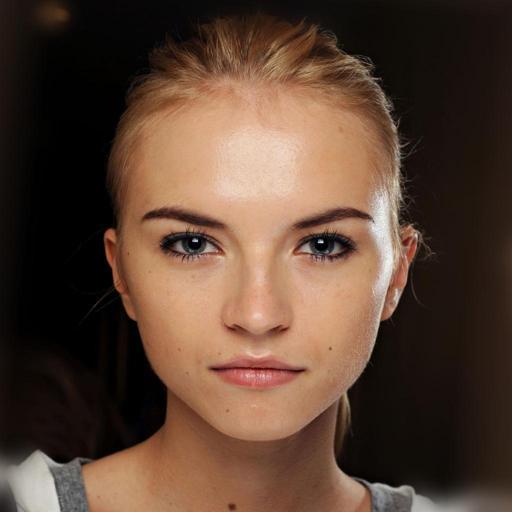}
\includegraphics[width=0.24\columnwidth]{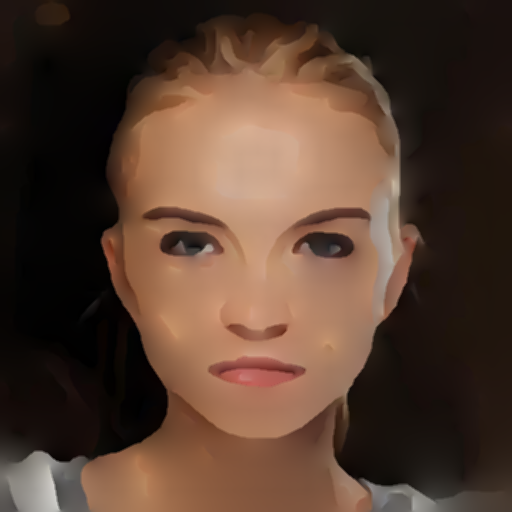}
\includegraphics[width=0.24\columnwidth]{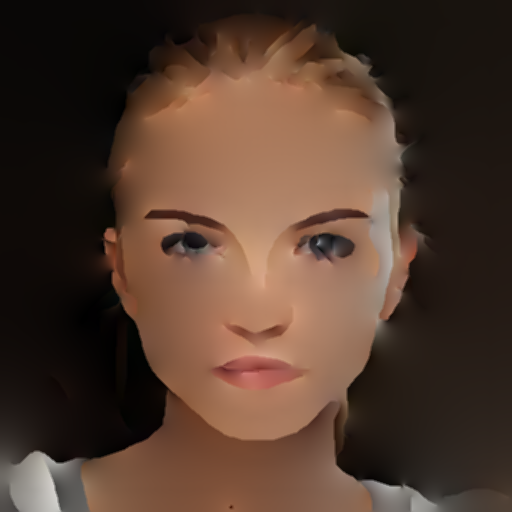}
\includegraphics[width=0.24\columnwidth]{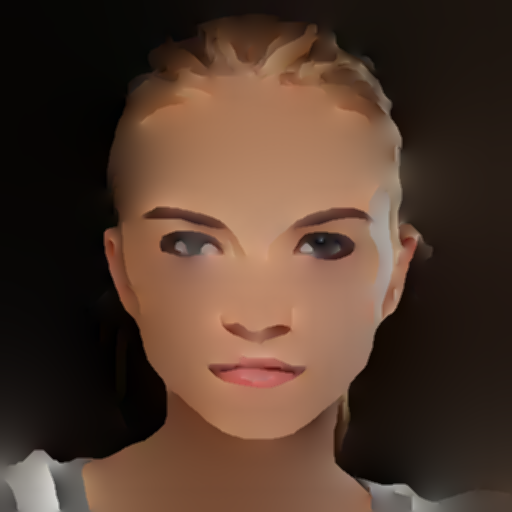}\\
\includegraphics[width=0.24\columnwidth]{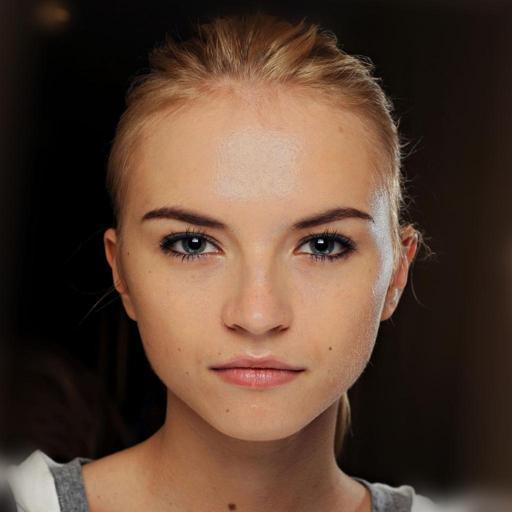}
\includegraphics[width=0.24\columnwidth]{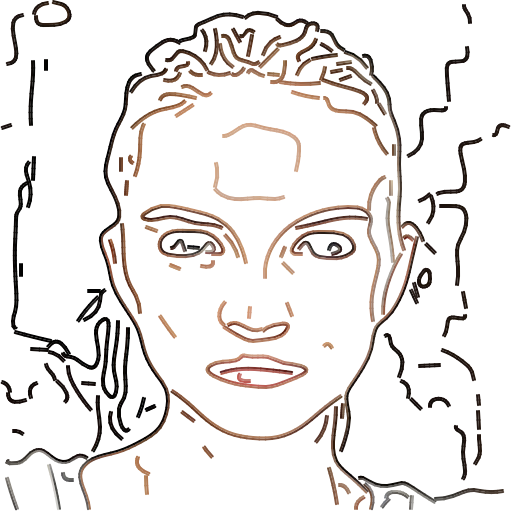}
\includegraphics[width=0.24\columnwidth]{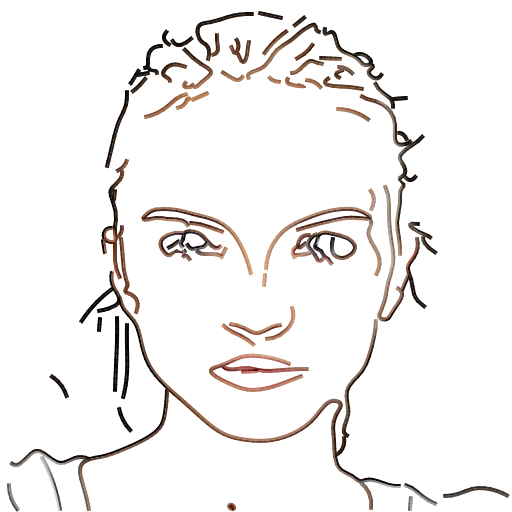}
\includegraphics[width=0.24\columnwidth]{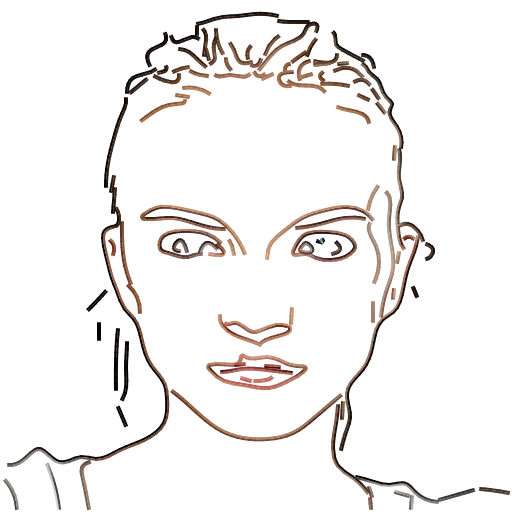}\\
{\scriptsize
\makebox[0.24\columnwidth]{(a)}
{
\makebox[0.24\columnwidth]{(b) 2.7\% px}
\makebox[0.24\columnwidth]{(c) 1.7\% px}
 \makebox[0.24\columnwidth]{(d) 1.8\% px}\\}}
\caption{DC extraction. {The percentages are the ratio of the number of pixels in DCs to the number of pixels in the image.} (a) The input image {$I$} (top) and the highlight-free image {$I-I_h$} (bottom). (b) The DCs $I_e$ generated by the probability edge method~\citep{leordeanu2012efficient} {often mix skin colors and illuminations}. (c) The DCs extracted by the DexiNed model can reduce the mixture of skin color and illumination. However, their locations are not as accurate as those of probability edges.
(d) {We compute an edge mask $I_m$ using  DexiNed and then apply it to the probability edges $I_e$. Experiments show that $I_e\cap I_m$, we can obtain high-quality DCs with reducing the interference of facial lighting, which can be used for facial color transfer and light editing.} }
\label{fig:dc}
\end{figure}

To address such issue, {we adopt the following strategy. First, we apply a classic edge detection algorithm~\citep{leordeanu2012efficient} to obtain strong feature lines, which are located at points with high color gradients. These lines are accurate in terms of positions, however, they often mix skin colors and illumination, which are supposed to be separated.}
{Second, we compute a highlight-compensated retouched image $I_{rh}$ by minimizing the following objective function
\begin{equation}
\label{eq:residual}
I_{rh}=\arg\min_{\varepsilon}(I-I_r-\varepsilon \times I_h)\times I_h+I_r,
\end{equation}
where $I$ is the input image, $I_h$ the highlight image and $I_r$ the retouched image.
Third, we adopt a modified DexiNed network (to be elaborated in the next paragraph) to extract edges in $I_{rh}$ and smooth them to produce an edge mask $I_m$. Fourth, applying the mask to the probability edges, we obtain diffusion curves which characterize the main geometric features and are less sensitive to illumination. Finally, we assign each pixel on either side of the extracted diffusion curve a color which corresponds to a pixel color of the largest gradient difference. See Figure~\ref{fig:dc_mask_example} for an example of edge mask and Figure~\ref{fig:dc} for the results of extracted DCs with and without the edge mask.} {In contrast with the previous methods~\cite{canny1986computational,leordeanu2012efficient}, which rely only on local color changes, our method considers both colors and facial features since the DC mask extraction model is trained on annotated portrait images.
}

\noindent\textbf{{Portrait-tailored DexiNed.}} {DexiNed \citep{poma2020dense} is a deep neural network based edge detector, which was originally trained on the BIPED dataset with $250$ annotated outdoor images.}
We manually annotate $27$ portrait images and combine them with the BIPED dataset, and re-train DexiNed. We also adopt the data augmentation script \footnote{\url{https://github.com/xavysp/MBIPED.git}} provided by BIPED for data augmentation. In addition, we apply self-training \citep{xie2020self, zoph2020rethinking}, where we use the model trained on manually annotated data to annotate $500$ randomly sampled portrait images, and then combine them with the hand-annotated data to train the final model. Given any portrait image, our re-trained model is able to generate the DC mask by assigning a probability to each pixel. We develop a smooth algorithm to further process the DC mask: 1) apply Gaussian filter to the DC mask and then set the pixels below a given threshold ({0.35 in our implementation}) to 0; 2) move a $5 \times 5$ sliding window though the DC mask with stride size 1, and filter out the bottom $20\%$ values in each window; 3) repeat step 1) to generate the final DC mask.

\section{{Hierarchical} PR Extraction}
\label{sec:pr}

In the hierarchical PVG, Poisson regions are in the middle and the top levels.
The middle level characterizes highlights and shadows, and the top level encodes residuals and fine details. There are two types of middle-level PRs, namely, hPRs and sPRs, which are large and editable regions. It is worth noting that a typical portrait PVG has only a small number of hPRs and sPRs (e.g., no more than ten) so that the user can edit them directly. The top-level fPRs and rPRs are small, pixel-sized regions, which encode high-frequency signals such as pimples, pigmentation and residuals. Note that the top-level PRs are not meant for direct editing due to their large numbers and small sizes.

\subsection{Highlights \& Shadows}
To facilitate highlight and shadow editing, we define 2 types of editable PRs in the middle level.
With the highlight removal algorithm~\citep{shen2013real}, {we can extract highlights $I_h$ and shadows $I_s$ from the original image $I$ and its inverse image $(1-I)$. }
After that, we apply the median filter to the highlight image $I_h$ {(resp. shadow image $I_s$)} and then extract boundaries to determine non-trivial hPRs {(resp. sPRs)} with relative smooth boundaries.

Unlike DCs, PRs are 2-dimensional objects.
Let $\Omega_i$ be a Poisson region. Following~\cite{hou2018poisson}, we require that the area integral of the Laplacian $\int_{\Omega_i} f(\mathbf{x})\mathrm{d}\mathbf{x}=0$ vanishes. The zero-sum requirement is a necessary condition for local shading control.
Denote by $\Omega_i'\subset\Omega_i$ the region such that the boundary $\partial\Omega_i$ is shrunk $\delta$ unit distance inwards along the boundary normals. In our implementation, we set $\delta=$ 5\% of the diagonal of $\Omega_i$. We control $\Omega_i'$ by specifying the integral Laplacian $P_i$ for $\Omega_i'$ as
\begin{equation}
\label{eqn:PR-control}
P_i=\int_{\Omega_i'} f(\mathbf{x})\mathrm{d}\mathbf{x}.
\end{equation}
Then we assign the Laplacian for each point of $\Omega_i\setminus\Omega_i'$ as
$$
\frac{-P_i}{\int_{\Omega_i\setminus\Omega_i'}\mathrm{d}\mathbf{x}}
$$
to satisfy the zero-sum condition, where $P_i$ is given by the mean Laplacian values of $\Omega_i'$.
In addition, the user can edit highlights and shadows by specifying different integral Laplacians $P_i$ for $\Omega_i'$ or directly modifying its geometry to produce various illumination effects.

\subsection{{Fine Details \& Residuals}}
{Fine details and residuals are the two types of PRs in the top level. Both are pixel-sized and usually outnumber the primitives in the base and middle levels significantly. Hence, these PRs are not meant for direct editing and they are also different in terms of usage and operations.}

{We define residual Poisson regions as the Laplacian of the highlight compensated retouched image $I_{rh}$, $f_r=\Delta I_{rh}$.
Then we define fine details Poisson regions as $\Delta I-f_r$, which contains facial details such as pimples and pigmentation.}

\begin{figure}[!htbp]
\centering
{\scriptsize
\includegraphics[width=0.158\columnwidth]{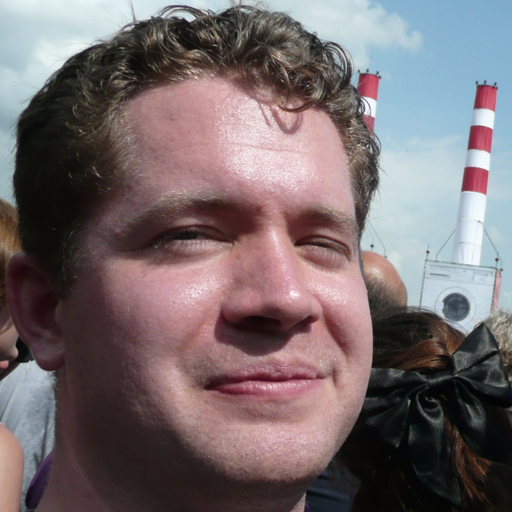}
\includegraphics[width=0.158\columnwidth]{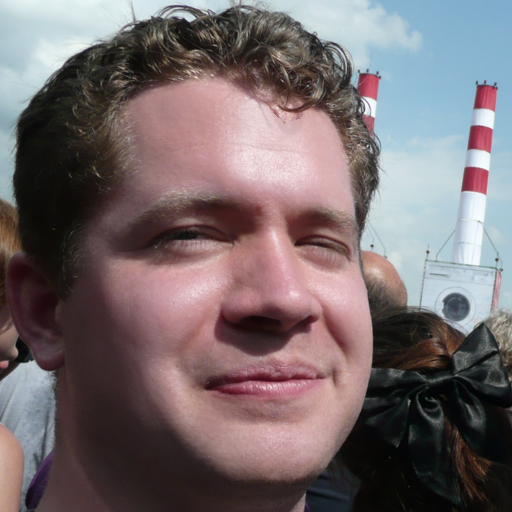}
\includegraphics[width=0.158\columnwidth]{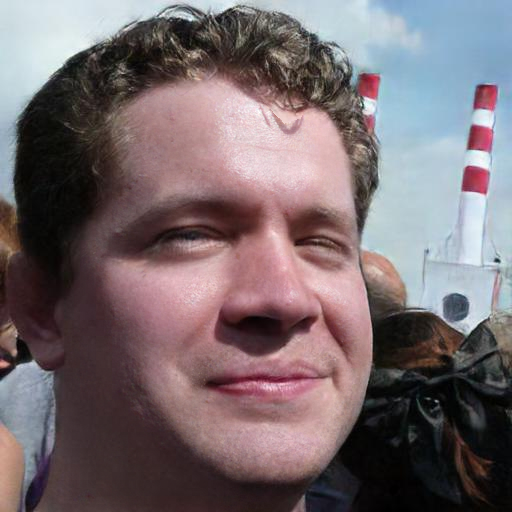}
\includegraphics[width=0.158\columnwidth]{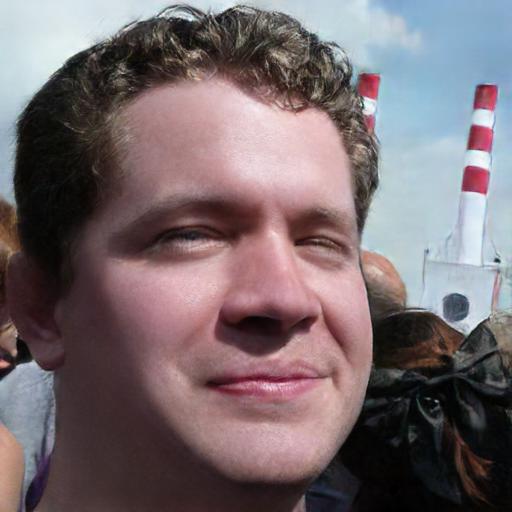}
\includegraphics[width=0.158\columnwidth]{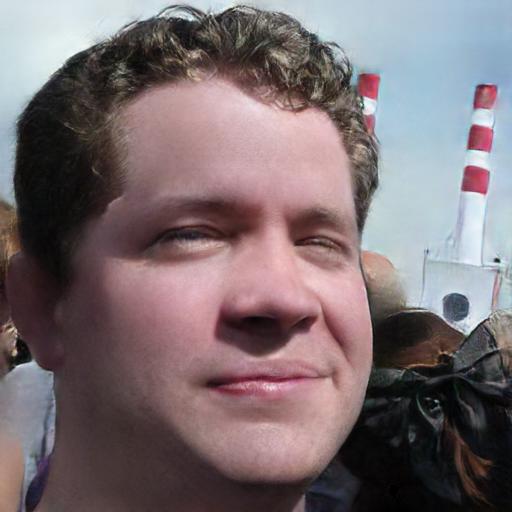}
\includegraphics[width=0.158\columnwidth]{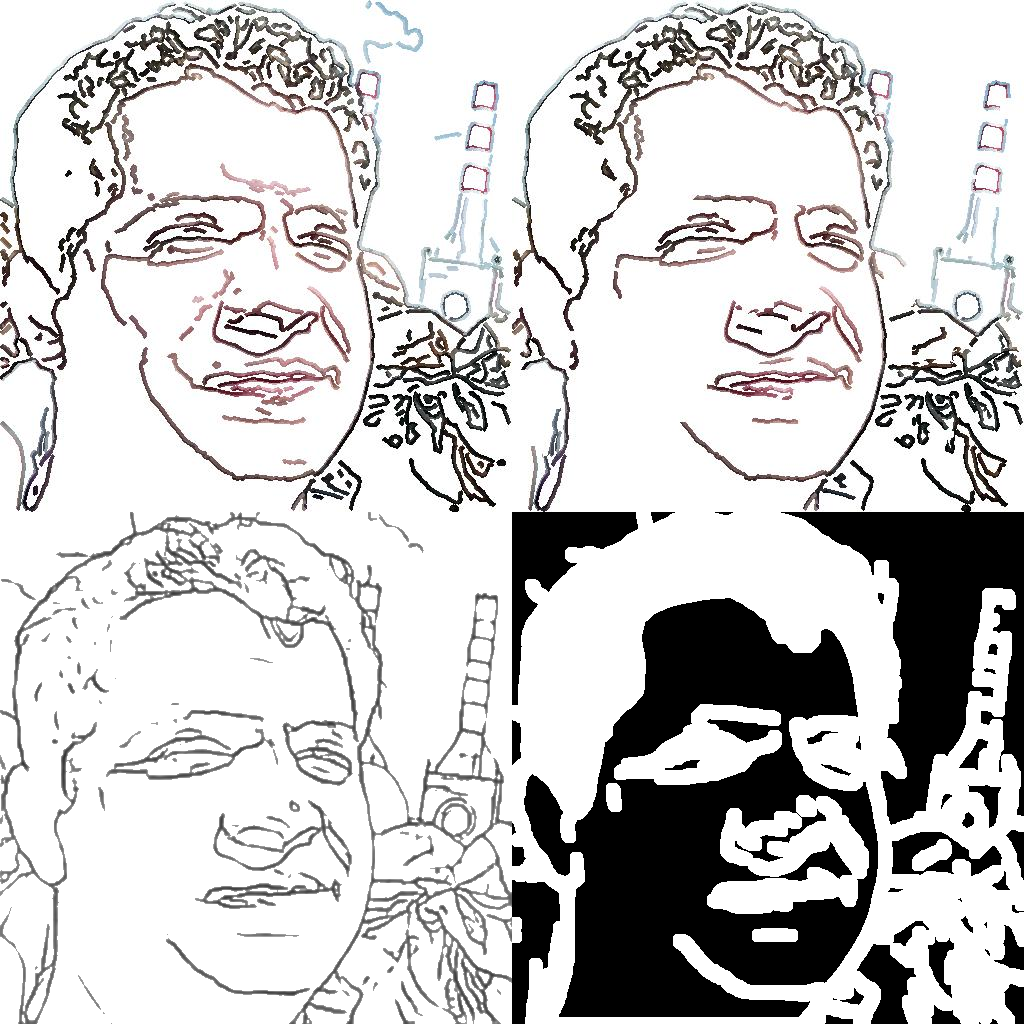}\\
\vspace{2pt}
\includegraphics[width=0.158\columnwidth]{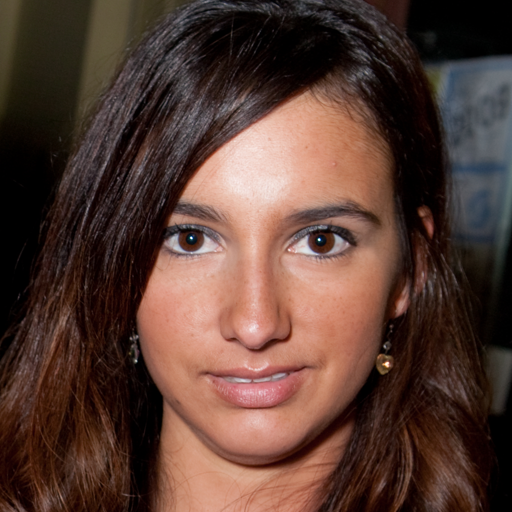}
\includegraphics[width=0.158\columnwidth]{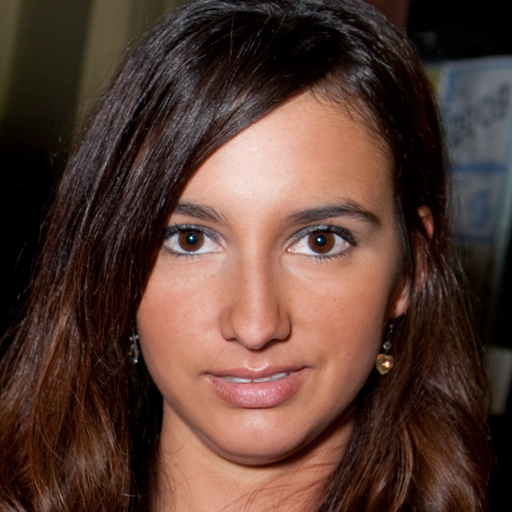}
\includegraphics[width=0.158\columnwidth]{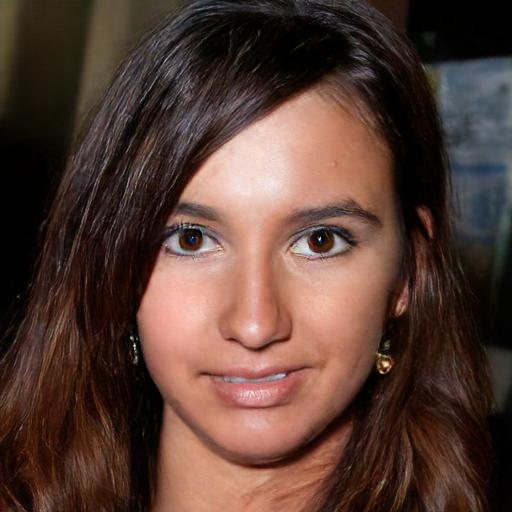}
\includegraphics[width=0.158\columnwidth]{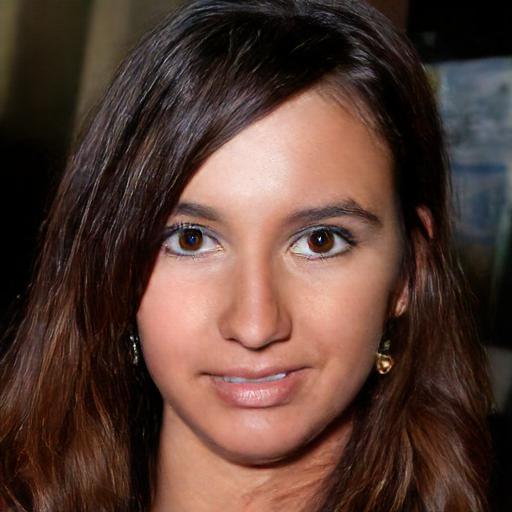}
\includegraphics[width=0.158\columnwidth]{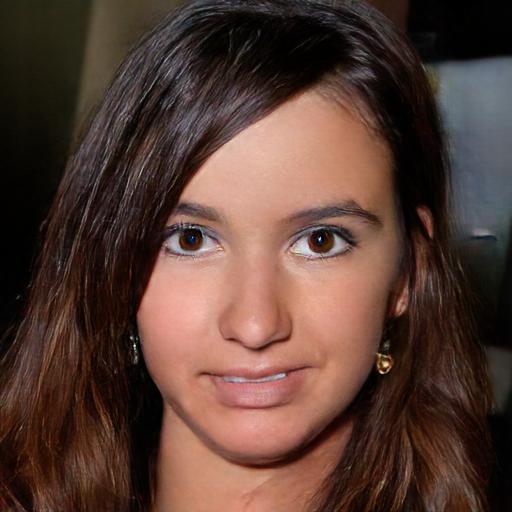}
\includegraphics[width=0.158\columnwidth]{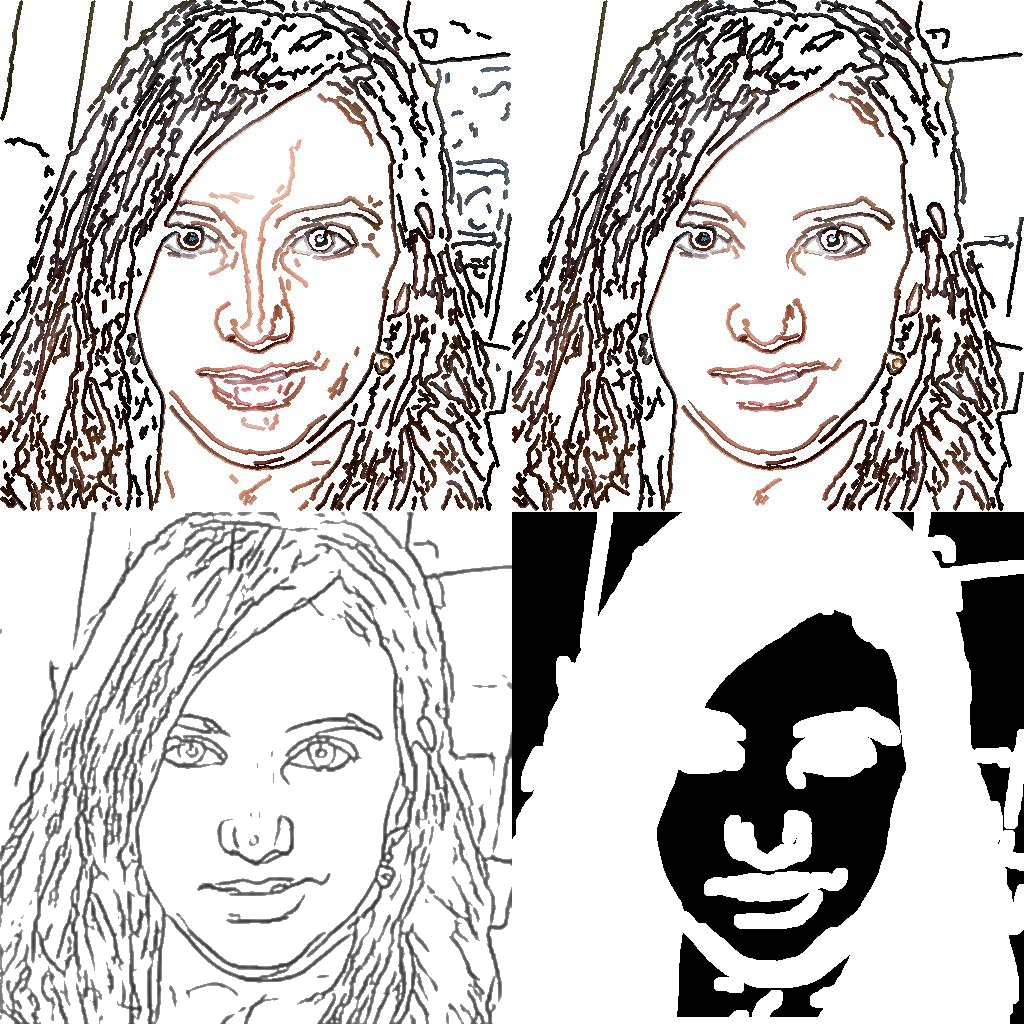}\\
\vspace{2pt}
\includegraphics[width=0.158\columnwidth]{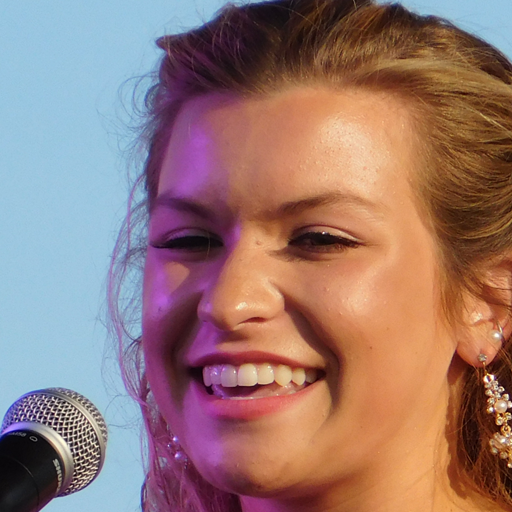}
\includegraphics[width=0.158\columnwidth]{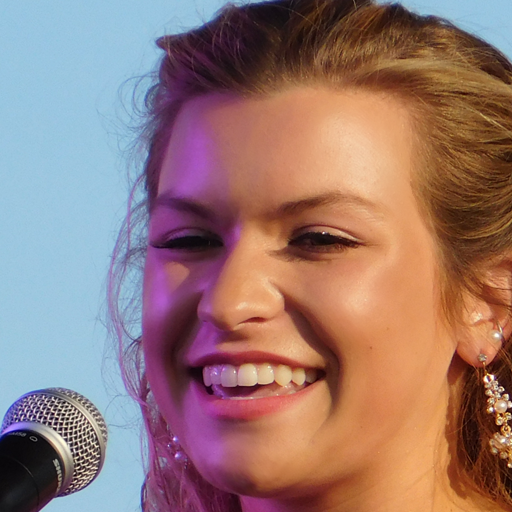}
\includegraphics[width=0.158\columnwidth]{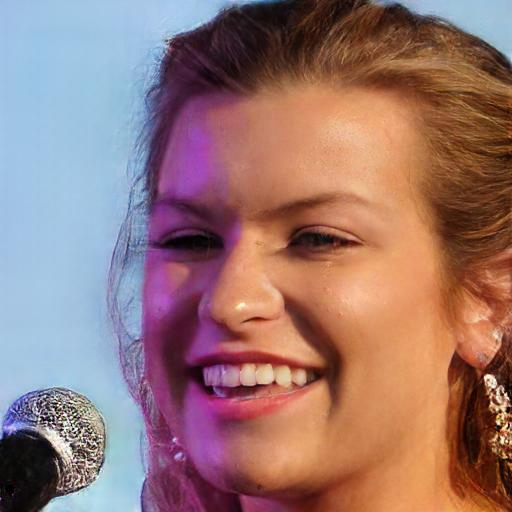}
\includegraphics[width=0.158\columnwidth]{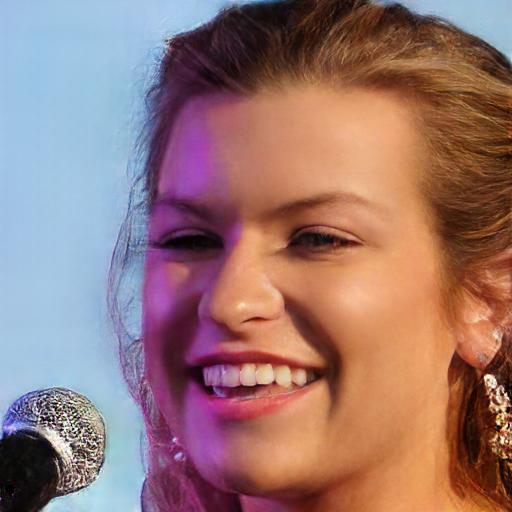}
\includegraphics[width=0.158\columnwidth]{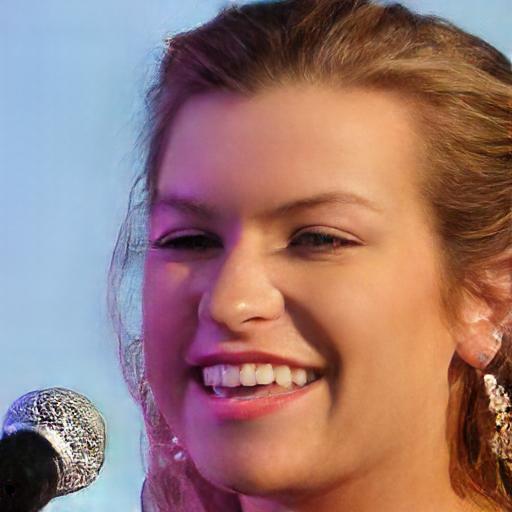}
\includegraphics[width=0.158\columnwidth]{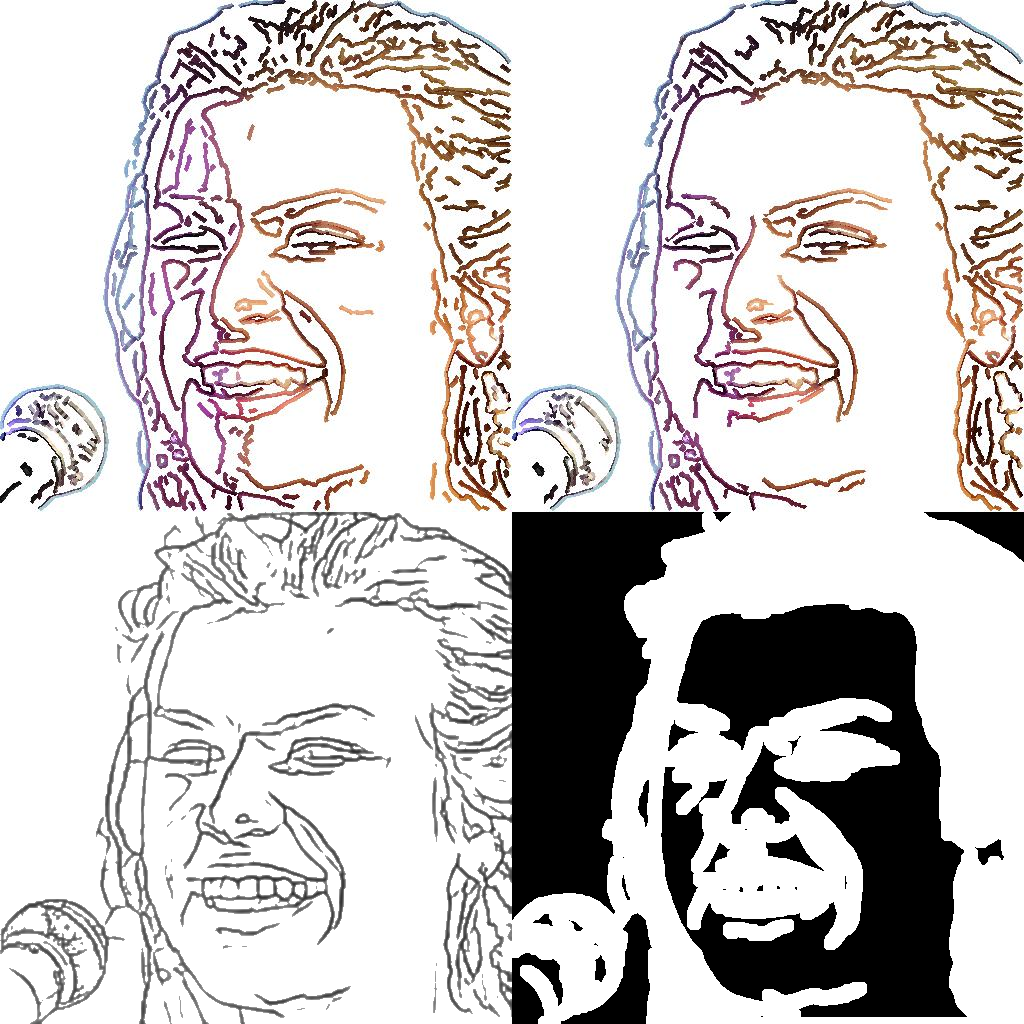}\\
\vspace{2pt}
\includegraphics[width=0.158\columnwidth]{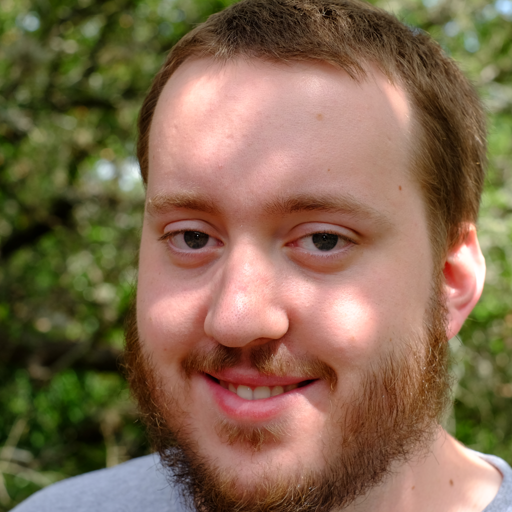}
\includegraphics[width=0.158\columnwidth]{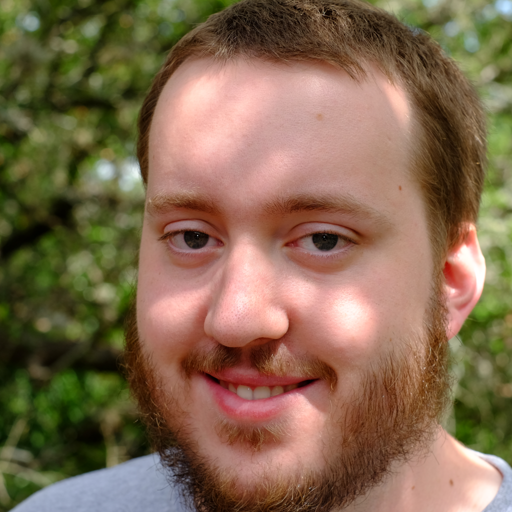}
\includegraphics[width=0.158\columnwidth]{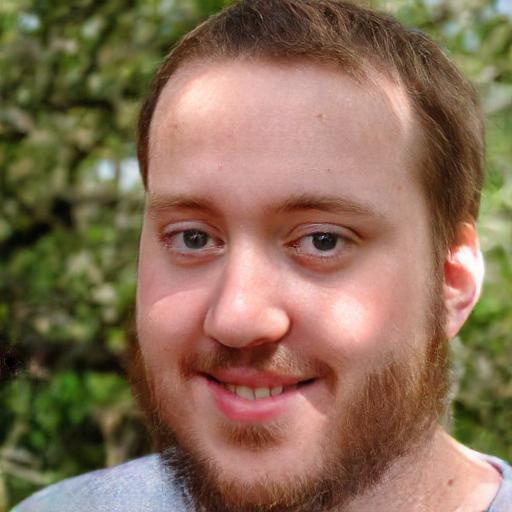}
\includegraphics[width=0.158\columnwidth]{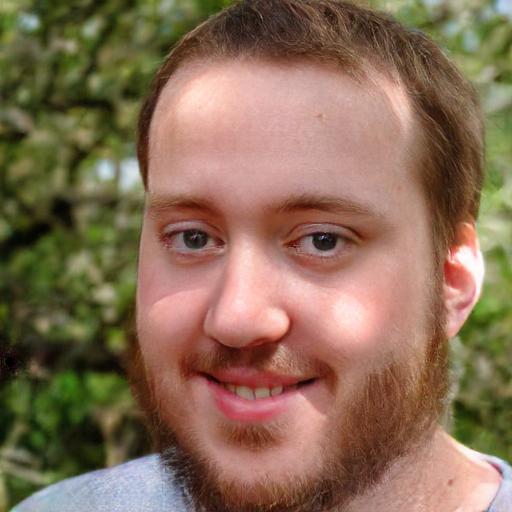}
\includegraphics[width=0.158\columnwidth]{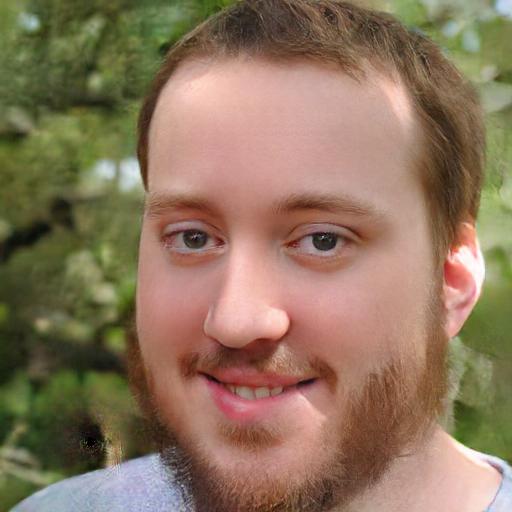}
\includegraphics[width=0.158\columnwidth]{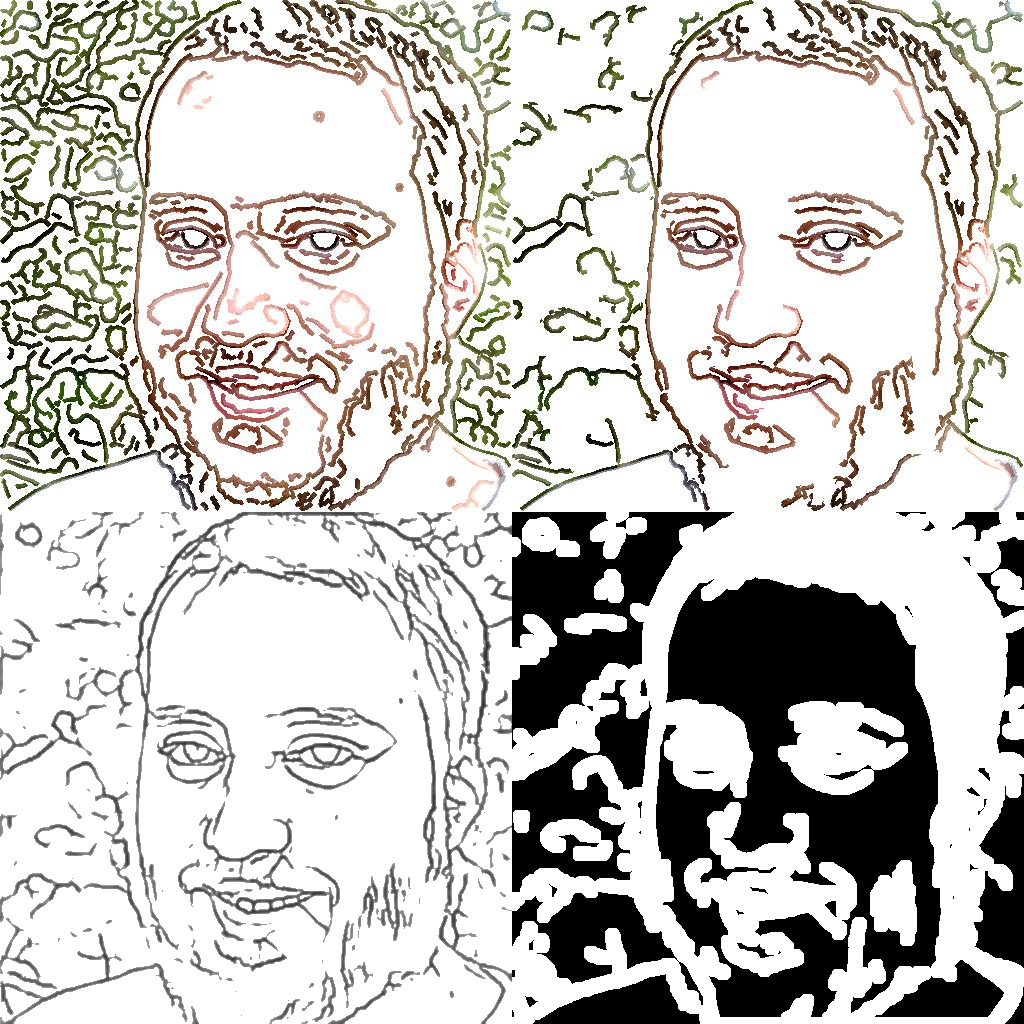}\\
\makebox[0.158\columnwidth]{(a) Original}
\makebox[0.158\columnwidth]{(b)}
\makebox[0.158\columnwidth]{(c)}
\makebox[0.158\columnwidth]{(d)}
\makebox[0.158\columnwidth]{(e)}
\makebox[0.158\columnwidth]{(f)}\\
\vspace{-2pt}
\makebox[0.158\columnwidth]{(DC+rPR+fPR)}
\makebox[0.158\columnwidth]{DC+rPR}
\makebox[0.158\columnwidth]{DC+rPR'+fPR}
\makebox[0.158\columnwidth]{DC+rPR'}
\makebox[0.158\columnwidth]{DC+rPR''}
\makebox[0.158\columnwidth]{DC primitives}}\\
\caption{{Generating residual PRs using deep learning.  {Clockwise from top left in the last column: the original DCs, the filtered DCs by applying DC masks, the DC masks and the extracted edges from portrait-tailored DexiNed. See main text for explanation.
}
The images are in high resolution, allowing zoom-in examination.}}
\label{fig:dc_ai}
\end{figure}

{Since residual PRs are coupled with the \textit{geometries} of diffusion curves, rPRs need to be updated whenever the user changes the shapes of DCs. Note that rPRs are pixel-sized Poisson regions that are not meant for direct editing. To address this issue, we train a deep generative model to encode high-frequency residuals. Specifically, we modify the image-to-image translation model, pix2pixHD \citep{wang2018pix2pixHD}, which takes diffusion curve images (DCI) as input and outputs residuals. We train the model to reconstruct the corresponding rPRs from the given DCIs.
As Figure \ref{fig:dc_ai} shows, the user only needs to modify sparse diffuse curves to edit expression without worrying about the high-frequency residuals. Throughout the paper, we use rPR' to denote the residual Poisson regions that are generated by the deep generative model, distinguishing itself from rPR, which is computed from the highlight compensated retouched image $I_{rh}$.
We also use rPR'' to denote the residual Poisson regions generated by the deep generative model that takes the original DCs \textit{with DC masks} as input.
In practice, rPR' is often used in geometry editing such that the original light is desired, whereas rPR'' is used in situations where light can be removed (Figure \ref{fig:dc_ai} (d) vs (e)). Also, with fine details Poisson regions, the reconstructed vector images are photo-realistic (Figure \ref{fig:dc_ai} (c) vs (d)).
}

\section{\normalsize Illumination-Sensitive Image Quality Metrics}
To quantitatively evaluate the results of color and highlight transfer, we propose a new metric, called illumination-sensitive IS-FLIP or IS-FLIP-ct, that measures the difference for the reconstruction images and the color-transferred images.
IS-FLIP is an extension of the recent FLIP metric~\citep{Andersson2020Flip}, which is a difference evaluator with a particular focus on the differences between rendered images and corresponding ground truths.
Given two images $I$ and $J$, FLIP computes the per-pixel color and feature differences $\delta E_c$ and $\delta E_f \in[0,1]$, respectively. Then the FLIP difference value, $\delta E_{F}$, per pixel is given by
\vspace*{-0.5\baselineskip}
$$\delta E_{F} = \delta E_c^{\left(1-\delta E_f\right)}.
\vspace*{-0.5\baselineskip}$$
FLIP produces a map that approximates the difference perceived by humans when alternating between $I$ and $J$~\citep{Andersson2020Flip}.
Though FLIP is a powerful difference evaluator, it does not consider illumination differences. Since human vision is more sensitive to large illumination differences such as highlights and shadows rather than small details in portrait images, FLIP is not an effective measure for portrait images.

IS-FLIP improves FLIP by adding a term of illumination difference $\delta E_i$,
\vspace*{-0.5\baselineskip}
$$\delta E_i(I,J)=\left(|I_h-J_h|+|I_s-J_s|\right)/2,
\vspace*{-0.5\baselineskip}$$
where {$I_h$} and {$J_h$} are
the extracted highlight obtained by the highlight removal algorithm \citep{shen2013real},
and {$I_s$} and {$J_s$} are the extracted shadow by applying the algorithm \citep{shen2013real} on the inverted images. We then define IS-FLIP $\delta E_I$ as
\vspace*{-0.5\baselineskip}
\begin{equation}
\delta E_I = \delta E_i^{\left(1- \delta E_F\right)}.
\label{eqn:hflip}
\vspace*{-0.5\baselineskip}
\end{equation}

We adopt $\delta E_{I}$ to evaluate the accuracy of vectorization, since it can capture the differences between an image and its alternative image based on human perception.

\begin{figure*}[htbp]
\centering
{\scriptsize
\includegraphics[width=0.11\columnwidth]{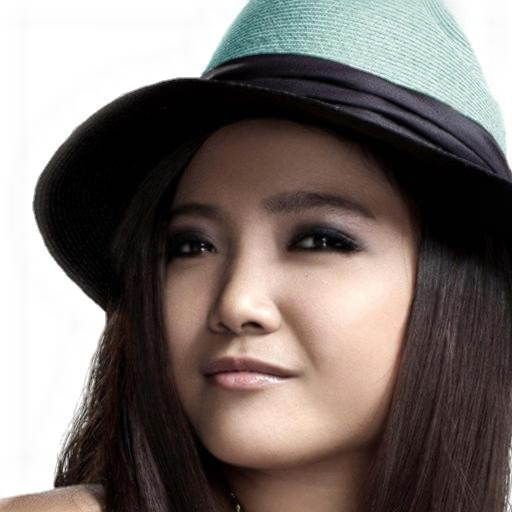}
\includegraphics[width=0.11\columnwidth]{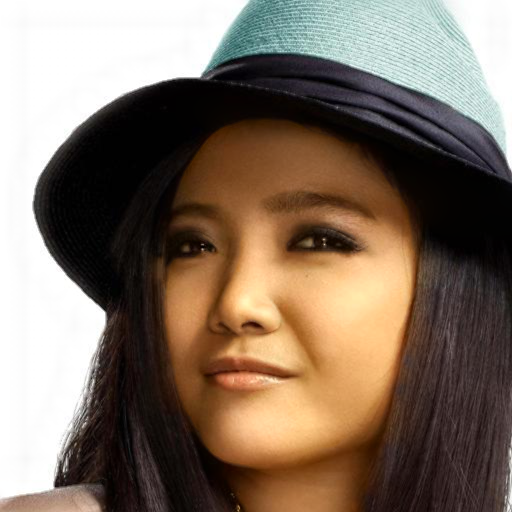}
\includegraphics[width=0.11\columnwidth]{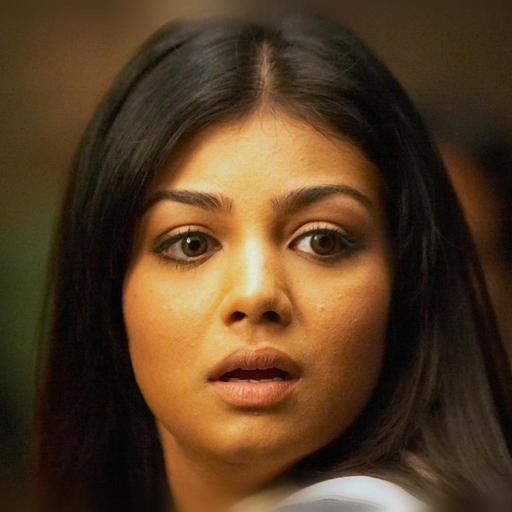}
\includegraphics[width=0.11\columnwidth]{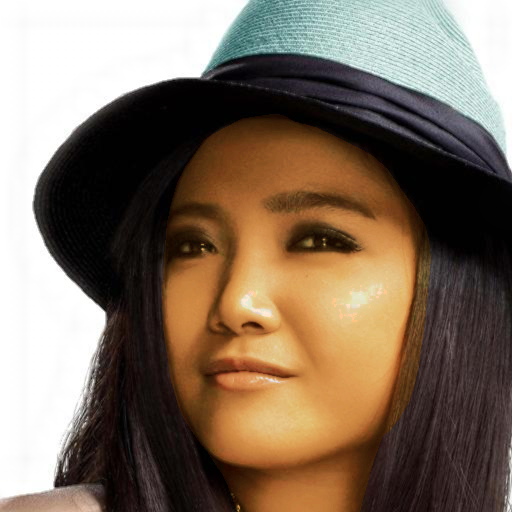}
\includegraphics[width=0.11\columnwidth]{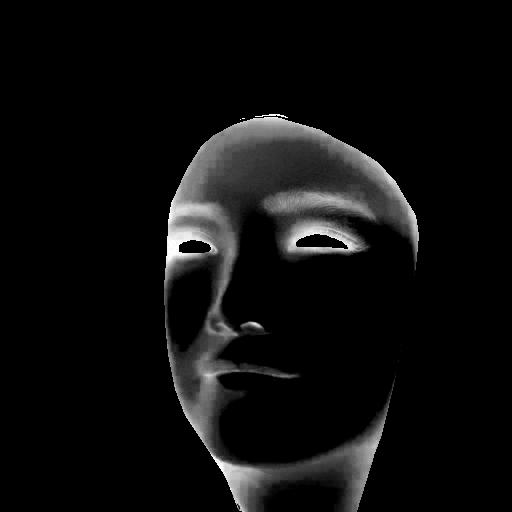}
\includegraphics[width=0.11\columnwidth]{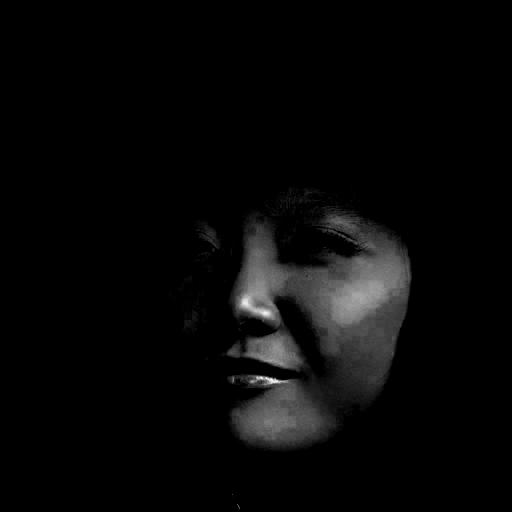}
\includegraphics[width=0.11\columnwidth]{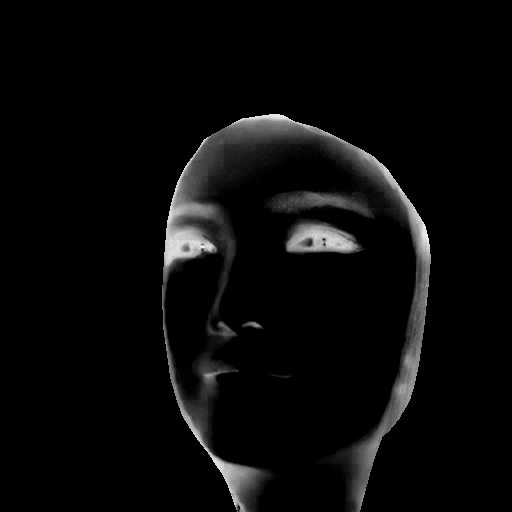}
\includegraphics[width=0.11\columnwidth]{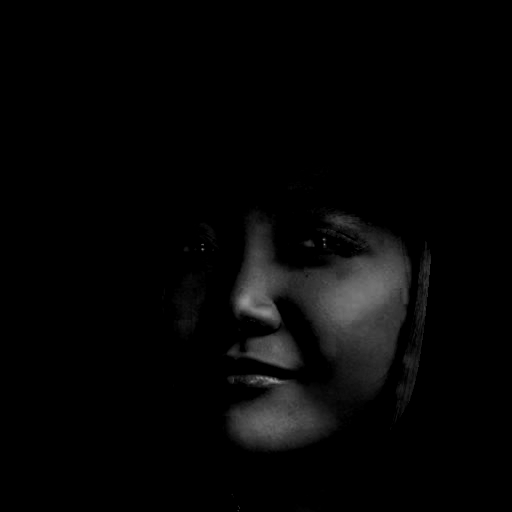}\\
\makebox[0.111\columnwidth]{$I$}
\makebox[0.111\columnwidth]{$J$}
\makebox[0.111\columnwidth]{$R$}
\makebox[0.111\columnwidth]{$M(I,R)$}
\makebox[0.111\columnwidth]{{$I_s$}}
\makebox[0.111\columnwidth]{{$I_h$}}
\makebox[0.111\columnwidth]{{$J_s$}}
\makebox[0.111\columnwidth]{{$J_h$}}\\
\vspace{1pt}
\includegraphics[width=0.106\columnwidth]{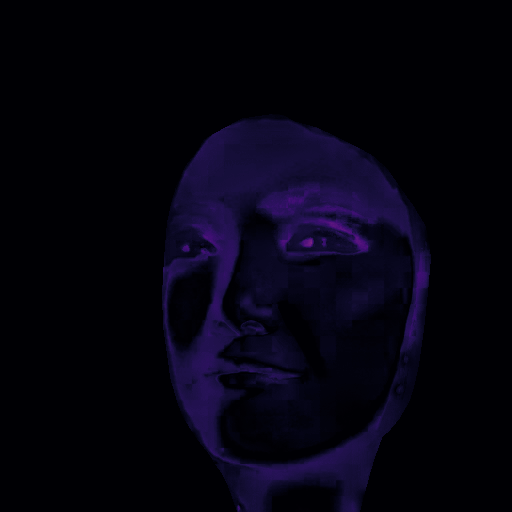}
\includegraphics[width=0.106\columnwidth]{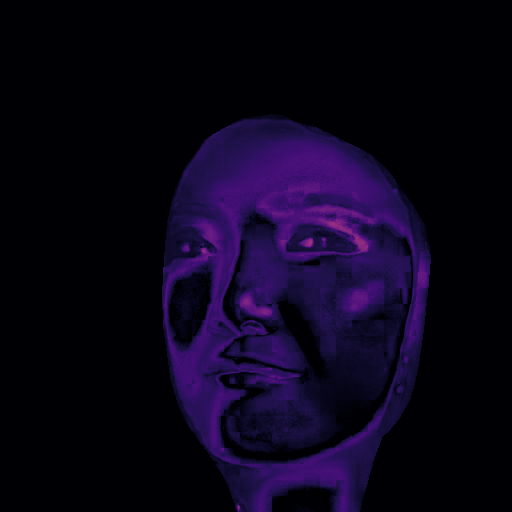}
\includegraphics[width=0.106\columnwidth]{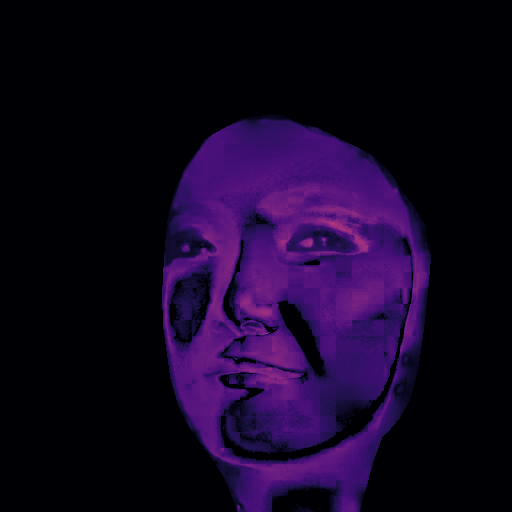}
\includegraphics[width=0.106\columnwidth]{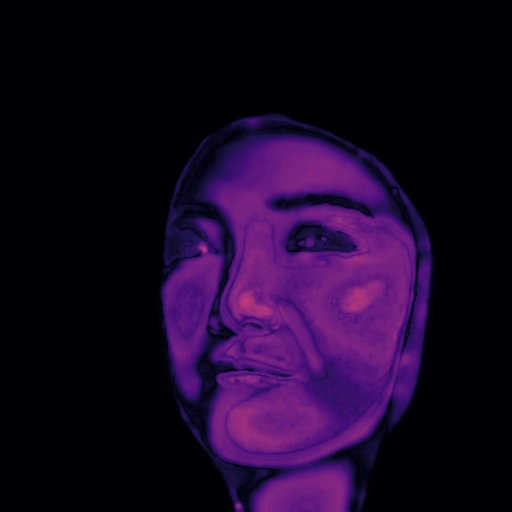}
\includegraphics[width=0.106\columnwidth]{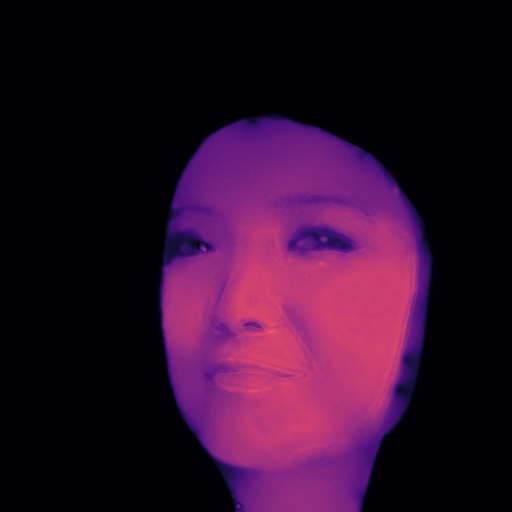}
\includegraphics[width=0.106\columnwidth]{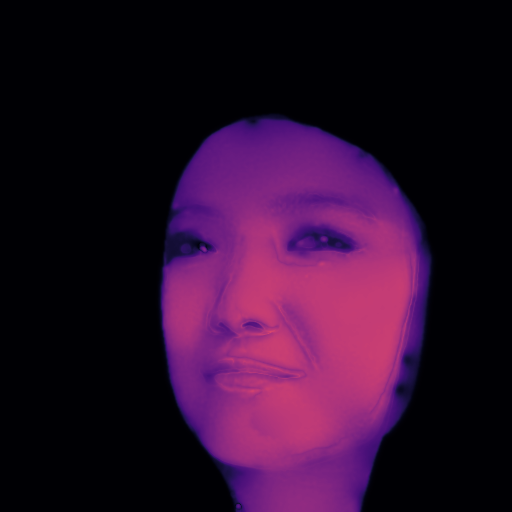}
\includegraphics[width=0.106\columnwidth]{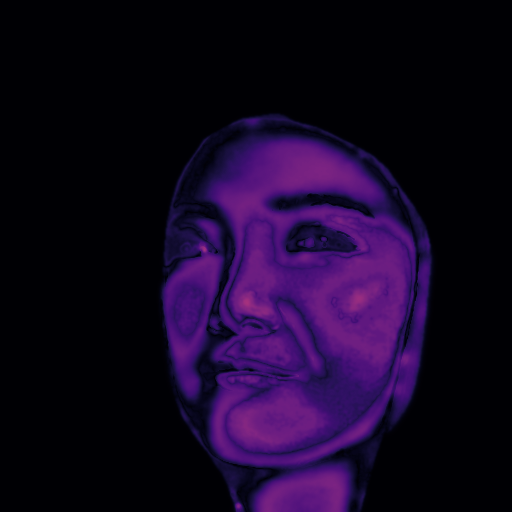}
\includegraphics[width=0.106\columnwidth]{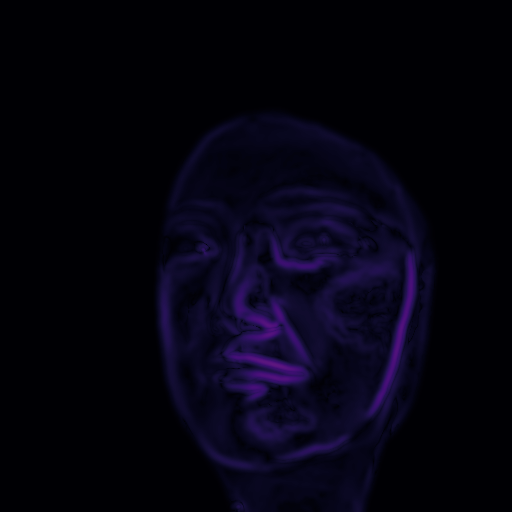}
\includegraphics[width=0.106\columnwidth]{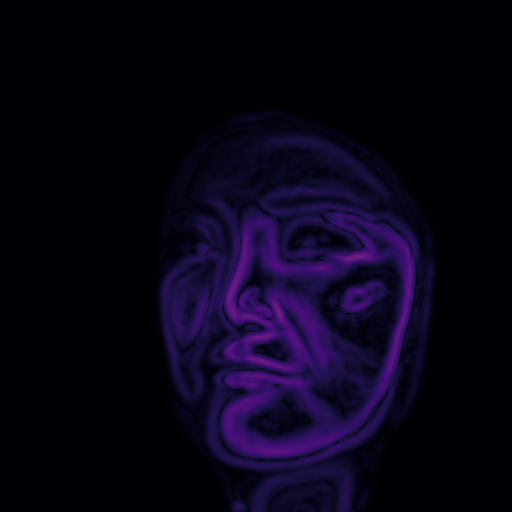}\\
\makebox[0.106\columnwidth]{$\delta E_i$}
\makebox[0.106\columnwidth]{$\delta E^{ct}_I$}
\makebox[0.106\columnwidth]{$\delta E_I$}
\makebox[0.106\columnwidth]{$\delta E^{ct}_F$}
\makebox[0.106\columnwidth]{$\delta E_F$}
\makebox[0.106\columnwidth]{$\delta E_c$}
\makebox[0.106\columnwidth]{$\delta E^{ct}_c$}
\makebox[0.106\columnwidth]{$\delta E_f$}
\makebox[0.106\columnwidth]{$\delta E^{ct}_f$}\\
\vspace{1pt}
\makebox[0.106\columnwidth]{2.36\%}
\makebox[0.106\columnwidth]{4.26\%}
\makebox[0.106\columnwidth]{6.51\%}
\makebox[0.106\columnwidth]{8.03\%}
\makebox[0.106\columnwidth]{13.44\%}
\makebox[0.106\columnwidth]{12.57\%}
\makebox[0.106\columnwidth]{6.76\%}
\makebox[0.106\columnwidth]{2.46\%}
\makebox[0.106\columnwidth]{3.64\%}\\
}
\caption{The IS-FLIP-ct metric $\delta E^{ct}_I$ is more effective than FLIP $\delta E_F$ and IS-FLIP $\delta E_I$ for evaluating color transfer results. The input image $I$ and the color transferred result $J$ have the same facial features but different colors. However, due to significant color change between $J$ and $I$, both FLIP and IS-FLIP, which take color changes as part of the difference metric, yield large error values. IS-FLIP-ct, in contrast, can reduce such effect with a reference color transfer result $M(I,R)$, hereby is more effective than FLIP and IS-FLIP for evaluating color transfer results. We visualize the error metrics to compare their effectiveness: the less features in the error maps, the more effective the error metric is for measuring color transfer quality. Human perception also confirms the efficacy of IS-FLIP-ct.}
\label{fig:hflip}
\vspace*{-0.5\baselineskip}
\end{figure*}

Notice that IS-FLIP involves only two images $I$ and $J$. To suit IS-FLIP for color transfer, we define a variant, called IS-FLIP-ct, which considers the input image $I$, the reference image $R$ and the transferred image $J$. Denote by $M(I,R)$ the transferred result by applying a baseline histogram matching method $M$ (e.g., MATLAB's 'imhistmatch' function) to $I$ and $R$.
Specifically, we can rewrite $\delta E_F$ by measuring the feature differences between $J$ and $M(I,R)$
\vspace*{-0.5\baselineskip}
$$\delta E^{ct}_F(I,J,R)\triangleq\delta E_F\left(M(I,R),J\right).
\vspace*{-0.5\baselineskip}$$
Then we define IS-FLIP-ct for color transfer as
\vspace*{-0.5\baselineskip}
\begin{equation}
\delta E^{ct}_I = \delta E_i^{\left(1- \delta E^{ct}_F\right)},
\label{eqn:hflip_c}
\vspace*{-0.5\baselineskip}
\end{equation}
where the superscript denotes the variable for color transfer.
As shown in Figure \ref{fig:hflip},  IS-FLIP-ct is more effective than FLIP and IS-FLIP for evaluating the quality of color transfer results.

There are also other metrics to evaluate the quality of color transfer, such as structural similarity index measure (SSIM)~\citep{wang2004image}, the KL divergence and Hellinger distance~\cite{afifi2021histogan}. In Figure \ref{fig:details_new}, we show FLIP, SSIM and the proposed IS-FLIP measures.
HistoGAN~\cite{afifi2021histogan} adopts the KL divergence and Hellinger distance to measure the similarity between the target histogram and the histogram of the color-changed images.
Notice that KL divergence and Hellinger distance focus on only the difference of two color distributions, they do not consider illumination changes.
Figure \ref{fig:color.transfer} shows the color transfer results of different methods. Judging by KL divergence and Hellinger distance which were used in \cite{afifi2021histogan}, the order from high to low quality is (g), (f), (d), (b), (c) and (h). However, if measured by our proposed IS-FLIP, the order is (b), (c), (f), (g), (h) and (d). Comparing with FLIP and SSIM, we observe that IS-FLIP is more consistent with human perception, since it considers color, feature as well as illumination.
We visualize the SSIM, FLIP and IS-FLIP measures in Figure~\ref{fig:details_new}.

\section{Experimental Results}
\textbf{Evaluation.} We evaluated our method on the CelebAMask-HQ dataset~\citep{CelebAMask-HQ}, which consists of 30K high-resolution photos of celebrities. Each image has segmentation mask of facial attributes, such as hair, face, brows, eyes, nose, lips, ear, neck and cloth.
To demonstrate the advantages of our {hierarchical} PVG, we applied it to color transfer, which can be mathematically modeled with optimal mass transport (OMT). We compared our method with 3 raster-image based OMT methods and 2 vectorization based OMT methods. The raster-image based methods apply OMT to the original images, the  intrinsic images, and the $L_1$ smoothed images. The vectorization based methods are the {2-level} PVG~\citep{fu2019vectorization}, and the intrinsic image-based PVG which vectorizes each decomposed layer using~\citep{fu2019vectorization}. Table \ref{tab:comparison_acc} reports the mean and variance of the IS-FLIP-ct and FLIP metrics on 500 representative images (No. 1500-1999) of the CelebAMask-HQ dataset.
To make a fair comparison with intrinsic image based methods, we calculate IS-FLIP and FLIP for the facial regions, since the existing intrinsic image decomposition can only process the face regions. For vectorization, our method can obtain similar accuracy of the conventional PVG vectorization method~\cite{fu2019vectorization}. For color transfer, our method yields results with the highest quality. The visual results also confirm that our method can effectively keep the original illumination and facial details in color transfer.

\begin{table}[htbp!]
 \centering
 \setlength\tabcolsep{1pt}
 \scalebox{0.85}{
 \begin{tabular}{lcccccccc}
 \toprule
  & \multicolumn{4}{c}{Vectorization} &  \multicolumn{4}{c}{Color Transfer} \\
 \midrule
   & \multicolumn{2}{c}{Mean (\%)} & \multicolumn{2}{c}{Variance} & \multicolumn{2}{c}{Mean (\%)} & \multicolumn{2}{c}{Variance} \\
   & IS-FLIP & FLIP& IS-FLIP & FLIP& IS-FLIP-ct & FLIP& IS-FLIP-ct & FLIP\\
 \midrule
  raster & - & -& - & - &6.02& 13.00 &1.57e-3& 3.68e-3 \\
 $L_1$ smoothing & - & -& - & - &6.09& 13.41 &1.54e-3& 3.56e-3  \\
 intrinsic raster &2.16& 6.82 &1.54e-4& 3.04e-4 &9.69& 19.42 &2.41e-3& 2.76e-3 \\
 intrinsic PVG &5.72& 15.96 &6.56e-4& 9.18e-4 &9.92& 19.48 &2.45e-3& 2.76e-3 \\
 conventional PVG &0.16& 0.81 &3.59e-7& 4.68e-6 &6.17 & 14.46 &1.62e-3 & 3.32e-3 \\
 {hierarchical} PVG &\bf{0.20} & \bf{1.02} &\bf{4.53e-7}& \bf{5.68e-6} &\bf{5.25}& \bf{12.99} &\bf{1.18e-3}& \bf{2.52e-3} \\
 {DL-PVG} &\bf{2.40} & \bf{7.55} &\bf{1.03e-4}& \bf{3.37e-4} & - & -& - & - \\
 \bottomrule
 \end{tabular}}
 \vspace{2pt}
 \caption{{Evaluating vectorization accuracy and color transfer quality using FLIP, IS-FLIP, IS-FLIP-ct measures. {DL-PVG is DCs combined with a deep generative model for geometry editing.}}}
 \label{tab:comparison_acc}
 \end{table}

\begin{figure}[htbp]
\centering
{\scriptsize
\includegraphics[height=0.195\columnwidth]{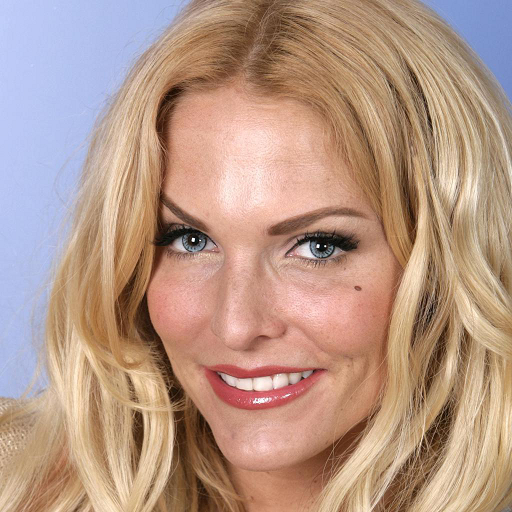}
\includegraphics[height=0.195\columnwidth]{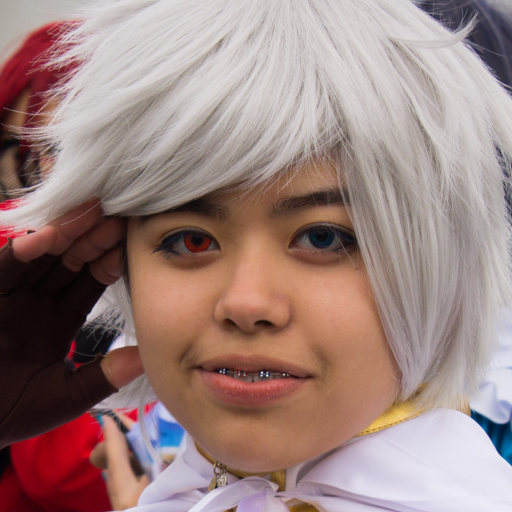}
\includegraphics[height=0.195\columnwidth]{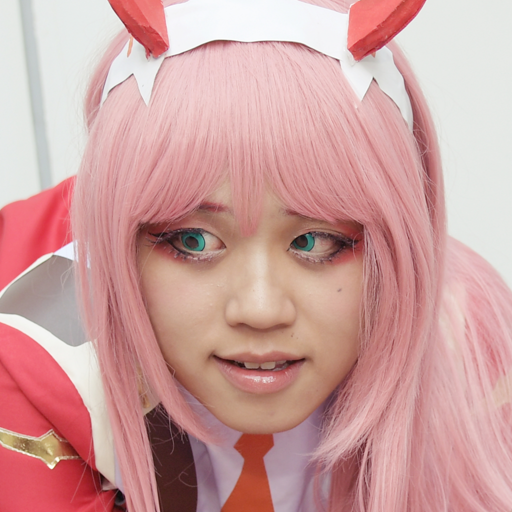}
\includegraphics[height=0.195\columnwidth]{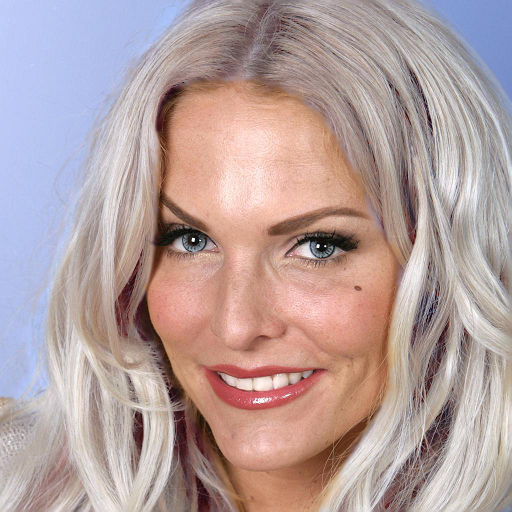}
\includegraphics[height=0.195\columnwidth]{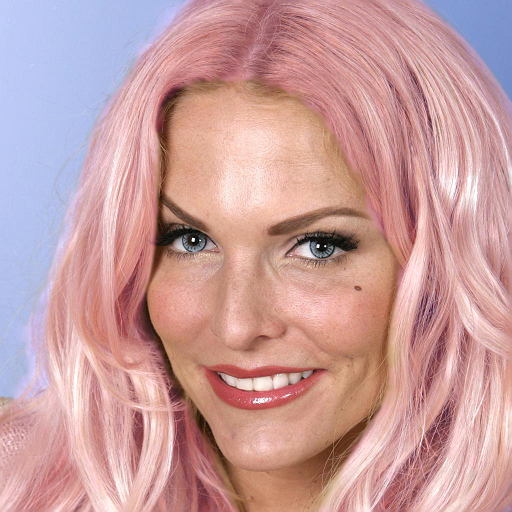}\\
\makebox[0.195\columnwidth]{(a)}
\makebox[0.195\columnwidth]{(b)}
\makebox[0.195\columnwidth]{(c)}
\makebox[0.195\columnwidth]{(d)}
\makebox[0.195\columnwidth]{(e)}\\
\vspace{2pt}
\includegraphics[height=0.195\columnwidth]{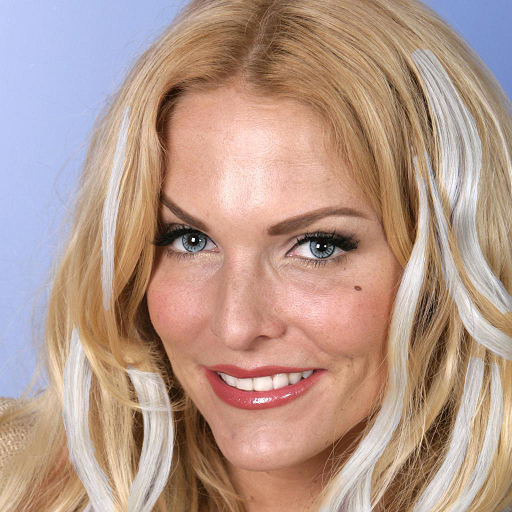}
\includegraphics[height=0.195\columnwidth]{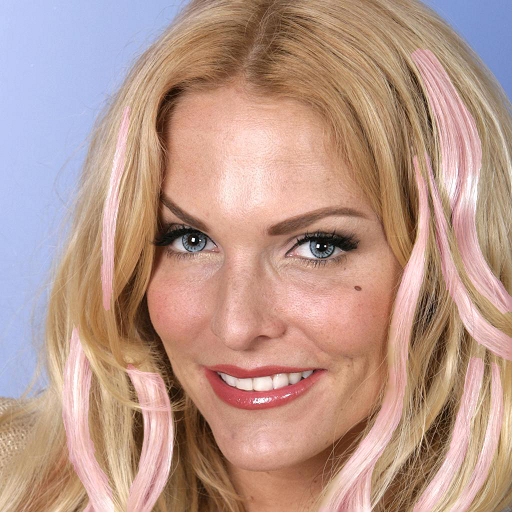}
\includegraphics[height=0.195\columnwidth]{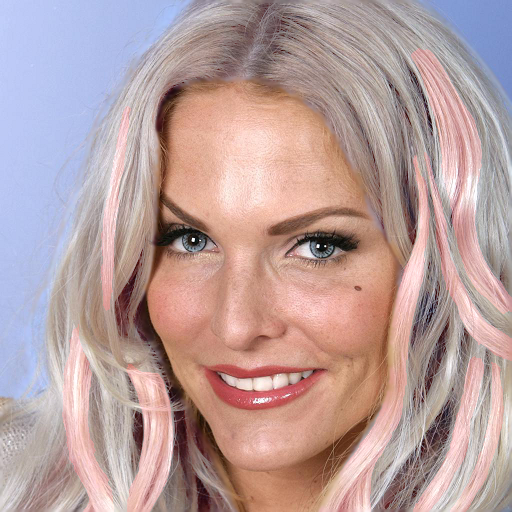}
\includegraphics[height=0.195\columnwidth]{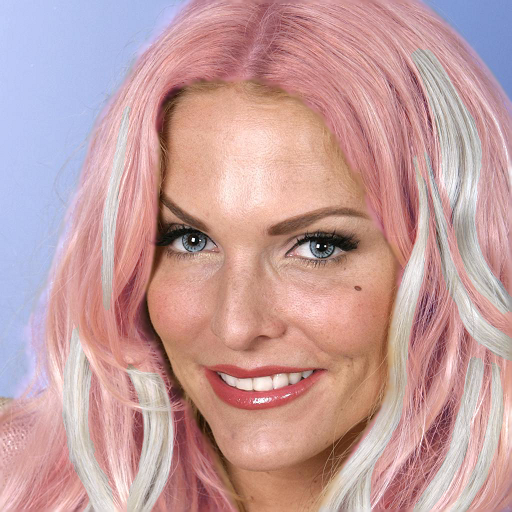}\\
\makebox[0.195\columnwidth]{(f)}
\makebox[0.195\columnwidth]{(g)}
\makebox[0.195\columnwidth]{(h)}
\makebox[0.195\columnwidth]{(i)}}
\caption{\fq{Hair color transfer by modifying colors of sparse diffusion curves. (a) original image. (b)-(c) reference images. (d)-(e) hair color transfer. (f)-(i) hair highlighting effects.}}
\label{fig:hair_color}
\end{figure}

\begin{figure*}[htbp]
\centering
{\scriptsize
\begin{minipage}[b]{0.74\linewidth}
\includegraphics[width=0.14\linewidth]{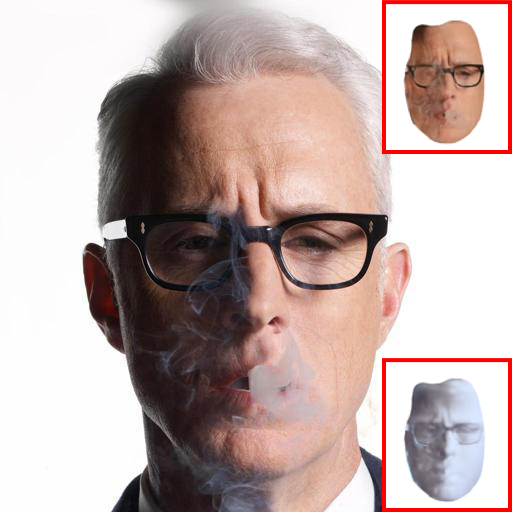}
\includegraphics[width=0.28\linewidth]{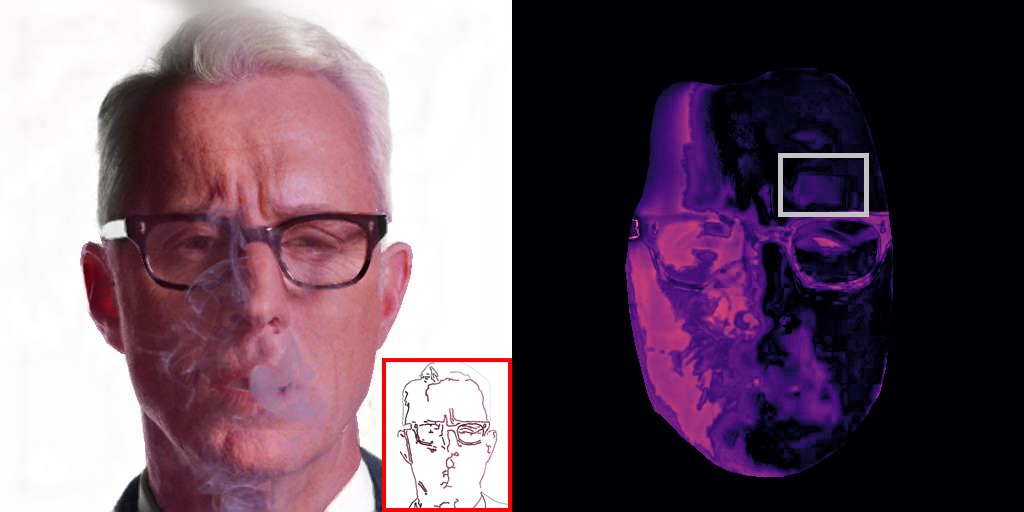}
\includegraphics[width=0.28\linewidth]{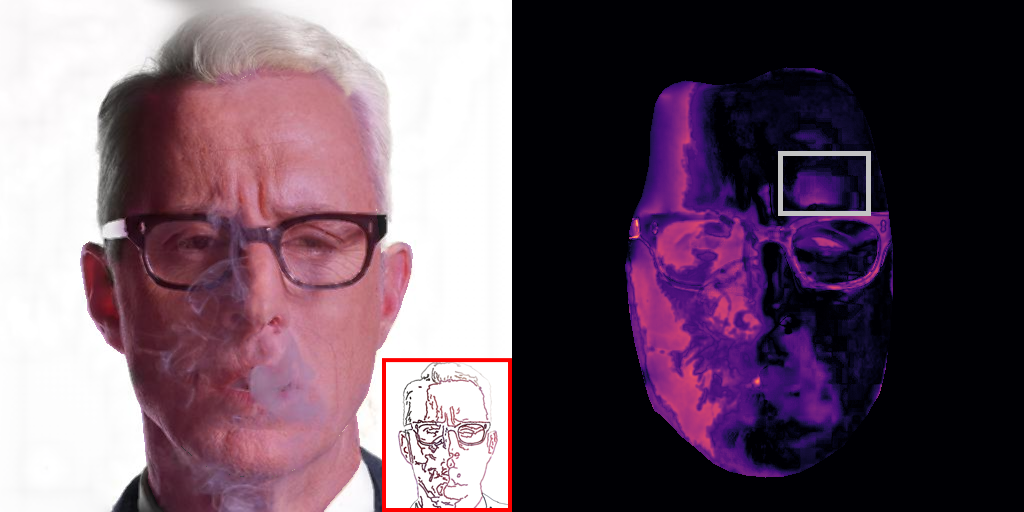}
\includegraphics[width=0.28\linewidth]{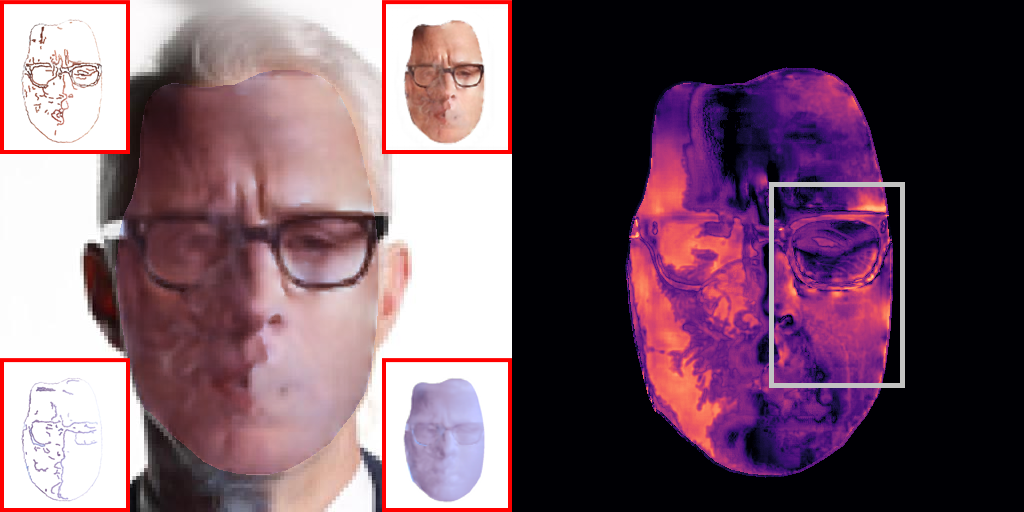}\\
\makebox[0.14\linewidth]{(a) Original}
\makebox[0.28\linewidth]{(b) Our method 6.41\%}
\makebox[0.28\linewidth]{(c) PVG \cite{fu2019vectorization} 6.59\%}
\makebox[0.28\linewidth]{(d) Intrinsic PVG 12.32\%}\\
\includegraphics[width=0.14\linewidth]{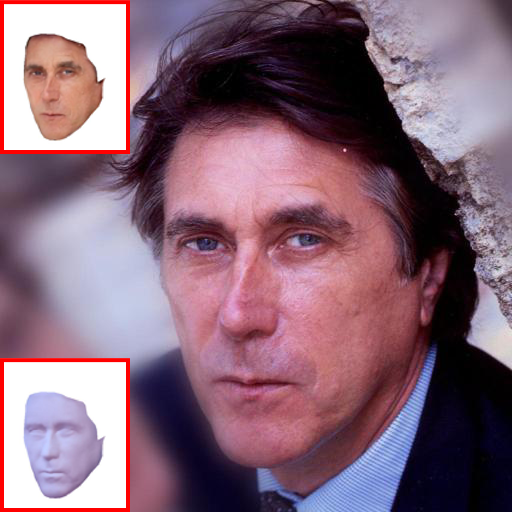}
\includegraphics[width=0.28\linewidth]{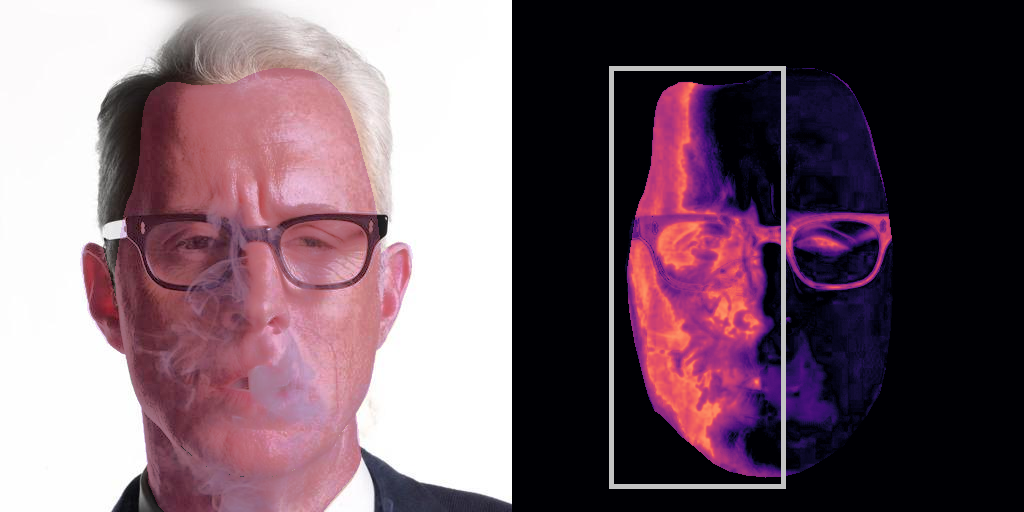}
\includegraphics[width=0.28\linewidth]{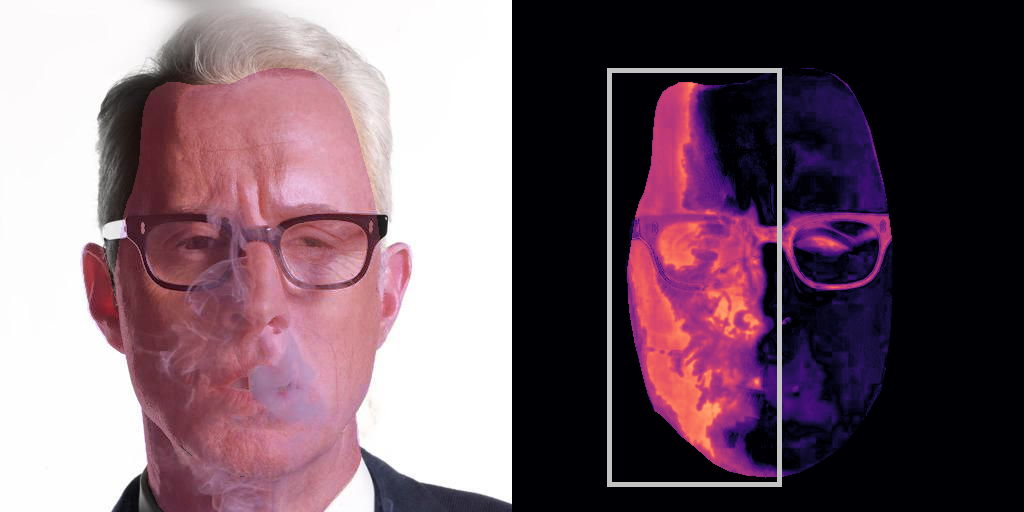}
\includegraphics[width=0.28\linewidth]{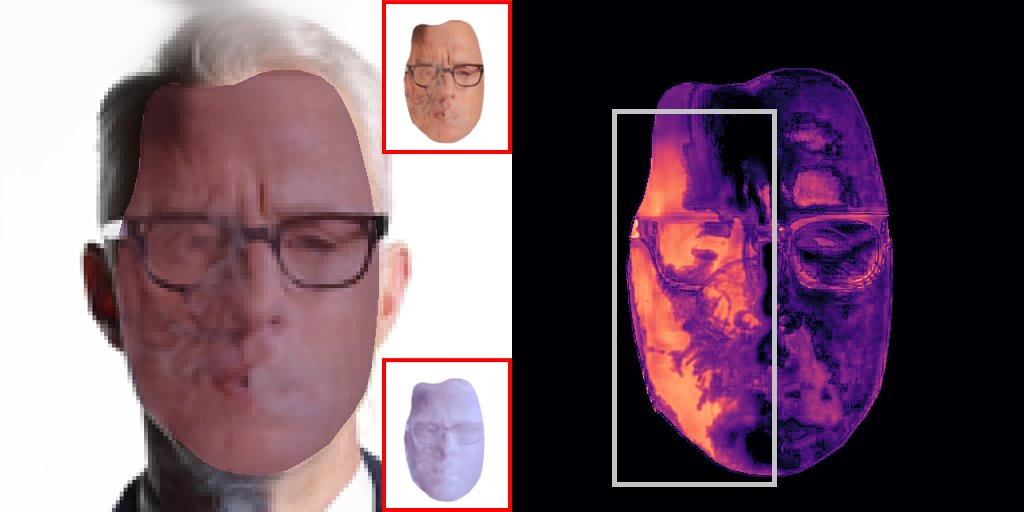}\\
\makebox[0.14\linewidth]{(e) Reference}
\makebox[0.28\linewidth]{(f) Raster 9.18\%}
\makebox[0.28\linewidth]{(g) Smooth 9.51\%}
\makebox[0.28\linewidth]{(h) Intrinsic raster 11.29\%}\\
\end{minipage}
\begin{minipage}[b]{0.25\linewidth}
\includegraphics[width=1.0\linewidth]{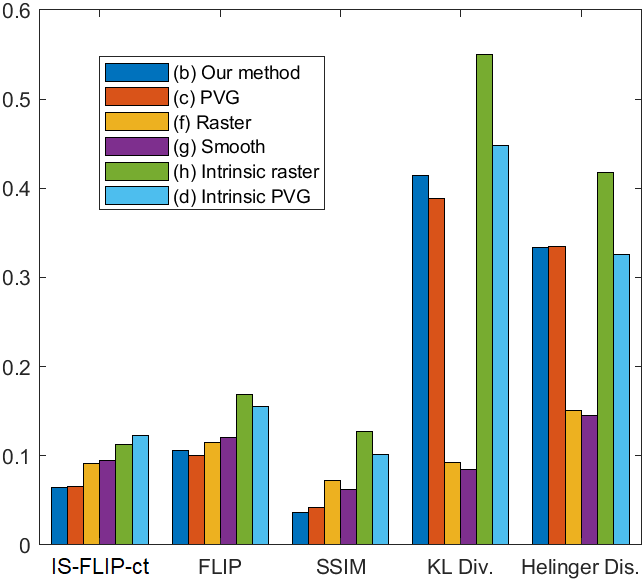}\\
\makebox[1.0\linewidth]{(i) IS-FLIP-ct \& other metrics}\\
\end{minipage}}\\
\vspace*{-0.5\baselineskip}
\caption{Facial color transfer on a challenging case with a wide range of brightness. We show the results of ours (b) and Fu et al.'s method \cite{fu2019vectorization} (c), intrinsic decomposition~\cite{sengupta2018sfsnet} applied to PVG~\cite{fu2019vectorization} (d),
raster image OMT (f), OMT combined with L1-smoothing~\cite{bi20151l1} (g), and OMT combined with intrinsic image~\cite{sengupta2018sfsnet} (h).
We also visualize the proposed IS-FLIP-ct error maps. The less facial details in the error maps, the higher the quality of color transfer. (i) shows the values of other popular metrics, including FLIP, SSIM, KL divergence and Helinger distance.
Both the quality metric IS-FLIP-ct and visual error maps confirm our method is superior than the other approaches.
}
\label{fig:color.transfer}
\end{figure*}

 \textbf{Semantic color transfer.} Since our method encodes low-frequency colors in a set of sparse diffusion curves with semantic meanings, semantic color transfer becomes easy. For example, \fq{the user can change colors semantically in the base level such as facial and hair color transfer shown in Figures \ref{fig:color.transfer} and \ref{fig:hair_color}. Note that we can definitely edit the targeted curves to generate hair highlighting effect thanks to vector representation, which is challenging if only using deep learning based methods.}

\begin{figure}[htbp]
\centering
{\scriptsize
\includegraphics[width=0.242\columnwidth]{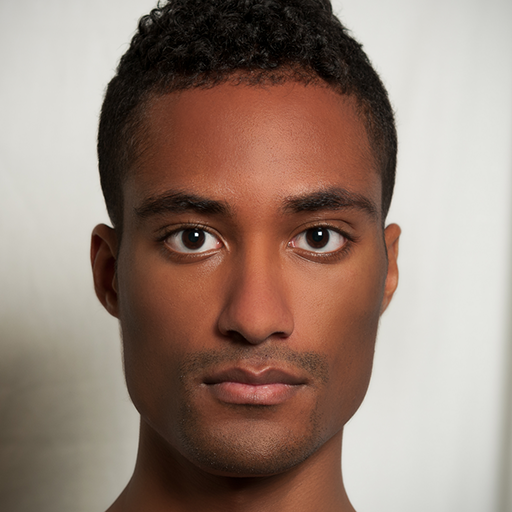}
\includegraphics[width=0.242\columnwidth]{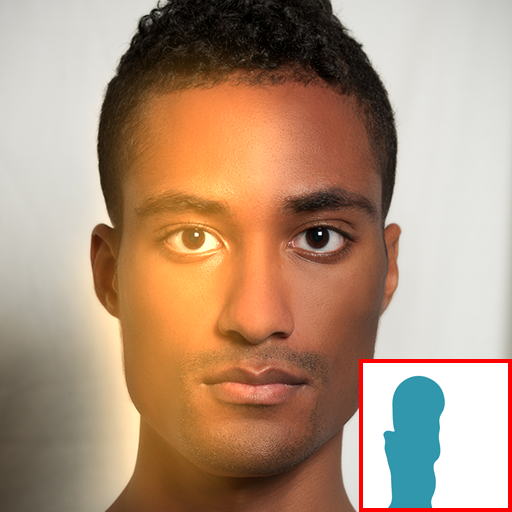}
\includegraphics[width=0.242\columnwidth]{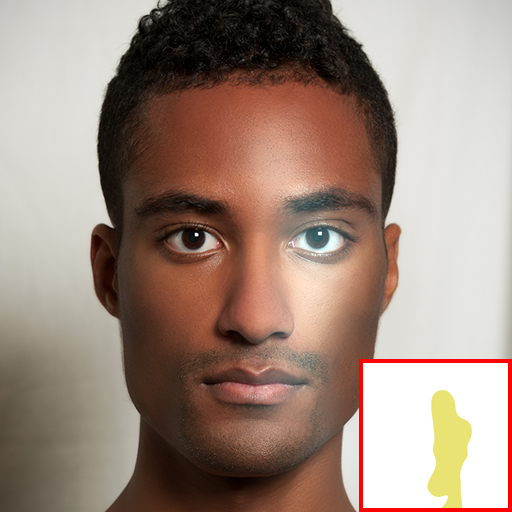}
\includegraphics[width=0.242\columnwidth]{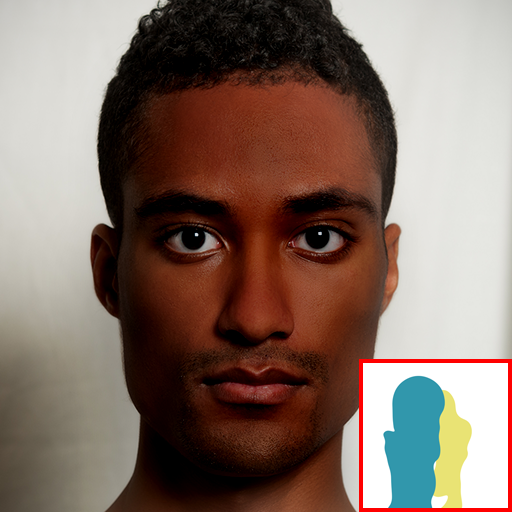}\\
\makebox[0.242\columnwidth]{Original}
\makebox[0.242\columnwidth]{D: (21,13,4); 70\%}
\makebox[0.242\columnwidth]{D: (24,44,47); 80\%}
\makebox[0.242\columnwidth]{B: (-22,-19,-13); 12\%}\\
\makebox[0.242\columnwidth]{}
\makebox[0.242\columnwidth]{}
\makebox[0.242\columnwidth]{}
\makebox[0.242\columnwidth]{B: (14,12,20); 25\%}\\
\vspace{1pt}
\includegraphics[width=0.242\columnwidth]{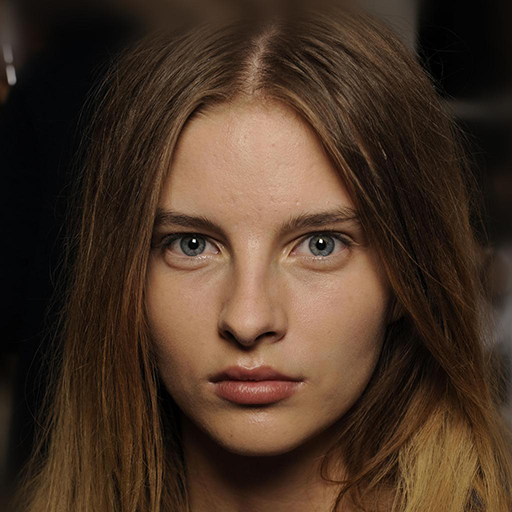}
\includegraphics[width=0.242\columnwidth]{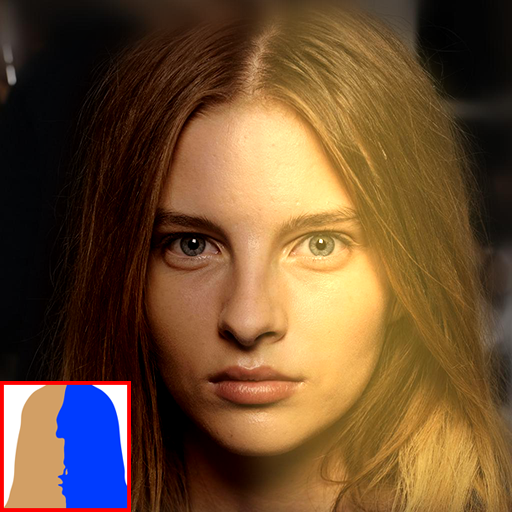}
\includegraphics[width=0.242\columnwidth]{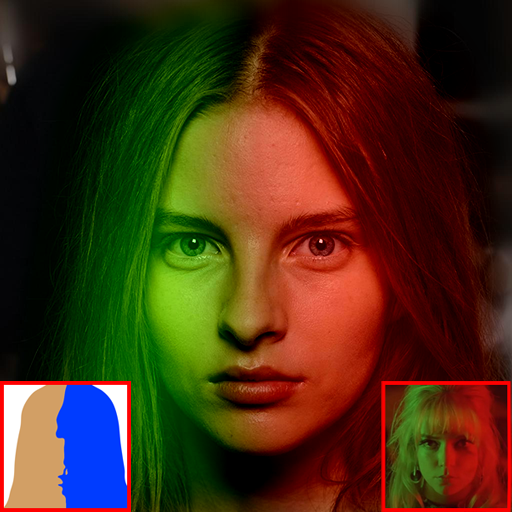}
\includegraphics[width=0.242\columnwidth]{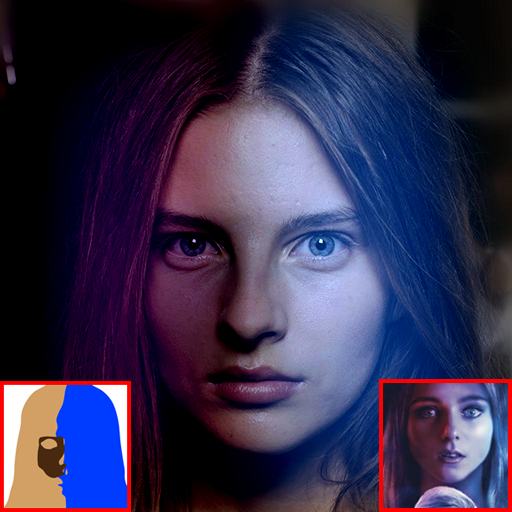}\\
\makebox[0.242\columnwidth]{Original}
\makebox[0.242\columnwidth]{B: (-9,-6,-5); 44\%}
\makebox[0.242\columnwidth]{B: (6,18,4); 85\%}
\makebox[0.242\columnwidth]{B: (7,5,9); 91\%}\\
\makebox[0.242\columnwidth]{}
\makebox[0.242\columnwidth]{D: (8,6,2); 81\%}
\makebox[0.242\columnwidth]{B: (8,3,3); 80\%}
\makebox[0.242\columnwidth]{B: (4,6,9); 100\%}\\
\makebox[0.242\columnwidth]{}
\makebox[0.242\columnwidth]{}
\makebox[0.242\columnwidth]{}
\makebox[0.242\columnwidth]{B: (-3,-9,-9); 58\%}\\
}
\caption{{Light editing. We apply linear dodge and linear burn to the middle-level PRs to modify highlights and shadows. D and B stand for linear dodge and linear burn, respectively. The 3-tuples are the integral of Laplacians of the PR and the percentage is the alpha value (opacity). We show the PR primitives and the references as insets.}}
\label{fig:light.transfer}
\end{figure}

{\textbf{Light editing.} Since our method separates illuminations from colors, the user can explicitly edit light using the PRs in the middle level. In Figure \ref{fig:light.transfer}, we show light editing results using the proposed blending functions.}

\begin{figure*}[htbp!]
\centering
{\scriptsize
\includegraphics[width=0.138\columnwidth]{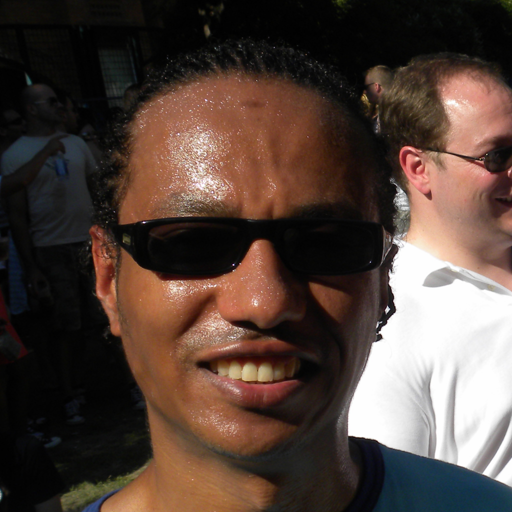}
\includegraphics[width=0.138\columnwidth]{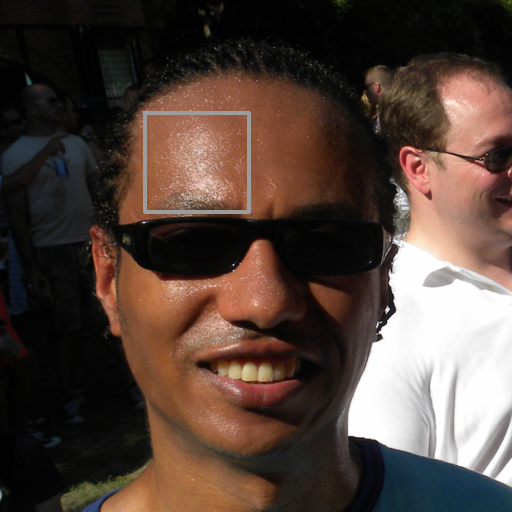}
\includegraphics[width=0.138\columnwidth]{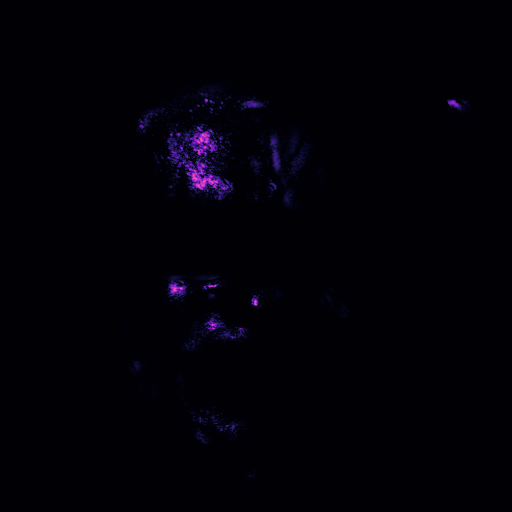}
\includegraphics[width=0.138\columnwidth]{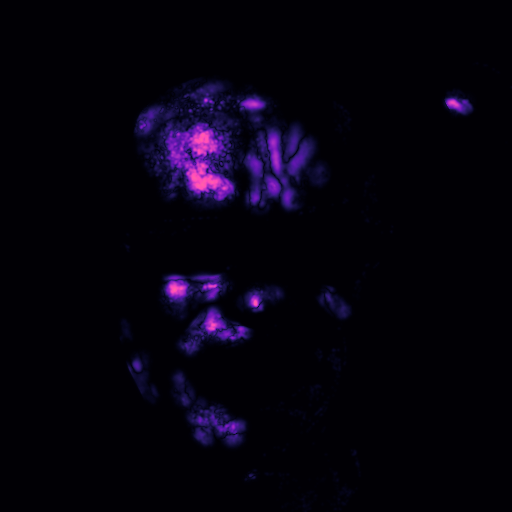}
\includegraphics[width=0.138\columnwidth]{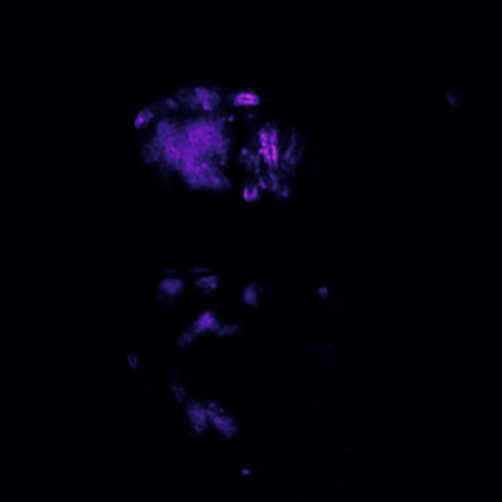}
\includegraphics[width=0.138\columnwidth]{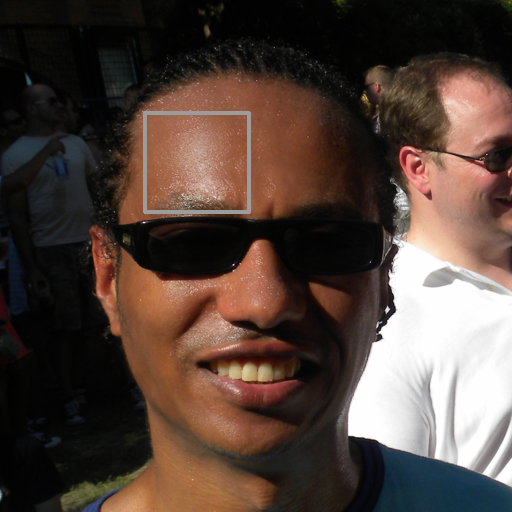}\\
\makebox[0.138\columnwidth]{Original}
\makebox[0.138\columnwidth]{Our method}
\makebox[0.138\columnwidth]{IS-FLIP}
\makebox[0.138\columnwidth]{FLIP}
\makebox[0.138\columnwidth]{SSIM}
\makebox[0.138\columnwidth]{Artist's}\\
\makebox[0.138\columnwidth]{}
\makebox[0.138\columnwidth]{}
\makebox[0.138\columnwidth]{0.39\%}
\makebox[0.138\columnwidth]{1.28\%}
\makebox[0.138\columnwidth]{0.72\%}
\makebox[0.138\columnwidth]{}\\
\vspace{1pt}
\includegraphics[width=0.138\columnwidth]{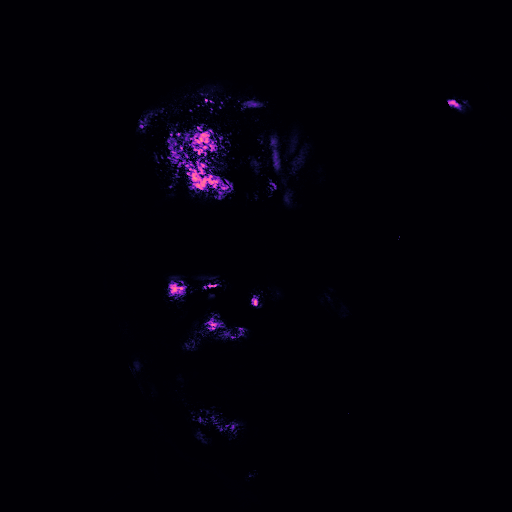}
\includegraphics[width=0.138\columnwidth]{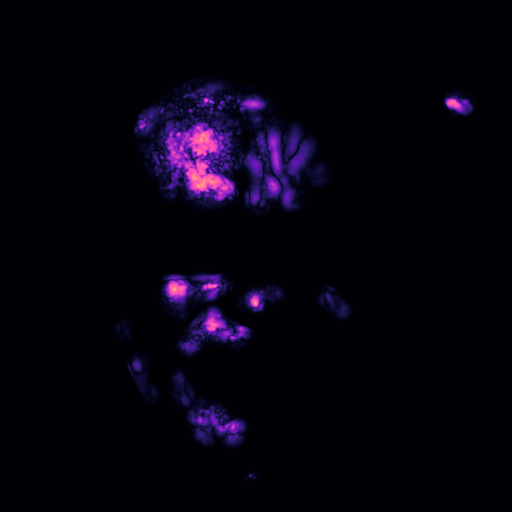}
\includegraphics[width=0.138\columnwidth]{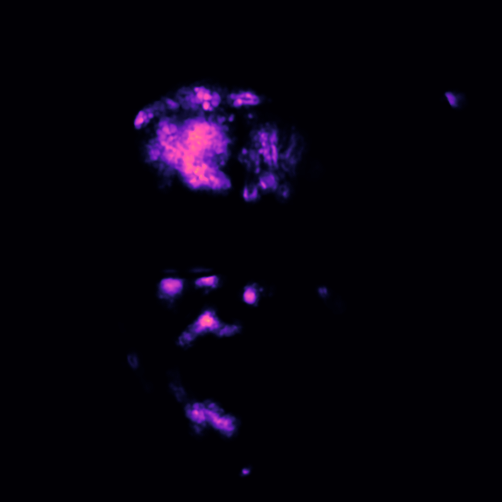}
\includegraphics[width=0.138\columnwidth]{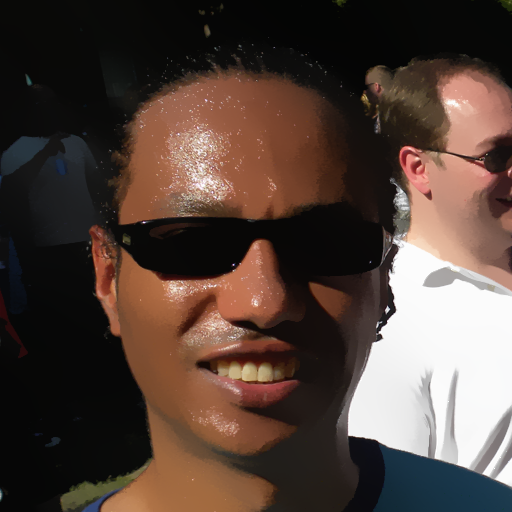}
\includegraphics[width=0.138\columnwidth]{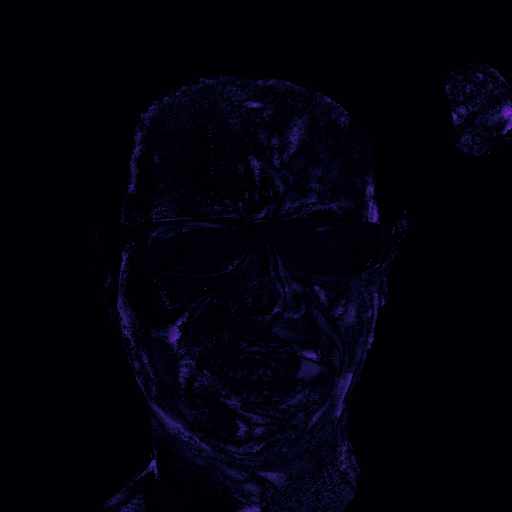}
\includegraphics[width=0.138\columnwidth]{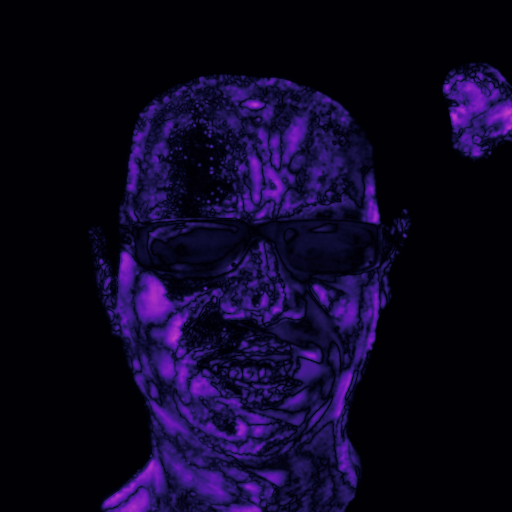}
\includegraphics[width=0.138\columnwidth]{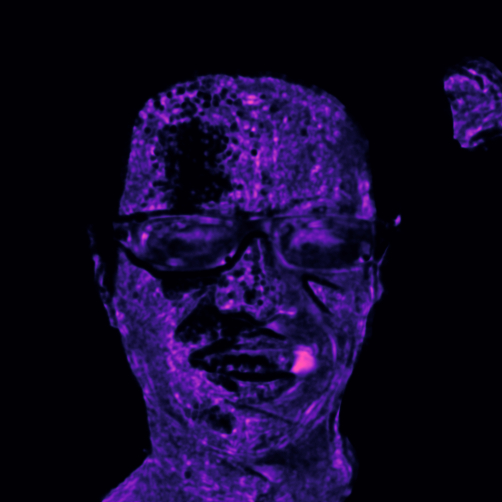}\\
\makebox[0.138\columnwidth]{IS-FLIP}
\makebox[0.138\columnwidth]{FLIP}
\makebox[0.138\columnwidth]{SSIM}
\makebox[0.138\columnwidth]{$L_1$ smoothing}
\makebox[0.138\columnwidth]{IS-FLIP}
\makebox[0.138\columnwidth]{FLIP}
\makebox[0.138\columnwidth]{SSIM}\\
\makebox[0.138\columnwidth]{0.52\%}
\makebox[0.138\columnwidth]{1.45\%}
\makebox[0.138\columnwidth]{1.3\%}
\makebox[0.138\columnwidth]{}
\makebox[0.138\columnwidth]{1.03\%}
\makebox[0.138\columnwidth]{4.02\%}
\makebox[0.138\columnwidth]{4.67\%}\\
\vspace{2pt}
\includegraphics[width=0.138\columnwidth]{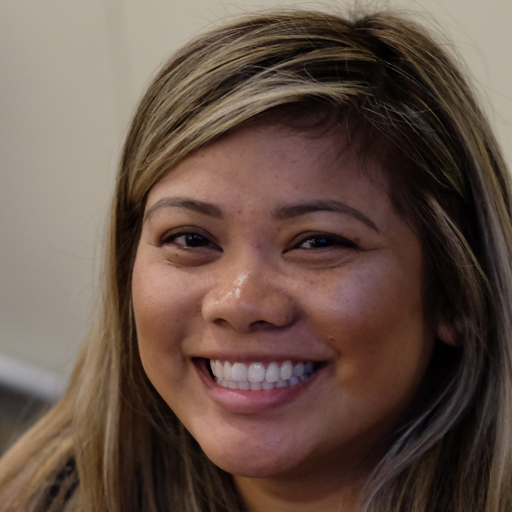}
\includegraphics[width=0.138\columnwidth]{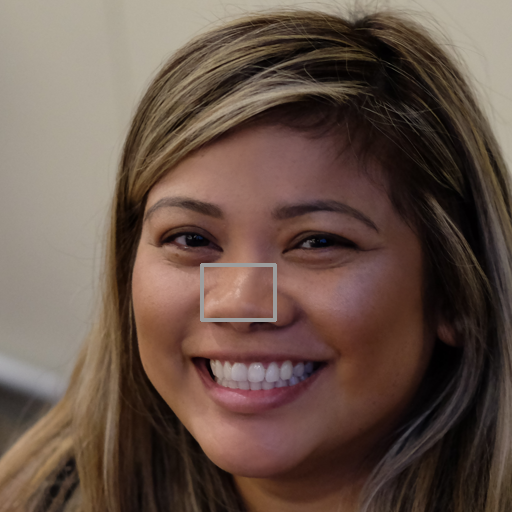}
\includegraphics[width=0.138\columnwidth]{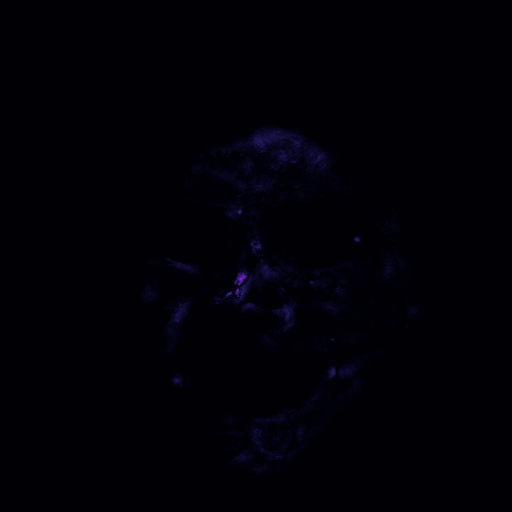}
\includegraphics[width=0.138\columnwidth]{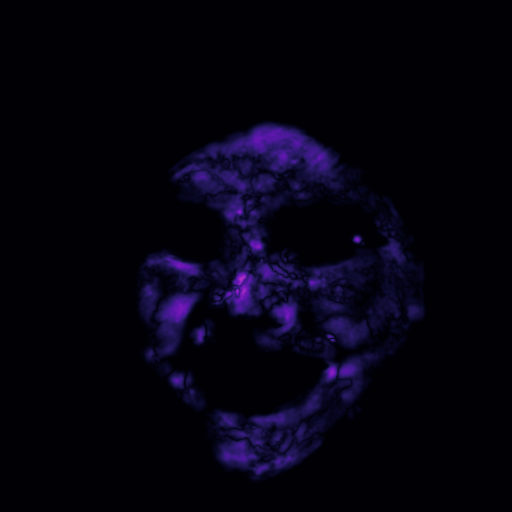}
\includegraphics[width=0.138\columnwidth]{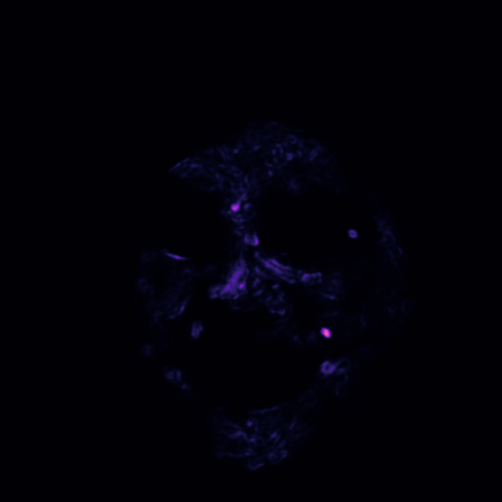}
\includegraphics[width=0.138\columnwidth]{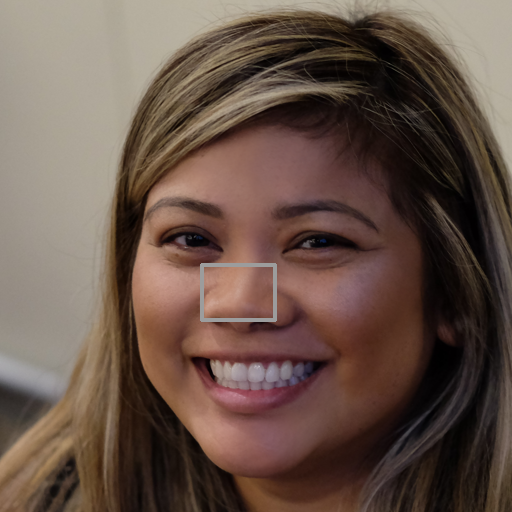}\\
\makebox[0.138\columnwidth]{Original}
\makebox[0.138\columnwidth]{Our method}
\makebox[0.138\columnwidth]{IS-FLIP}
\makebox[0.138\columnwidth]{FLIP}
\makebox[0.138\columnwidth]{SSIM}
\makebox[0.138\columnwidth]{Artist's}\\
\makebox[0.138\columnwidth]{}
\makebox[0.138\columnwidth]{}
\makebox[0.138\columnwidth]{0.31\%}
\makebox[0.138\columnwidth]{1.45\%}
\makebox[0.138\columnwidth]{0.48\%}
\makebox[0.138\columnwidth]{}\\
\vspace{1pt}
\includegraphics[width=0.138\columnwidth]{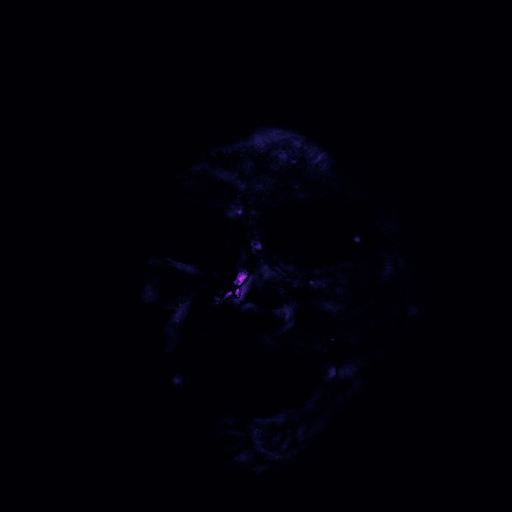}
\includegraphics[width=0.138\columnwidth]{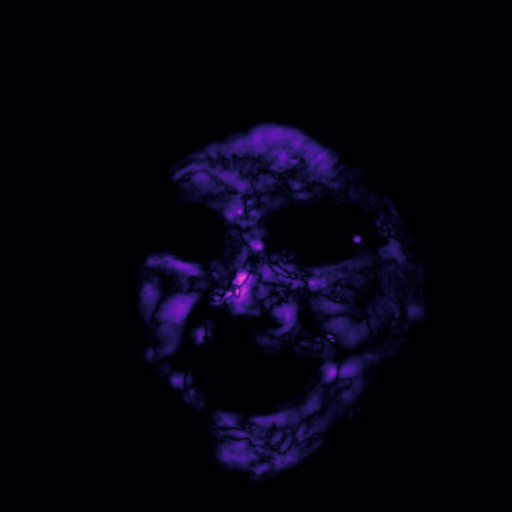}
\includegraphics[width=0.138\columnwidth]{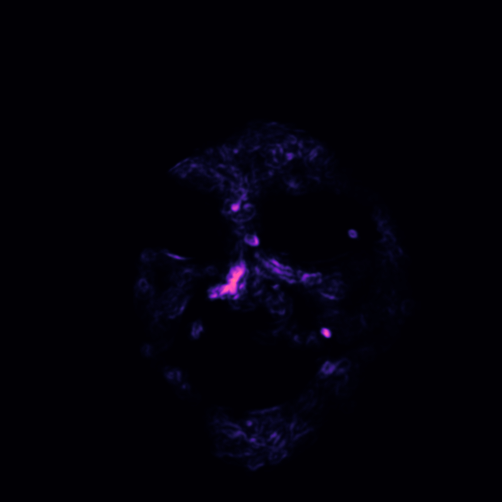}
\includegraphics[width=0.138\columnwidth]{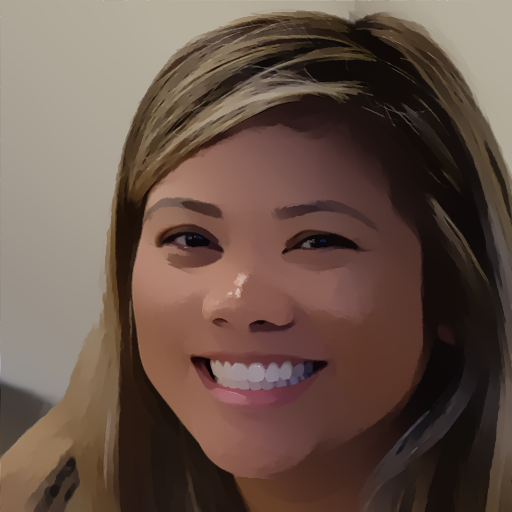}
\includegraphics[width=0.138\columnwidth]{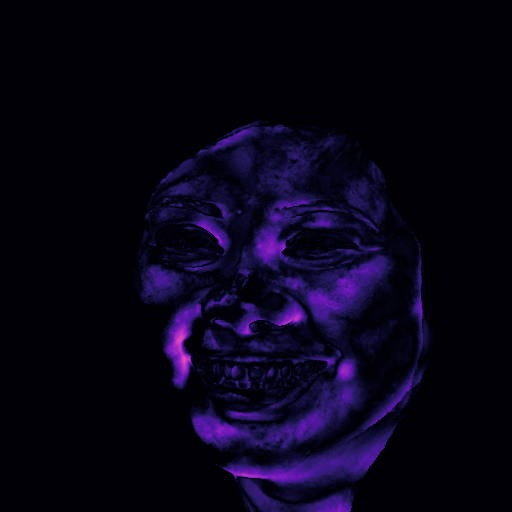}
\includegraphics[width=0.138\columnwidth]{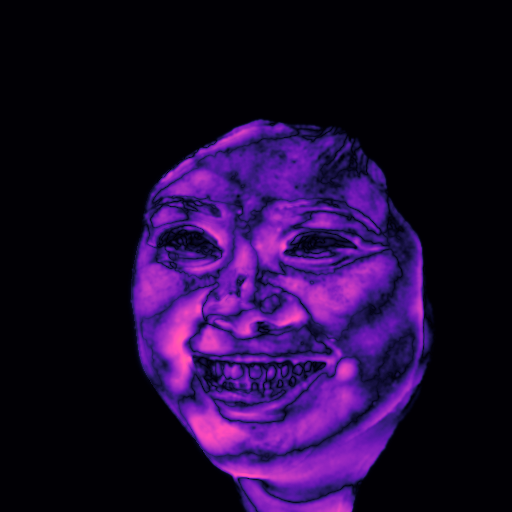}
\includegraphics[width=0.138\columnwidth]{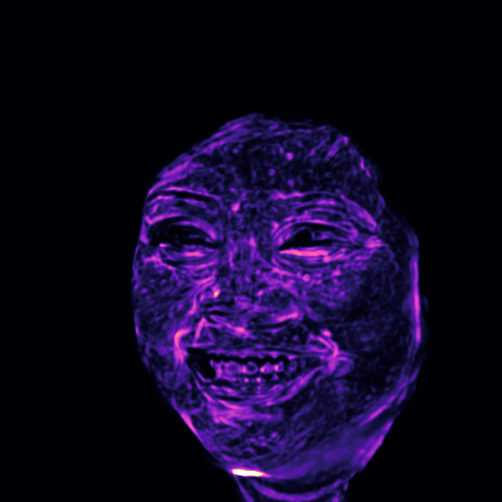}\\
\makebox[0.138\columnwidth]{IS-FLIP}
\makebox[0.138\columnwidth]{FLIP}
\makebox[0.138\columnwidth]{SSIM}
\makebox[0.138\columnwidth]{$L_1$ smoothing}
\makebox[0.138\columnwidth]{IS-FLIP}
\makebox[0.138\columnwidth]{FLIP}
\makebox[0.138\columnwidth]{SSIM}\\
\makebox[0.138\columnwidth]{0.34\%}
\makebox[0.138\columnwidth]{1.54\%}
\makebox[0.138\columnwidth]{0.63\%}
\makebox[0.138\columnwidth]{}
\makebox[0.138\columnwidth]{2.32\%}
\makebox[0.138\columnwidth]{6.82\%}
\makebox[0.138\columnwidth]{4.5\%}\\
\vspace{2pt}
\includegraphics[width=0.138\columnwidth]{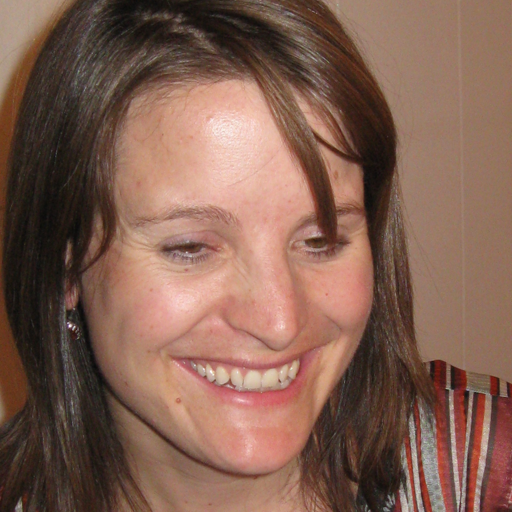}
\includegraphics[width=0.138\columnwidth]{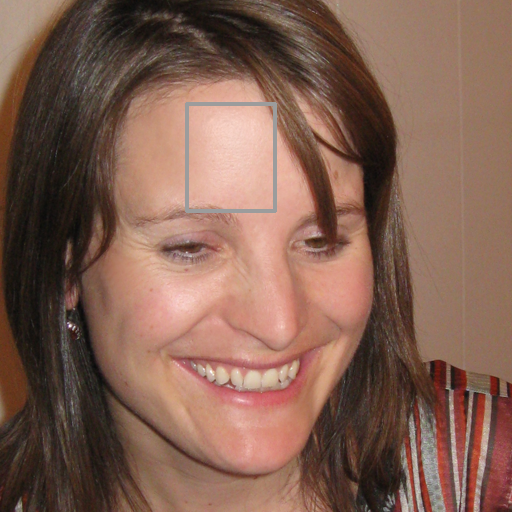}
\includegraphics[width=0.138\columnwidth]{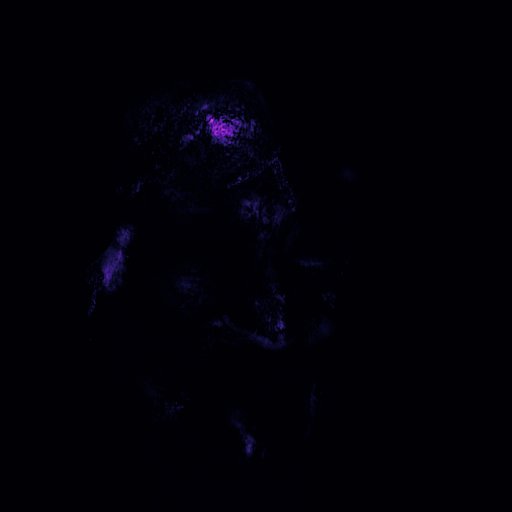}
\includegraphics[width=0.138\columnwidth]{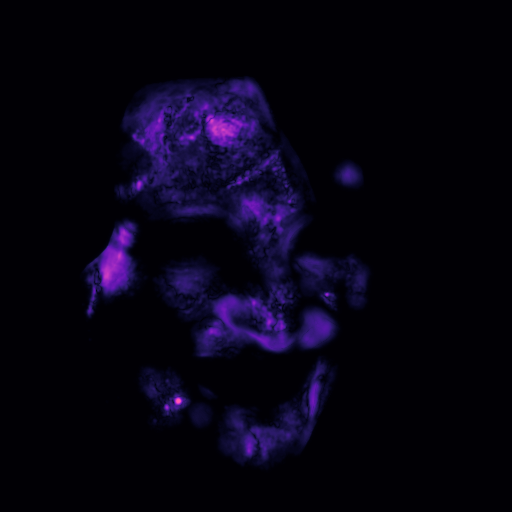}
\includegraphics[width=0.138\columnwidth]{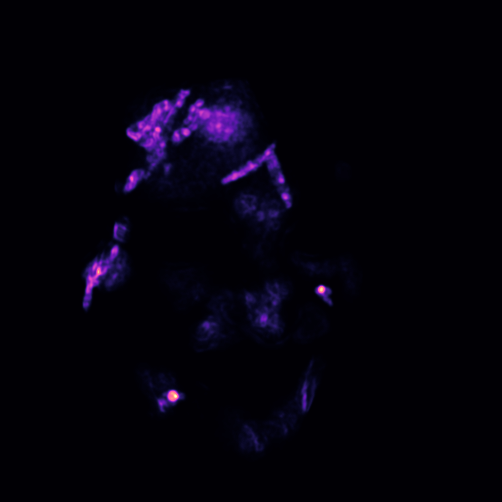}
\includegraphics[width=0.138\columnwidth]{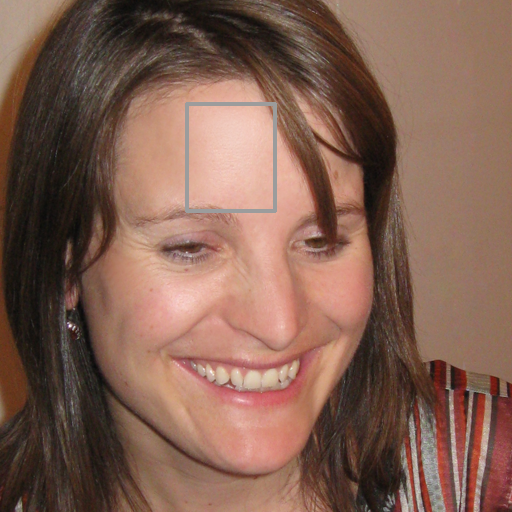}\\
\makebox[0.138\columnwidth]{Original}
\makebox[0.138\columnwidth]{Our method}
\makebox[0.138\columnwidth]{IS-FLIP}
\makebox[0.138\columnwidth]{FLIP}
\makebox[0.138\columnwidth]{SSIM}
\makebox[0.138\columnwidth]{Artist's}\\
\makebox[0.138\columnwidth]{}
\makebox[0.138\columnwidth]{}
\makebox[0.138\columnwidth]{0.37\%}
\makebox[0.138\columnwidth]{1.86\%}
\makebox[0.138\columnwidth]{0.98\%}
\makebox[0.138\columnwidth]{}\\
\vspace{1pt}
\includegraphics[width=0.138\columnwidth]{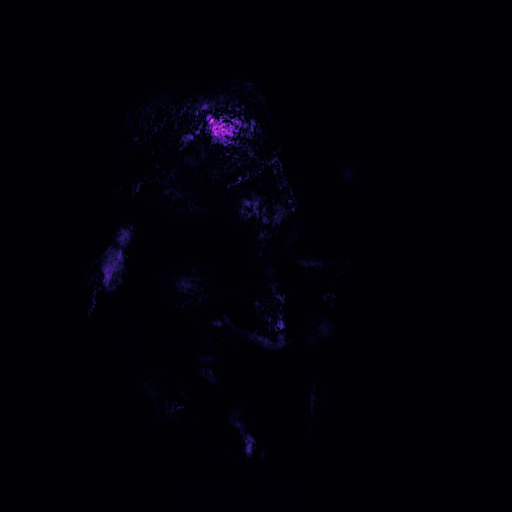}
\includegraphics[width=0.138\columnwidth]{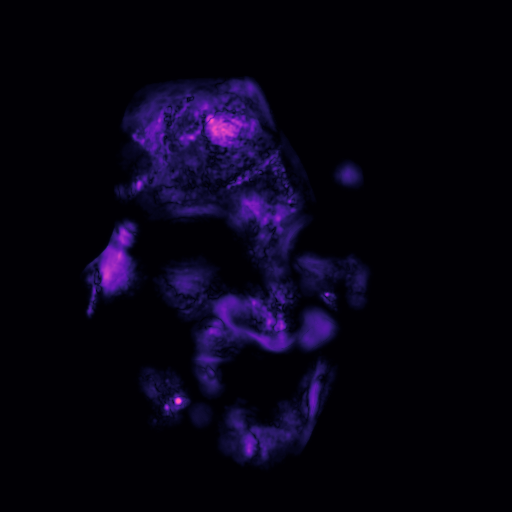}
\includegraphics[width=0.138\columnwidth]{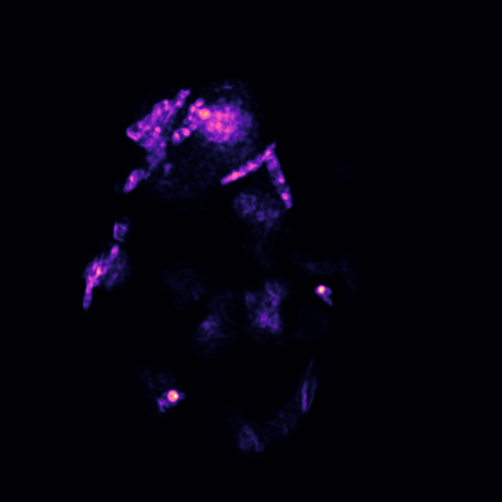}
\includegraphics[width=0.138\columnwidth]{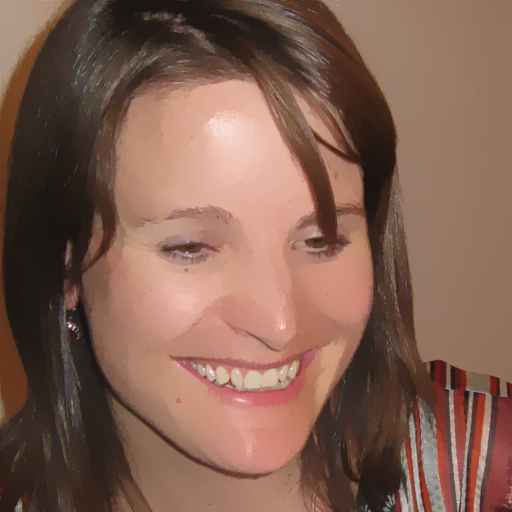}
\includegraphics[width=0.138\columnwidth]{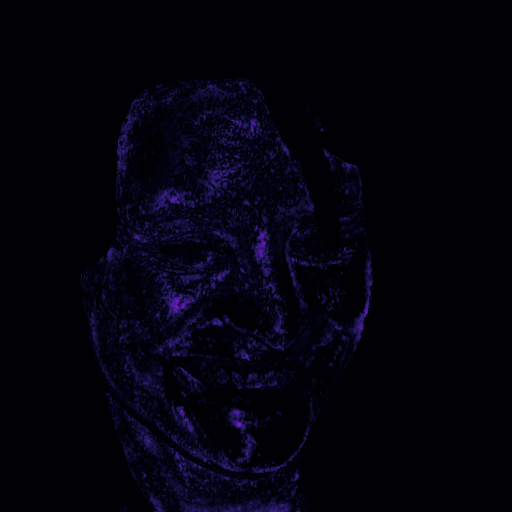}
\includegraphics[width=0.138\columnwidth]{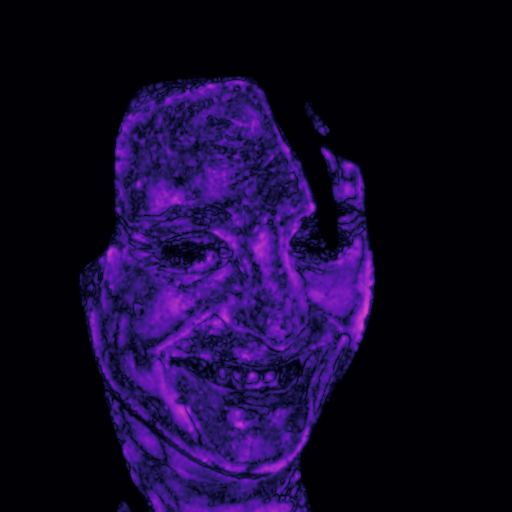}
\includegraphics[width=0.138\columnwidth]{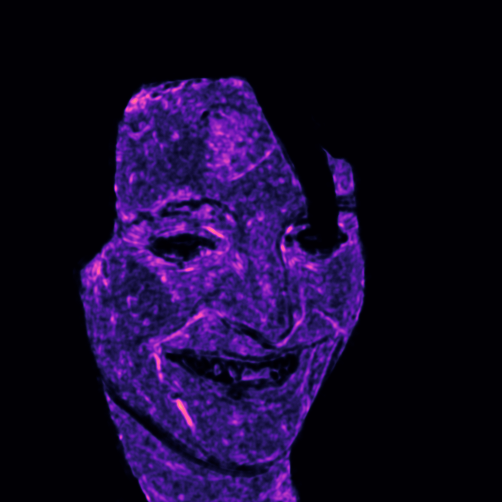}\\
\makebox[0.138\columnwidth]{IS-FLIP}
\makebox[0.138\columnwidth]{FLIP}
\makebox[0.138\columnwidth]{SSIM}
\makebox[0.138\columnwidth]{$L_1$ smoothing}
\makebox[0.138\columnwidth]{IS-FLIP}
\makebox[0.138\columnwidth]{FLIP}
\makebox[0.138\columnwidth]{SSIM}\\
\makebox[0.138\columnwidth]{0.4\%}
\makebox[0.138\columnwidth]{1.97\%}
\makebox[0.138\columnwidth]{1.18\%}
\makebox[0.138\columnwidth]{}
\makebox[0.138\columnwidth]{1.15\%}
\makebox[0.138\columnwidth]{5.25\%}
\makebox[0.138\columnwidth]{5.25\%}\\}
\caption{{Comparing our face retouching results with the $L_1$ smoothing method \cite{bi20151l1} and manual editing by professional artists \citep{shafaei2021auturetouch}. Our method simply removes fPRs, and keeps other primitives unchanged.
The $L_1$ smoothing method can keep illumination, but often over-smoothens salient features. Therefore, our method can effectively preserve illumination, yielding more realistic results than the $L_1$ smoothing method. }
{Our results are also better than manual retouching, which are justified by the three quantitative measures. This is because that manual editing hardly reaches pixel-level precision and often compromises original features, such as highlights.} Quantitative evaluations in terms of FLIP, IS-FLIP and SSIM also confirm our observation. The images are in high resolution, allowing zoom-in examination.}
\label{fig:details_new}
\end{figure*}

\begin{figure}[htbp]
\centering
{\scriptsize
\includegraphics[width=0.158\columnwidth]{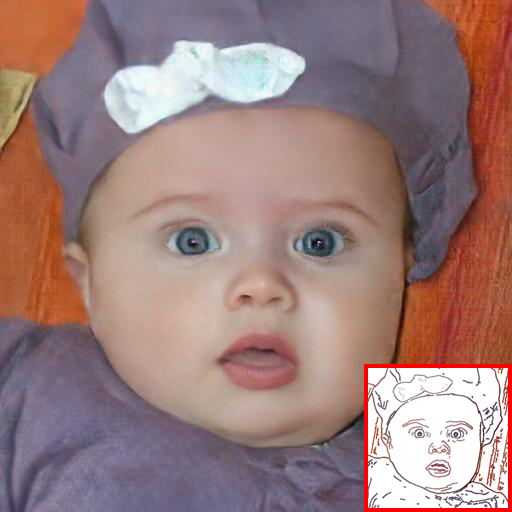}
\includegraphics[width=0.158\columnwidth]{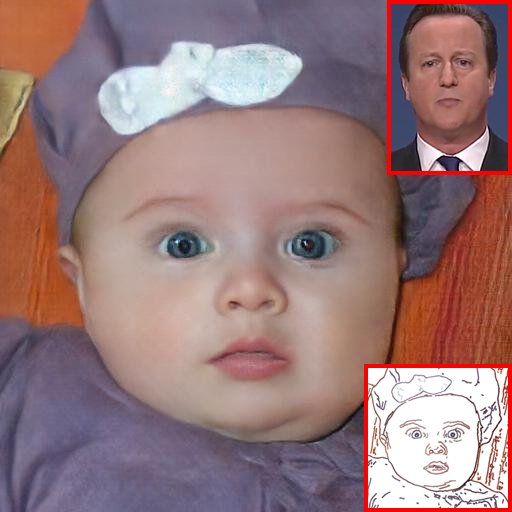}
\includegraphics[width=0.158\columnwidth]{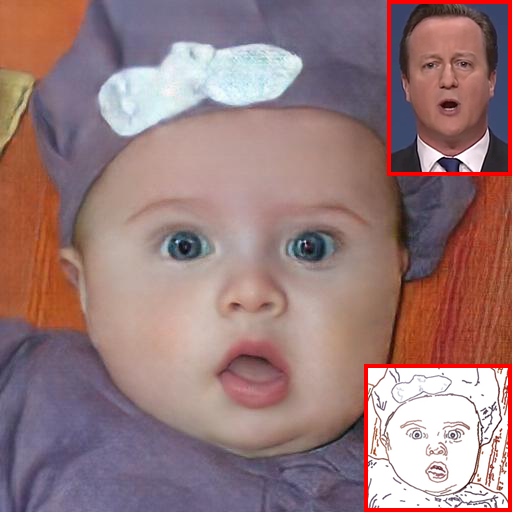}
\includegraphics[width=0.158\columnwidth]{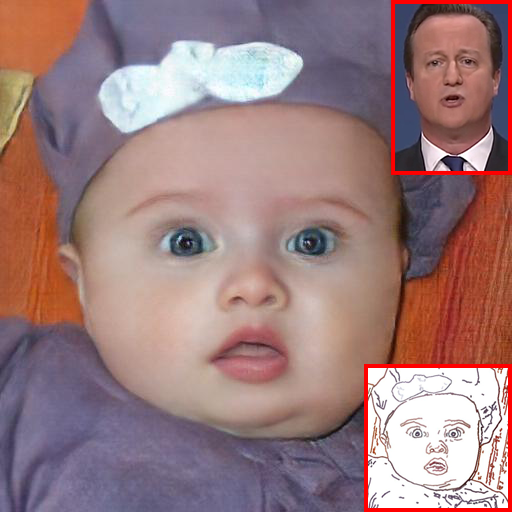}
\includegraphics[width=0.158\columnwidth]{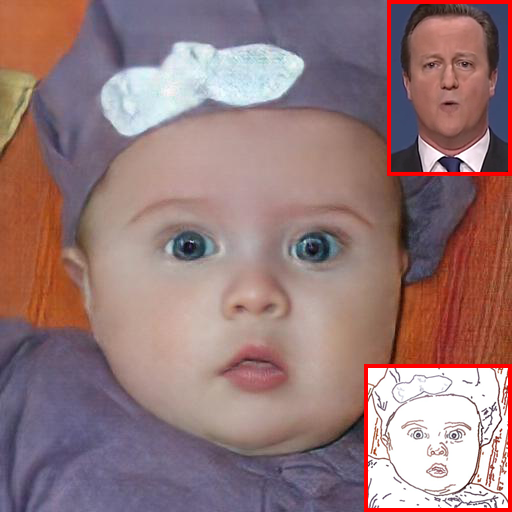}
\includegraphics[width=0.158\columnwidth]{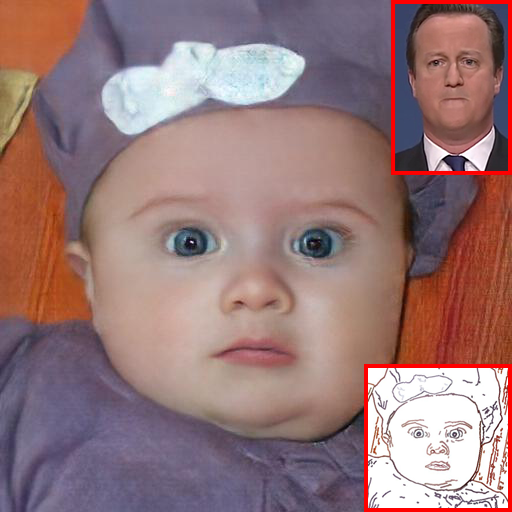}\\
\vspace{2pt}
\includegraphics[width=0.158\columnwidth]{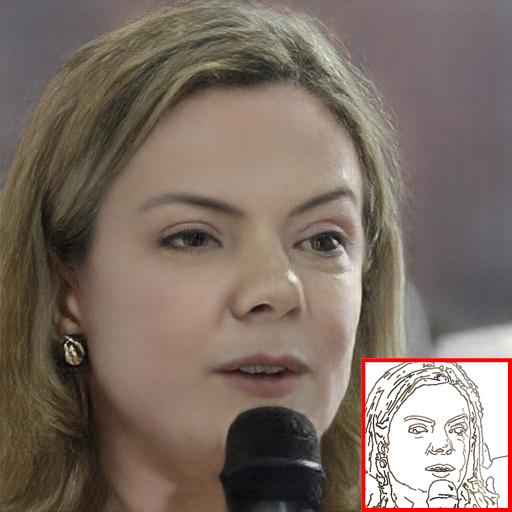}
\includegraphics[width=0.158\columnwidth]{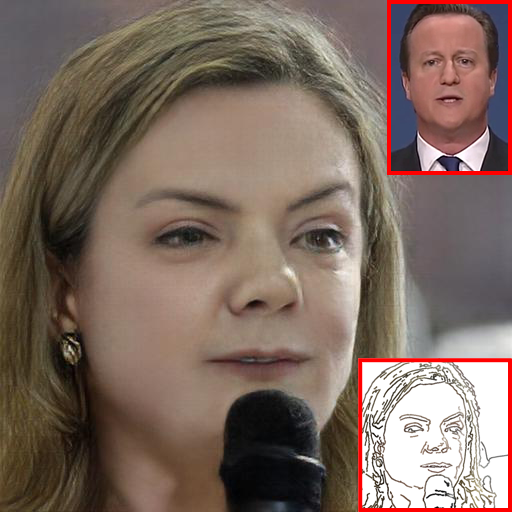}
\includegraphics[width=0.158\columnwidth]{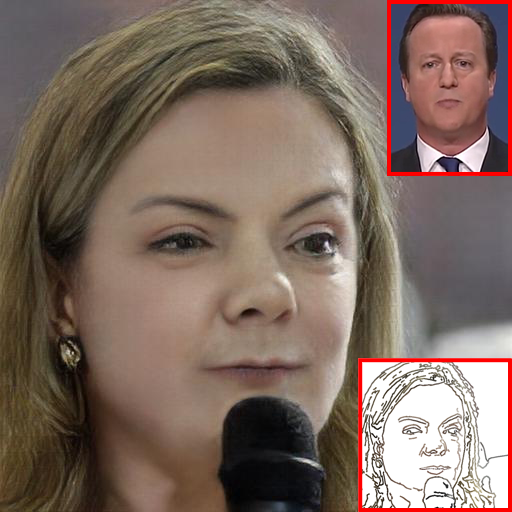}
\includegraphics[width=0.158\columnwidth]{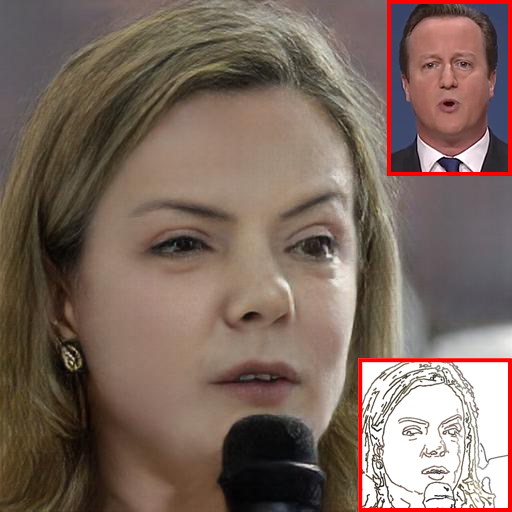}
\includegraphics[width=0.158\columnwidth]{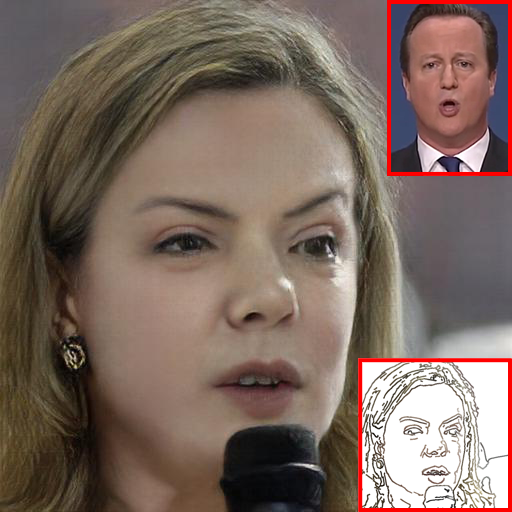}
\includegraphics[width=0.158\columnwidth]{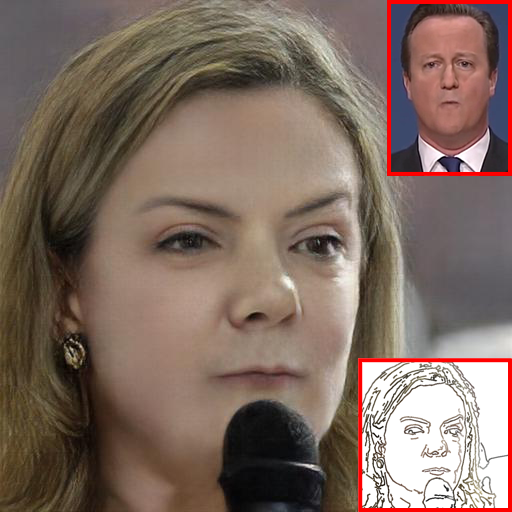}\\
\vspace{2pt}
\includegraphics[width=0.158\columnwidth]{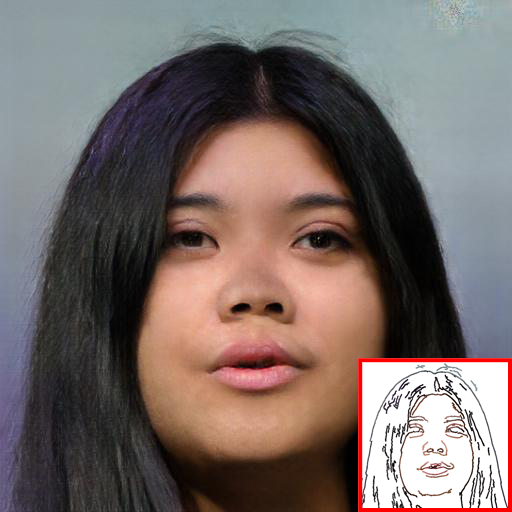}
\includegraphics[width=0.158\columnwidth]{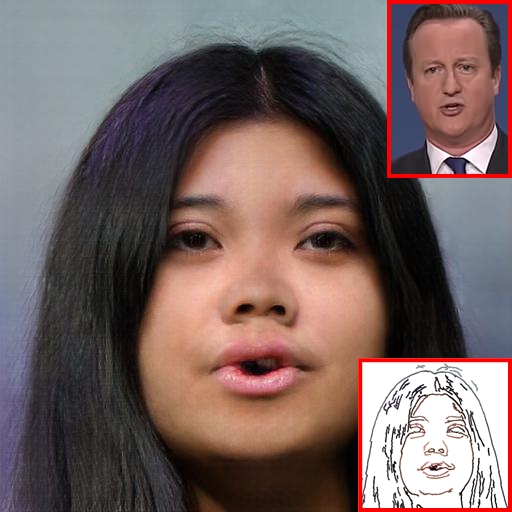}
\includegraphics[width=0.158\columnwidth]{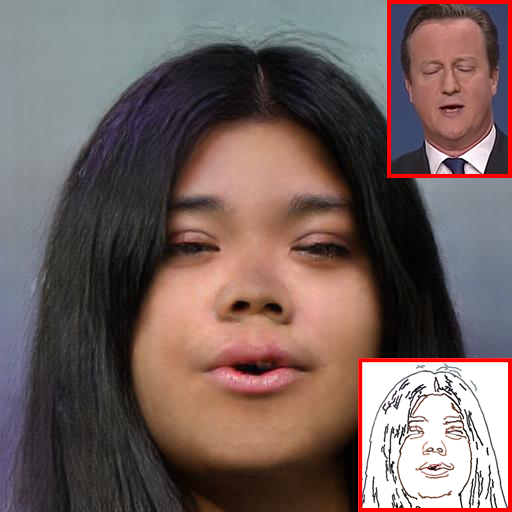}
\includegraphics[width=0.158\columnwidth]{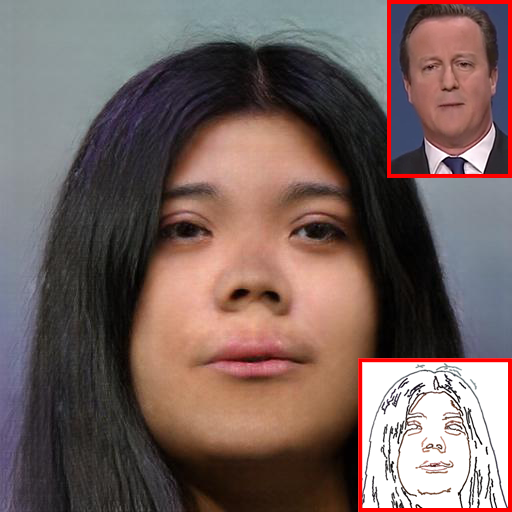}
\includegraphics[width=0.158\columnwidth]{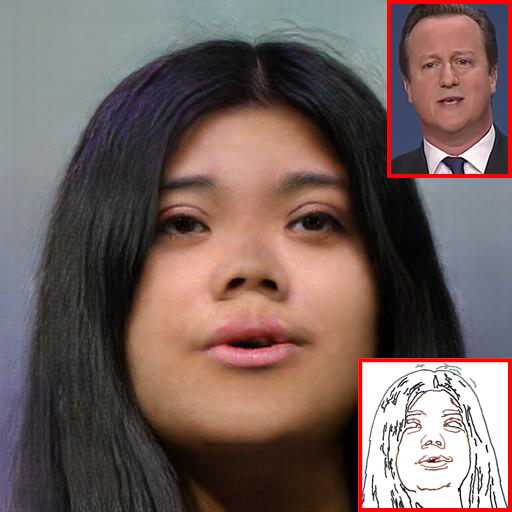}
\includegraphics[width=0.158\columnwidth]{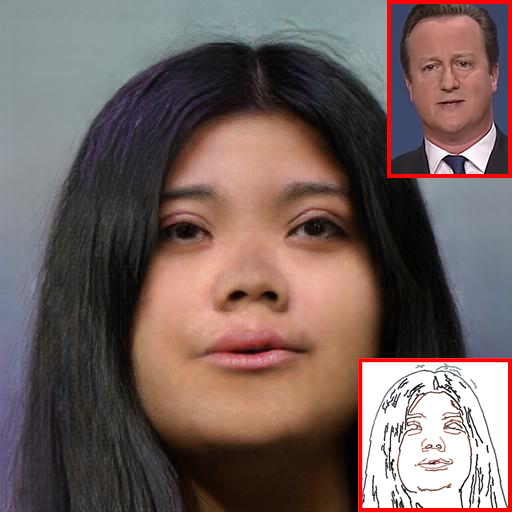}\\
}
\caption{{Expression editing via changing the geometries of DCs in the base level. The first column shows the original images, and the others are the editing results.} {We show the references and the original and updated DCs as insets.}
}
\label{fig:geometry}
\end{figure}

{\textbf{Expression editing.} Since there are only a few sparse diffusion curves in the base level, geometry editing becomes easy. Figure \ref{fig:geometry} shows example of expression editing, in which the user changes the geometries of the diffusion curves. The fine details and high-frequency residuals are generated automatically by a deep generative model. }

\textbf{Comparison with DC vectorization.} The existing diffusion curve based vectorization methods \citep{orzan2008diffusion,favreau2017photo2clipart,zhao2017inverse,xie2014hierarchical} often produce a large amount of vector primitives that make post-editing difficult.
Our {hierarchical PVG can significantly reduce the user's burden since the base and middle levels contain a small number of vector primitives (e.g., a few tens). \fq{The user can edit the middle-level PRs by using linearly blending to modify highlights and shadows (see Figure \ref{fig:light.transfer}).} Though there are a large amount of pixel-sized PRs in the top level, they are not meant for direct editing by the user.
{Turning off fPRs yields face retouching results. As Figure \ref{fig:details_new} shows, our retouched results are of higher quality than artist's manual retouching \citep{shafaei2021auturetouch} and L1-smoothing~\cite{bi20151l1}. }
In geometry editing, when diffusion curves in the base level are changed, we adopt a deep neural network model to generate residual PRs automatically (see Figure \ref{fig:geometry}).
Also applying the DC masks to the changed DCs, we can remove lights in the generated residual Poisson regions (see Figure \ref{fig:dc_ai}(e)).}

\textbf{Comparison with~\citep{fu2019vectorization}.} The existing PVG vectorization method~\citep{fu2019vectorization} converts a portrait photo into a PVG with only two {levels}: the base {level} consists of diffusion curves for the low-frequency skin and hair colors, and the top {level} contains everything else. {Their 2-level decomposition works well for color transfer, which involves diffusion curves only. However, since their method does not {separate illumination and facial details from skin colors}, it is difficult apply their method to  illumination editing, geometry editing, and face retouching. It is also worth noting that though we adopt the same method for transferring boundary colors between diffusion curves as in \citet{fu2019vectorization},
our color transfer results are better in terms of both visual comparison and quantitative measure as shown in Figure~\ref{fig:color.transfer}. This is due to the fact that our DC extraction can separate illumination from low-frequency colors better than \citet{fu2019vectorization}.}

\begin{figure}[!htbp]
\setlength\tabcolsep{1pt}
\centering
\begin{tabular}{c|ccc|cc}
\includegraphics[width=0.158\columnwidth]{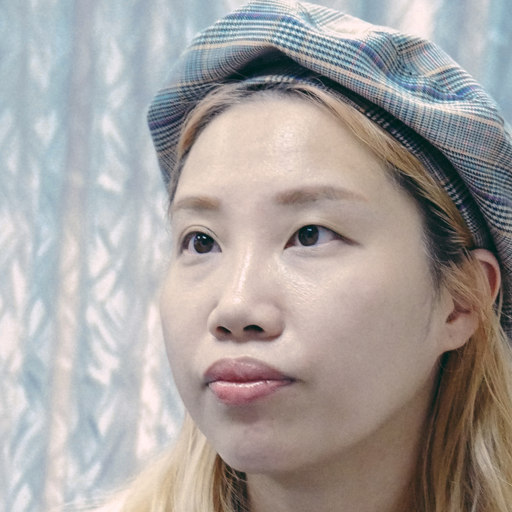} &
\includegraphics[width=0.158\columnwidth]{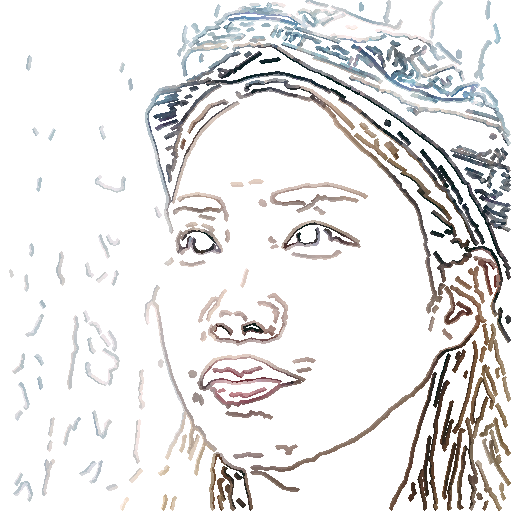} &
\includegraphics[width=0.158\columnwidth]{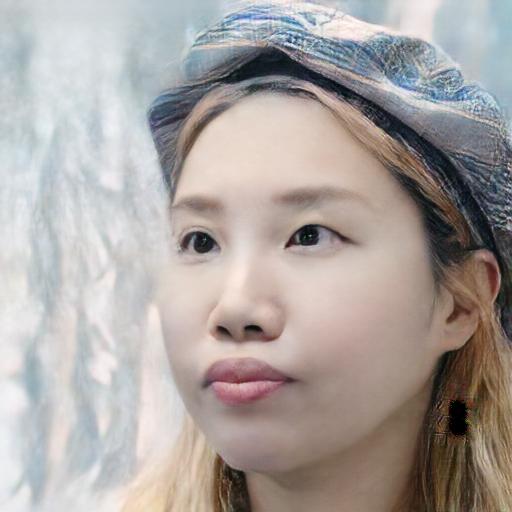} &
\includegraphics[width=0.158\columnwidth]{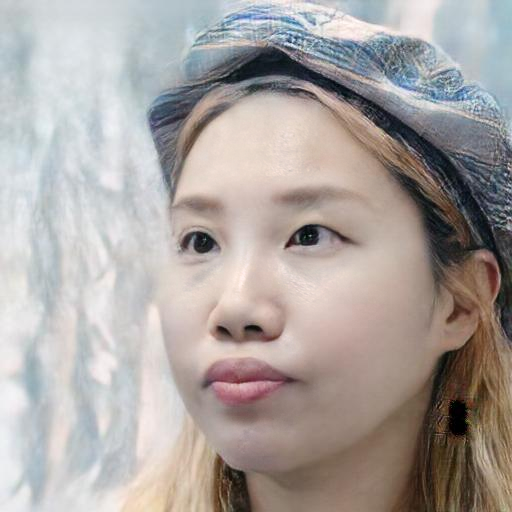} &
\includegraphics[width=0.158\columnwidth]{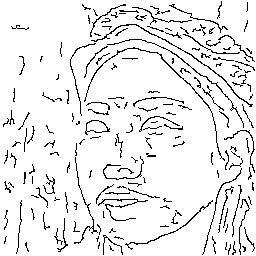} &
\includegraphics[width=0.158\columnwidth]{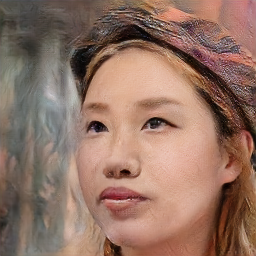}\\
& {\scriptsize4.0\% px} & & & {\scriptsize3.1\% px} & \\
\includegraphics[width=0.158\columnwidth]{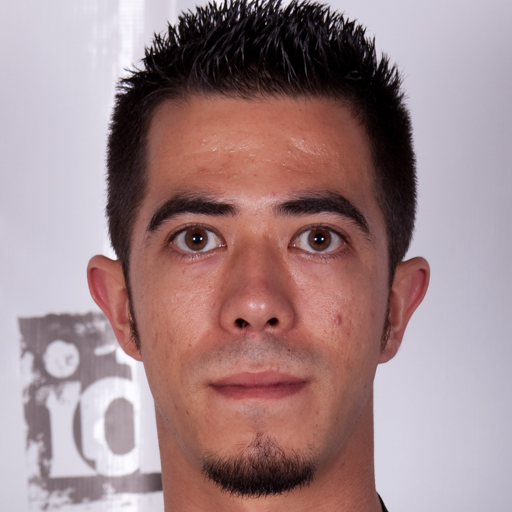}&
\includegraphics[width=0.158\columnwidth]{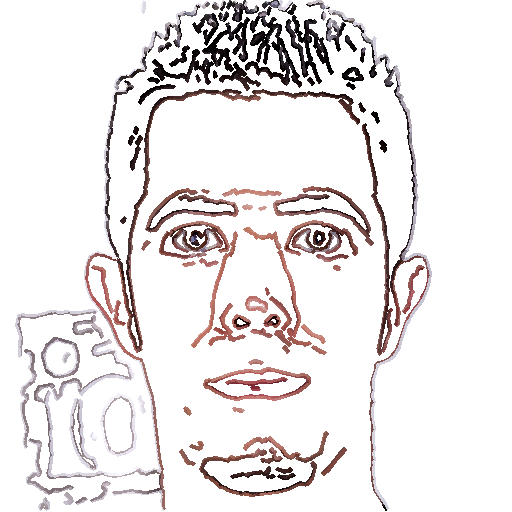}&
\includegraphics[width=0.158\columnwidth]{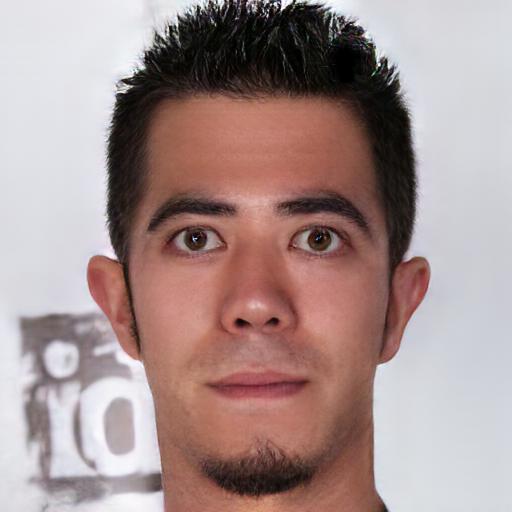}&
\includegraphics[width=0.158\columnwidth]{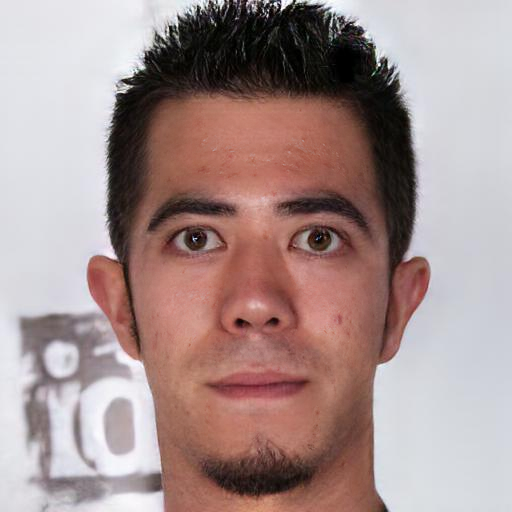}&
\includegraphics[width=0.158\columnwidth]{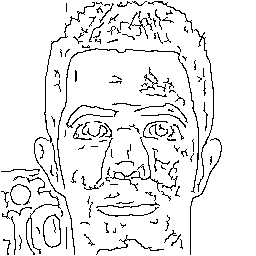}&
\includegraphics[width=0.158\columnwidth]{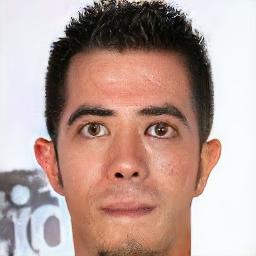}\\
&{\scriptsize2.9\% px}&&&{\scriptsize3.1\% px}&\\
\includegraphics[width=0.158\columnwidth]{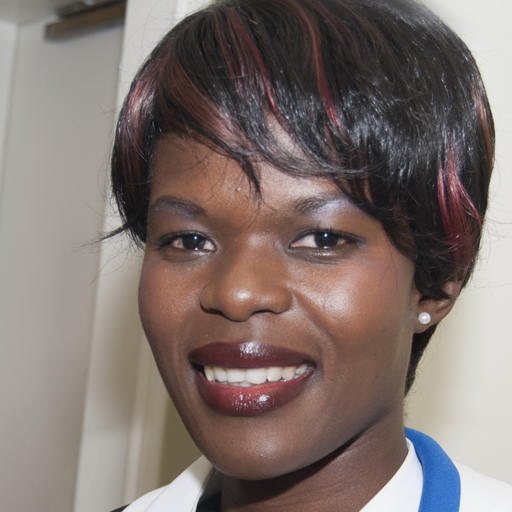}&
\includegraphics[width=0.158\columnwidth]{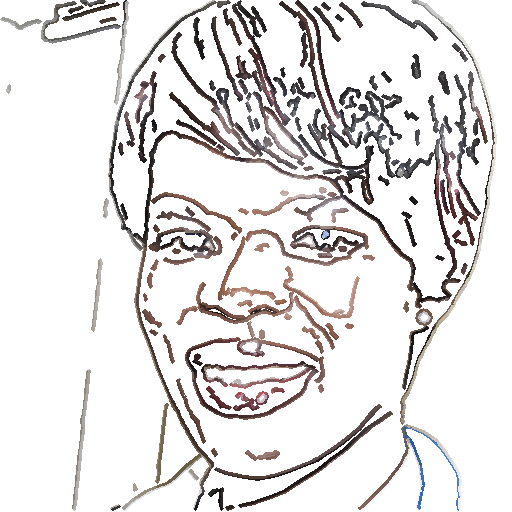}&
\includegraphics[width=0.158\columnwidth]{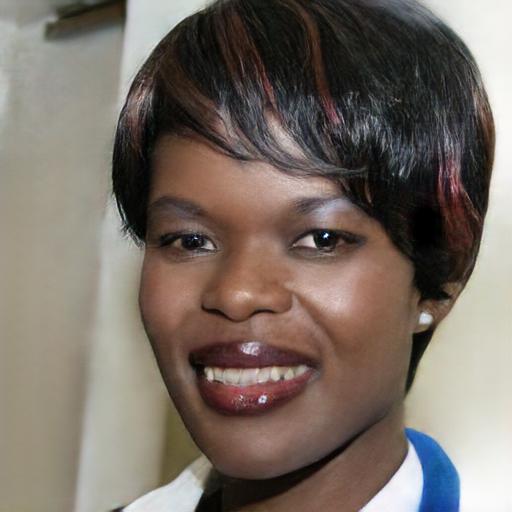}&
\includegraphics[width=0.158\columnwidth]{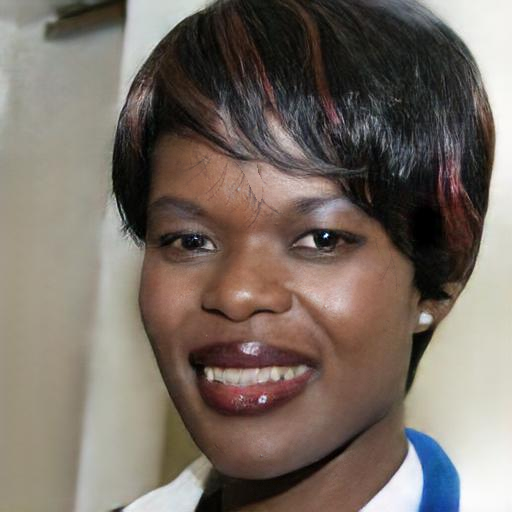}&
\includegraphics[width=0.158\columnwidth]{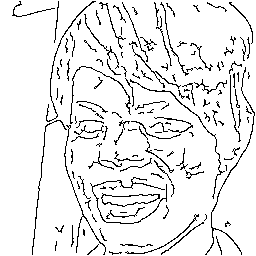}&
\includegraphics[width=0.158\columnwidth]{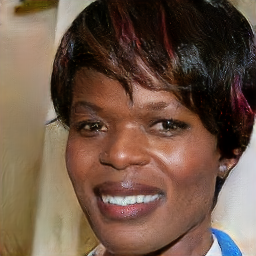}\\
&{\scriptsize3.5\% px}&&&{\scriptsize3.2\% px}&\\
\includegraphics[width=0.158\columnwidth]{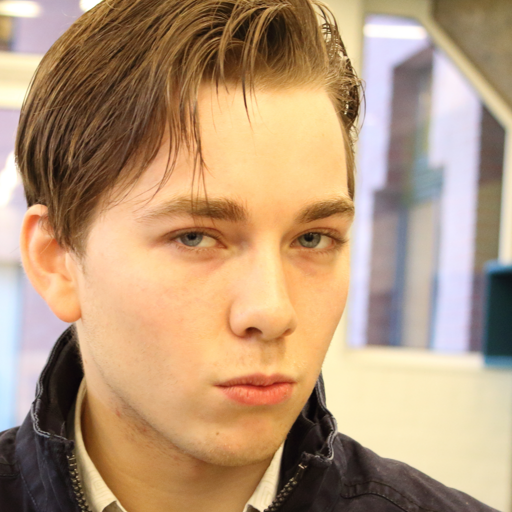}&
\includegraphics[width=0.158\columnwidth]{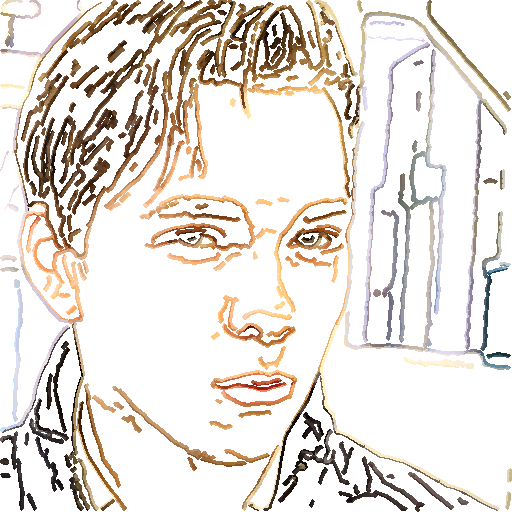}&
\includegraphics[width=0.158\columnwidth]{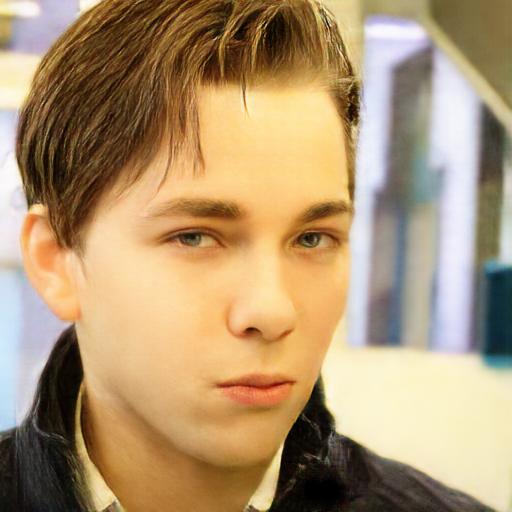}&
\includegraphics[width=0.158\columnwidth]{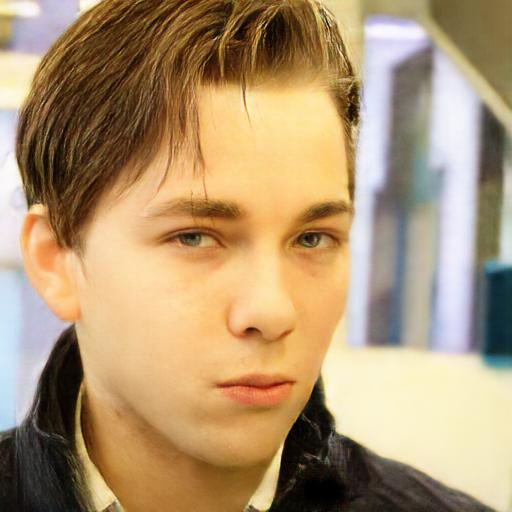}&
\includegraphics[width=0.158\columnwidth]{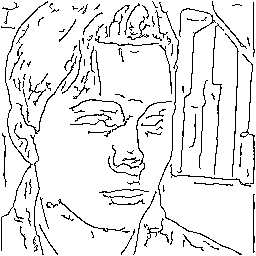}&
\includegraphics[width=0.158\columnwidth]{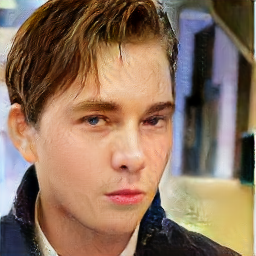}\\
{\scriptsize(a)}&{\scriptsize(b) 4.8\% px}&{\scriptsize(c) DC}&{\scriptsize(d) DC+rPR'}&{\scriptsize(e) 3.4\% px}&{\scriptsize(f)}\\
{\scriptsize Original}&{\scriptsize DC}&{\scriptsize+rPR'}&{\scriptsize+fPR}&{\scriptsize contour}&{\scriptsize\cite{dekel2018sparse}}\\
\end{tabular}
\caption{{Comparison with deep sparse, smart contours~\cite{dekel2018sparse} in image reconstruction.
Our method produces diffusion curves in a similar amount as smart contours.
To make a fair comparison, we generate the residual Poisson regions rPR' using the deep neural network.
We observe that our method can better preserve lights and colors than their method even without fPRs (see (c) vs (f)). The px ratios show the complexity of vector primitives (DC or contours).}}
\label{fig:dc_ai_compare}
\end{figure}

\textbf{Comparison with deep learning based methods.}
The smart contour method~\cite{dekel2018sparse} uses a cascade of two deep neural networks, where the low-frequency network encodes a raster image by a sparse set of contours, which are associated with color gradients, for overall colors and geometries, and the high-frequency network generates residuals. Their method is flexible and efficient in that it allows the user to freely edit the contours, such adding and deleting contours and changing their geometries, and obtain photo-realistic results in near real-time. Our method also allows the user to edit geometries by changing diffusion curves and then automatically generates residuals using deep learning. As Figure \ref{fig:dc_ai_compare} shows, both methods produce similar amount of vector primitives (i.e., diffusion curves and smart contours) in the base level. However, thanks to Poisson regions in the middle level, our method allows the user to edit illumination in an easy and intuitive manner, whereas their method cannot. We also observe that our method can better preserve fine details in image reconstruction thanks to fPRs in the top level.
\fq{Moreover, our hierarchical representation provides an integral framework for portrait coarse-to-fine editing, which conjoins the advantages of both vectorization and deep learning based methods.
Except for expression manipulation that requiring high-frequency signals, the other presented editing procedures only consist of vector primitive (curves or regions, i.e., DCs or PRs) modifications and PVG rendering.
Comparing with deep learning method, vectorization based editing can provide more subtle control, for example, in Figure \ref{fig:hair_color}, we can control exactly which hairlines are modified.
And also it largely economizes computational complexity and memory since the editing process does not involve loading or running any neural models during vector primitives modification and rendering.
}

\begin{figure}[htbp]
\centering
{\scriptsize
\includegraphics[width=0.158\columnwidth]{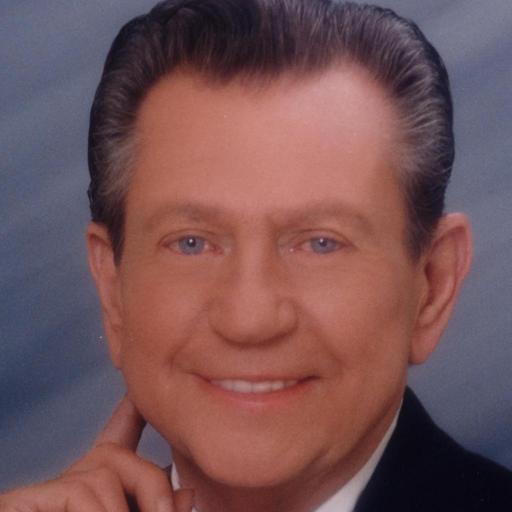}
\includegraphics[width=0.158\columnwidth]{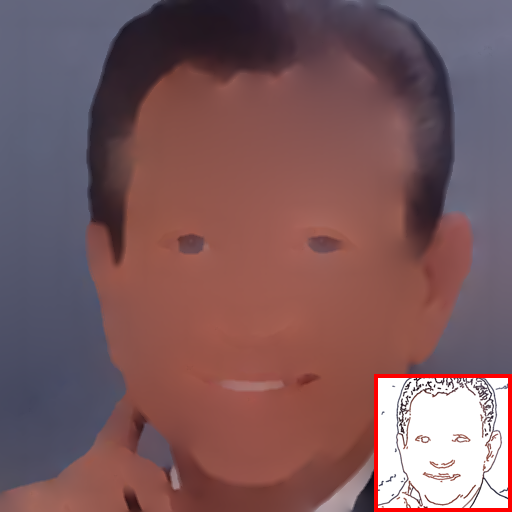}
\includegraphics[width=0.158\columnwidth]{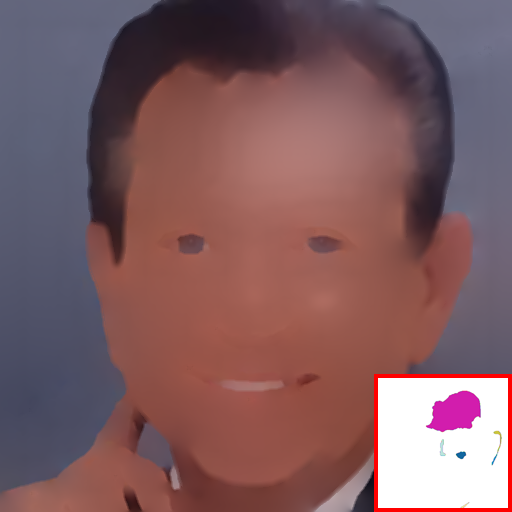}
\includegraphics[width=0.158\columnwidth]{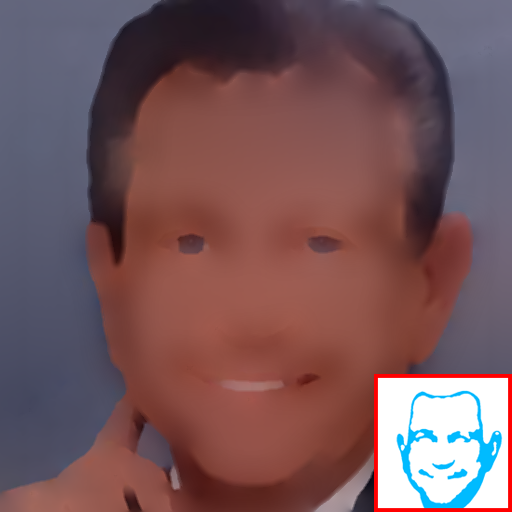}
\includegraphics[width=0.158\columnwidth]{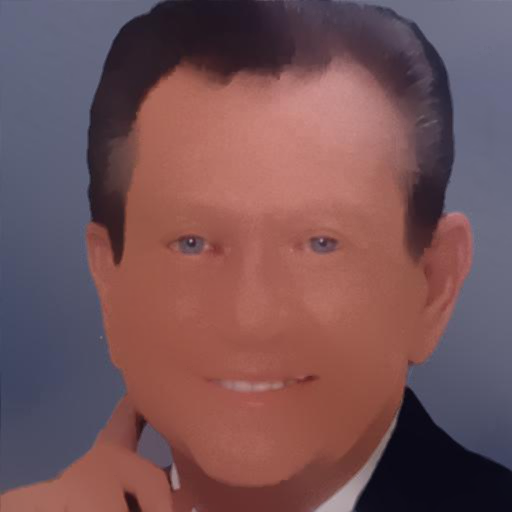}
\includegraphics[width=0.158\columnwidth]{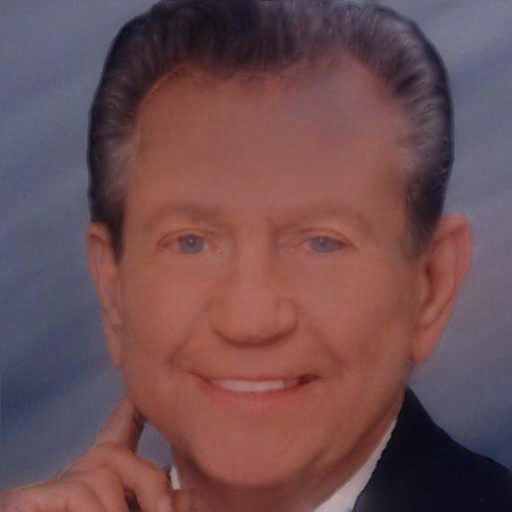}\\
\makebox[0.158\columnwidth]{Original}
\makebox[0.158\columnwidth]{DC}
\makebox[0.158\columnwidth]{DC+hPR}
\makebox[0.158\columnwidth]{DC+sPR}
\makebox[0.158\columnwidth]{DC+rPR}
\makebox[0.158\columnwidth]{DC+fPR}\\
}
\caption{A failed case. The vector primitives in the base level are not able to capture the facial features due to blurred input.}
\label{fig:fail}
\end{figure}

\textbf{Failed case.} Our method may fail if the input images are blurred due to out of focus and/or low contrast. This is because our method could not extract sufficient DCs for representing salient facial features.
Figure \ref{fig:fail} shows a failed example.


\section{Conclusion}
We developed an automatic method for vectorizing portrait images. In contrast to the existing methods that generate a large number of diffusion curves for reconstructing fine details of the input, our method produces a reasonable amount of editable vector primitives without compromising the accuracy. We organize the primitives into three levels. The base level is a set of sparse diffusion curves for representing low-frequency colors and salient facial features. The middle level consists of a few large Poisson regions for specular highlights and shadows. The top level contains pixel-sized Poisson regions for fine details and high-frequency residuals.
We demonstrate that the hierarchical vector representation facilitates portrait editing, such as color transfer, light and expression editing.

Although our proposed method has a range of desirable properties, such as its full automation and its high flexibility, there are also open points that we aim to address in the future.
{For example, we can further improve the hierarchical vectorization by adding other semantic levels for additional facial features, such as tattoos. We only develop 3 simple blending functions, i.e., alpha blending, linear dodge and linear burn, in our current implementation. In the future, we will add more linear and non-linear blending modes to further expand the applications of PVG.
We will also extend our method from still images to videos by considering spatial and temporal coherence of diffusion curves and Poisson regions between frames.}
{Last but not the least, our current method is particularly designed for portrait images, thanks to various portrait-tailored computational tools available, such as segmentation and highlight removal. We believe that the proposed hierarchical PVG vectorization framework is general and can be extended to other images given proper pre-processing tools.}

\bibliography{hpvg}

\end{document}